\documentclass{article}

\usepackage[nonatbib,preprint]{neurips_2026}

\usepackage[utf8]{inputenc} 
\usepackage[T1]{fontenc}    
\usepackage{url}            
\usepackage{booktabs}       
\usepackage{amsfonts}       
\usepackage{nicefrac}       
\usepackage{microtype}      
\usepackage{xcolor}         

\usepackage{comment}

\usepackage{amsmath}
\usepackage{amssymb} 
\usepackage{verbatim}
\usepackage{pifont} 
\usepackage{fontawesome} 
\usepackage[clock]{ifsym} 
\usepackage{upgreek} 
\usepackage{xspace} 
\usepackage{upquote}
\usepackage{bbding}

\usepackage{soul} 
\usepackage{transparent}

\usepackage[most]{tcolorbox} 
\usepackage{mdframed} 

\usepackage{adjustbox}

\usepackage{xparse}
\usepackage{scalerel} 
\usepackage{pgfplots}
\usepackage{pgfplotstable}
\usepgfplotslibrary{groupplots}
\usepgfplotslibrary{statistics}
\usetikzlibrary{matrix} 
\pgfplotsset{compat=newest}
\usetikzlibrary{pgfplots.fillbetween,patterns,patterns.meta} 
\usetikzlibrary{decorations.markings}
\usetikzlibrary{babel}
\usepackage{tikz}
\usepgfplotslibrary{patchplots} 
\usepackage{wheelchart} 
\usepackage{tikzviolinplots} 
\usepackage{xfp} 
\usetikzlibrary{tikzmark}
\usetikzlibrary{calc}
\usetikzlibrary{fit}
\usetikzlibrary{positioning}
\usepackage{caption}
\usepackage{subcaption}
\pgfplotsset{compat=1.18,
    /pgfplots/xbar legend/.style={
    /pgfplots/legend image code/.code={%
       \draw[##1,/tikz/.cd,yshift=-0.25em]
        (0cm,0cm) rectangle (3pt,0.8em);},
   },
   /pgfplots/ybar legend/.style={
    /pgfplots/legend image code/.code={%
       \draw[##1,/tikz/.cd,yshift=-0.25em]
        (0cm,0cm) rectangle (3pt,0.8em);},
   },
}

\usepackage{array}
\usepackage{multirow}
\usepackage{colortbl}
\usepackage{arydshln} 
\usepackage{tabularx} 
\usepackage[para,online,flushleft]{threeparttable} 
\usepackage{stackengine}
\usepackage{hhline} 
\usepackage{diagbox} 
\usepackage{booktabs}
\usepackage{nicematrix}
\usepackage{bm}
\usepackage{makecell}

\newcolumntype{C}[1]{>{\centering\arraybackslash}p{#1}}

\usepackage{wrapfig}

\usepackage{bm} 
\usepackage{amsfonts} 
\usepackage{mathtools}

\usepackage{enumitem} 

\usepackage[pagebackref, breaklinks=true, colorlinks, citecolor=citecolor, linkcolor=linkcolor, urlcolor=urlcolor, bookmarks=false]{hyperref}
\usepackage[nameinlink,capitalise]{cleveref}
\crefname{section}{\S\hspace{-0.0em}}{\S}

\usepackage[toc,page]{appendix}
\usepackage{titletoc}

\usepackage{xskak} 
\usepackage{chessfss}
\usepackage[LSBC5,OT1]{fontenc}

\definecolor{citecolor}{HTML}{0071BC}
\definecolor{linkcolor}{HTML}{9D5917}
\definecolor{urlcolor}{HTML}{9D5917}

\definecolor{tablehighlight}{HTML}{F2F2F2}
\definecolor{tablebordercolor}{HTML}{363636}
\definecolor{cellhighlight}{HTML}{E9E9E9}

\storechessboardstyle{classic}{
    boardfontencoding=LSBC5,
    pgfstyle=color,
    color=brown!88, 
    trimtocolor=black,
    backboard,
    trimtocolor=white,
    color=brown!33, 
    backboard,
    whitepiecemaskcolor=white,
    blackpiecemaskcolor=white,
    setfontcolors,
    trimtocolor=false}
\definecolor{chessback}{HTML}{99BB9D}
\definecolor{chessarrow}{HTML}{32864A}

\definecolor{crayonRed}{HTML}{EE4266}
\definecolor{crayonBlue}{HTML}{5C95FF}
\definecolor{crayonYellow}{HTML}{FFD23F}

\newcommand{\legendBox}[1]{\tikz{\fill[#1] (0,0) rectangle (0.25cm,0.25cm);}}

\definecolor{tagbackground}{HTML}{FCF1E3}
\definecolor{tagforeground}{HTML}{7A4304}
\definecolor{rqtagbackground}{HTML}{772E11}
\definecolor{tokentagbackground}{HTML}{E7E7E7}
\definecolor{errortagbackground}{HTML}{FDECEC}
\definecolor{errortagforeground}{HTML}{850A0A}

\definecolor{survey-question-background}{HTML}{FFFBF6}
\definecolor{strongly-disagree-color}{HTML}{E26986}
\definecolor{disagree-color}{HTML}{E89AA9}
\definecolor{neutral-color}{HTML}{E2E2E2}
\definecolor{agree-color}{HTML}{A3ACE6}
\definecolor{strongly-agree-color}{HTML}{768BE6}
\definecolor{pgnmentor-color}{HTML}{B2B2B2}
\definecolor{lichess-color}{HTML}{CCCCCC}
\definecolor{chesscom-color}{HTML}{E6E6E6}
\definecolor{yes-color}{HTML}{77B6AE}
\definecolor{no-color}{HTML}{E2E2E2}
\definecolor{attacking-color}{HTML}{F6574C}
\definecolor{positional-color}{HTML}{BA90F9}
\definecolor{solid-color}{HTML}{87B4FD}
\definecolor{creative-color}{HTML}{F7BC6E}
\definecolor{do-not-know-color}{HTML}{868686}
\definecolor{other-color}{HTML}{B68277}
\definecolor{survey-bar-color}{HTML}{772E11}

\definecolor{challenge-background}{HTML}{FEF1F1}

\definecolor{plotbackground}{HTML}{FAF7FB}
\definecolor{uniquegamebar}{HTML}{D4AA97}
\definecolor{nonuniquegamebar}{HTML}{772E11}
\definecolor{ssl-color}{HTML}{D4AA97}
\definecolor{sslrl-color}{HTML}{772E11}
\definecolor{bar-color}{HTML}{D4AA97}
\definecolor{delta-color}{HTML}{772E11}
\definecolor{sim-color}{HTML}{D4AA97}
\definecolor{pos-delta-color}{HTML}{3E9D3A}
\definecolor{neg-delta-color}{HTML}{EF4444}
\definecolor{momline}{HTML}{772E11}
\definecolor{modelsoupline}{HTML}{D29E66}
\definecolor{transcendenceline}{HTML}{1F65D6}
\definecolor{karvonenline}{HTML}{CBD77F}
\definecolor{mixedgmline}{HTML}{99BF31}
\definecolor{expertarea}{HTML}{729153}
\definecolor{randompartitionarea}{HTML}{DCC389}
\definecolor{legendbackground}{HTML}{FFFBF6}
\definecolor{trainingbar}{HTML}{D4AA97}
\definecolor{expertmark}{HTML}{C78F57}

\definecolor{fidescorecolor}{HTML}{F5C28E}

\definecolor{concentrationcolor}{HTML}{729153}
\definecolor{sharpnesscolor}{HTML}{BE7D37}
\definecolor{topkborder}{HTML}{8BB050}
\definecolor{topkfill}{HTML}{E8F2D2}
\definecolor{bellcolor}{HTML}{772E11}
\definecolor{colHd}{HTML}{F0F0F0}   
\definecolor{panSep}{HTML}{A0A0A0}  

\pgfplotsset{colormap={chess}{
    rgb255(0pt)=(119,46,17);
    rgb255(1000pt)=(255,242,204)
}}

\newcommand{\cmark}{\ding{51}}
\newcommand{\xmark}{\ding{55}}

\newcommand{\momLong}{\textsc{Mixture-of-Masters}\xspace}
\newcommand{\mom}{\textsc{MoM}\xspace}
\newcommand{\gm}{GM\xspace}
\newcommand{\gms}{GMs\xspace}

\newtcbox{\roundedtag}[1][]{
  on line,
  colback=tagbackground,    
  colframe=white,           
  coltext=tagforeground,    
  boxrule=0pt,              
  arc=3pt,                  
  outer arc=3pt,
  top=-2pt,
  bottom=-2pt,
  left=-1pt,
  right=-1pt,
  #1
}
\newtcbox{\errortag}[1][]{
  on line,
  colback=errortagbackground,    
  colframe=white,                
  coltext=errortagforeground,   
  boxrule=0pt,                   
  arc=3pt,                       
  outer arc=3pt,
  top=-2pt,
  bottom=-2pt,
  left=-1pt,
  right=-1pt,
  #1
}

\newcommand\encircle[2][]{
    \tikz[overlay]
    \node[
        inner sep=1.4pt,
        anchor=text,
        rectangle,
        rounded corners=1mm,
        #1
    ] {#2};
    \phantom{#2}
}
\newcommand\encirclesmooth[2][]{
    \tikz[overlay]
    \node[
        inner sep=3pt,
        anchor=text,
        rectangle,
        rounded corners=1mm,
        #1
    ] {#2};
    \phantom{#2}
}
\newcommand{\rqtag}[1]{\encircle[fill=rqtagbackground, text=white]{#1}}
\newcommand{\tokentag}[1]{\encirclesmooth[fill=tokentagbackground, text=black]{#1}}

\newcommand{\quotes}[1]{``#1''}

\newcommand{\percentscale}{100}
\newcommand{\barwidth}{5}
\newcommand{\barheight}{0.3cm}
\def\pcb#1{%
   {\color{survey-bar-color}\rule{\fpeval{#1/\percentscale*\barwidth} cm}{\barheight}} #1
}

\newenvironment{surveyquestion}{
  \begin{mdframed}[
      leftmargin=0cm,
      rightmargin=0cm,
      innerleftmargin=4pt,
      innerrightmargin=4pt,
      innertopmargin=4pt,
      innerbottommargin=4pt,
      linewidth=0.8pt,
      backgroundcolor=survey-question-background,
    ]
    \fontsize{9pt}{9pt}\selectfont
}{
  \end{mdframed}\vspace{-0.3\baselineskip}
}

\newenvironment{challenge}{
  \begin{mdframed}[
      leftmargin=0cm,
      rightmargin=0cm,
      innerleftmargin=4pt,
      innerrightmargin=4pt,
      innertopmargin=4pt,
      innerbottommargin=4pt,
      linewidth=0.8pt,
      backgroundcolor=challenge-background,
    ]
}{
  \end{mdframed}\vspace{-0.3\baselineskip}
}

\title{Mixture of Masters: Sparse Chess Language Models with Player Routing}

\author{%
  Giacomo Frisoni, Lorenza Molfetta, Davide Freddi, Gianluca Moro \\
  Department of Computer Science and Engineering \\
  University of Bologna \\
  \texttt{\{giacomo.frisoni, lorenzo.molfetta, d.freddi, gianluca.moro\}@unibo.it} \\
}

\begin{document}

\maketitle

\begin{abstract}
Modern chess language models are dense transformers trained on millions of games played by thousands of high-rated individuals. However, these monolithic networks tend to collapse into mode-averaged behavior, where stylistic boundaries are blurred, and rare but effective strategies are suppressed. To counteract homogenization, we introduce \momLong (\mom), the first chess mixture-of-experts model with small-sized GPT experts emulating world-class grandmasters. For each move, a post-hoc learnable gating network selects the most appropriate persona to channel depending on the game state, allowing \mom to switch its style dynamically---e.g., Tal’s offensive vocation or Petrosian’s defensive solidity. When evaluated against Stockfish on unseen standard games, \mom outperforms both dense individual expert networks and popular GPT baselines trained on aggregated data, while ensuring generation variety, control, and interpretability.\\[1mm]
\raisebox{-0.25\height}{\hspace{0.05cm}\includegraphics[width=0.42cm]{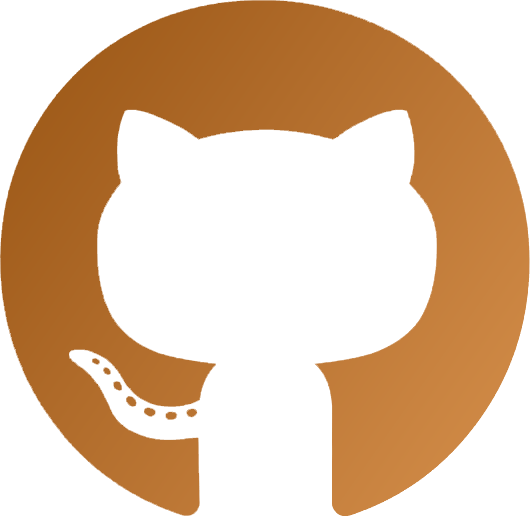}} \small \textbf{\mbox{Data, Code \& Model Weights:}} \href{https://anonymous.4open.science/r/mixture-of-masters}{anonymous.4open.science/r/mixture-of-masters}\\[1mm]
\raisebox{-0.25\height}{\hspace{0.05cm}\includegraphics[width=0.42cm]{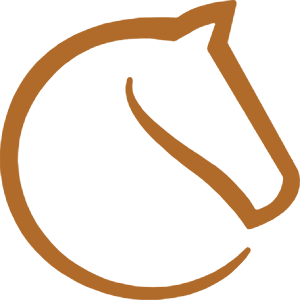}} \small \textbf{\mbox{Lichess Bot:}} \href{https://lichess.org/@/mixture-of-masters}{https://lichess.org/@/mixture-of-masters}
\end{abstract}

\section{Introduction}
\label{sec:introduction}

Originating nearly 1,500 years ago, chess ranks among the oldest and most thoroughly studied board games.
With a game-tree complexity of ${\sim}10^{120}$ (Shannon number)---far exceeding the number of estimated atoms in the observable universe---it demands strategic planning and creative thinking.
AI surpassed human chess capability roughly two decades ago, beginning with IBM's Deep Blue defeating world champion Garry Kasparov in 1997 using specialized hardware and tree search algorithms~\cite{DBLP:journals/ai/CampbellHH02}.
DeepMind's AlphaZero revolutionized the field in 2017 by removing human input through reinforcement learning (RL) and self-play~\cite{DBLP:journals/corr/abs-1712-01815}, inspiring contemporary engines such as Stockfish and Leela Chess Zero to adopt neural-network evaluation~\cite{DBLP:journals/corr/abs-2209-01506,DBLP:journals/entropy/MaharajPT22}.
The latest paradigm shift reframes chess as a language modeling problem, with transformer-based models learning rules and patterns from game transcripts in algebraic notation without explicit search mechanisms~\cite{DBLP:journals/corr/abs-2403-15498,DBLP:conf/aaai/ToshniwalWLG22}.

Chess history teaches us there is no single optimal way to play---champions with contrasting styles (e.g., positional, tactical, defensive) have all succeeded at the highest level~\cite{kasparov2003predecessors}.
In particular, creativity is a hallmark of chess excellence, embodying the ability to find unexpected, unconventional, yet valid moves that defy standard patterns.
However, contemporary chess language models are led by monolithic architectures that struggle with creative play.
A dense model, trained to minimize error across billions of moves from millions of players, might hesitate to choose rare or eccentric lines, preferring safe options that conform to dataset statistics.
This carries the risk of strategic conservatism and stylistic flattening: the unique traits of individual players may get diluted into a generic behavior.
Many chess professionals have warned against a possible homogenization of play with the widespread adoption of AI~\cite{alimpic2024impact,DBLP:conf/acg/BarrishKM23}.
We further substantiate these concerns through a survey of expert student and faculty players from 18 universities in 10 countries and 3 continents, detailed in \cref{app:survey}.
If everyone studies the same moves recommended by the engines, the players may adopt similar strategies and openings, reducing the diversity of ideas in the games.
This concern mirrors the trends observed in text generation, where studies indicate that the use of large language models (LLMs) results in a decline in expressive diversity, with writing styles converging toward dominant expressions while less common traits are suppressed~\cite{DBLP:conf/iclr/Padmakumar024,DBLP:journals/corr/abs-2502-11266}.

Recent developments suggest that sparse and modular mixture-of-experts (MoE) models may hold promise in computer chess~\cite{DBLP:journals/corr/abs-2401-16852}.
From 2021 to 2025, MoE architectures have undergone significant evolution, progressively redefining the notion of \quotes{experts}---shifting from feed-forward layers~\cite{DBLP:journals/jmlr/FedusZS22,DBLP:journals/corr/abs-2401-04088,DBLP:conf/iclr/LepikhinLXCFHKS21} to adapters~\cite{DBLP:conf/icml/MuqeethLLR24,DBLP:conf/emnlp/WuZH024} and full-model branches~\cite{simonds-etal-2024-modem,DBLP:conf/emnlp/ZhangBBCFFKSSDGL25}.
As the complexity and expressiveness of experts have increased, a natural question arises: \textit{Can we envision a persona-based MoE for chess?}

Building on this reasoning, we introduce \momLong (\mom), the first chess MoE for next move prediction\footnote[2]{We use the term \textit{\quotes{move}} to refer to an individual action by either White or Black (i.e., a semi-move or ply). We use the term \textit{\quotes{turn}} for a complete move cycle, commonly known as a full move.} with experts emulating world-class grandmasters (\gms).
We train multiple small-scale GPT models independently, each on the games of a specific \gm, preserving their distinctive styles without cross-contamination.
These specialized models are then combined into a unified sparse language model following a \quotes{wisdom of the crowd} paradigm.

\mom acts as a coalition of renowned players, each contributing their situational insight to the evolving board.
This approach is attractive for two main reasons.
First, \mom can prevent collapse toward the majority style and better handle out-of-distribution positions.
Second, it provides greater interpretability, enabling analysts to trace decisions back to identifiable chess personas.

\section{Related work}
\label{sec:related_work}

\mom is a human-centric MoE architecture for autoregressive chess language modeling~\cite{DBLP:journals/corr/abs-2403-15498,DBLP:conf/nips/ZhangZSKETKM24}, combining independently tuned \gm-specific expert models.
This design is orthogonal to previous chess MoEs~\cite{DBLP:journals/corr/abs-2401-16852}, which partition expertise by game phase and select moves via tree search.
Our central question is whether persona-based specialization improves playing strength over dense and sparse baselines while preserving stylistic diversity.
Behavioral stylometry offers tools to probe for player fingerprints, but has so far been applied successfully to large amateur populations only~\cite{DBLP:conf/nips/McIlroy-YoungWS21,DBLP:journals/corr/abs-2502-14998}.
Our experts, instead, refer to a small cohort of elite players whose styles largely overlap.
We tackle this challenge by moving beyond move-emulation scores alone, localizing persona-specific separation in expert activation space and testing whether \mom preserves that diversity through sharp, sparse routing decisions---a more robust and reliable diagnostic than move emulation alone.
Extended discussion appears in \cref{app:related_work}.

\section{Method}
\label{sec:method}

\subsection{\mom}

\begin{figure}[!htb]
    \centering
    \includegraphics[width=\linewidth]{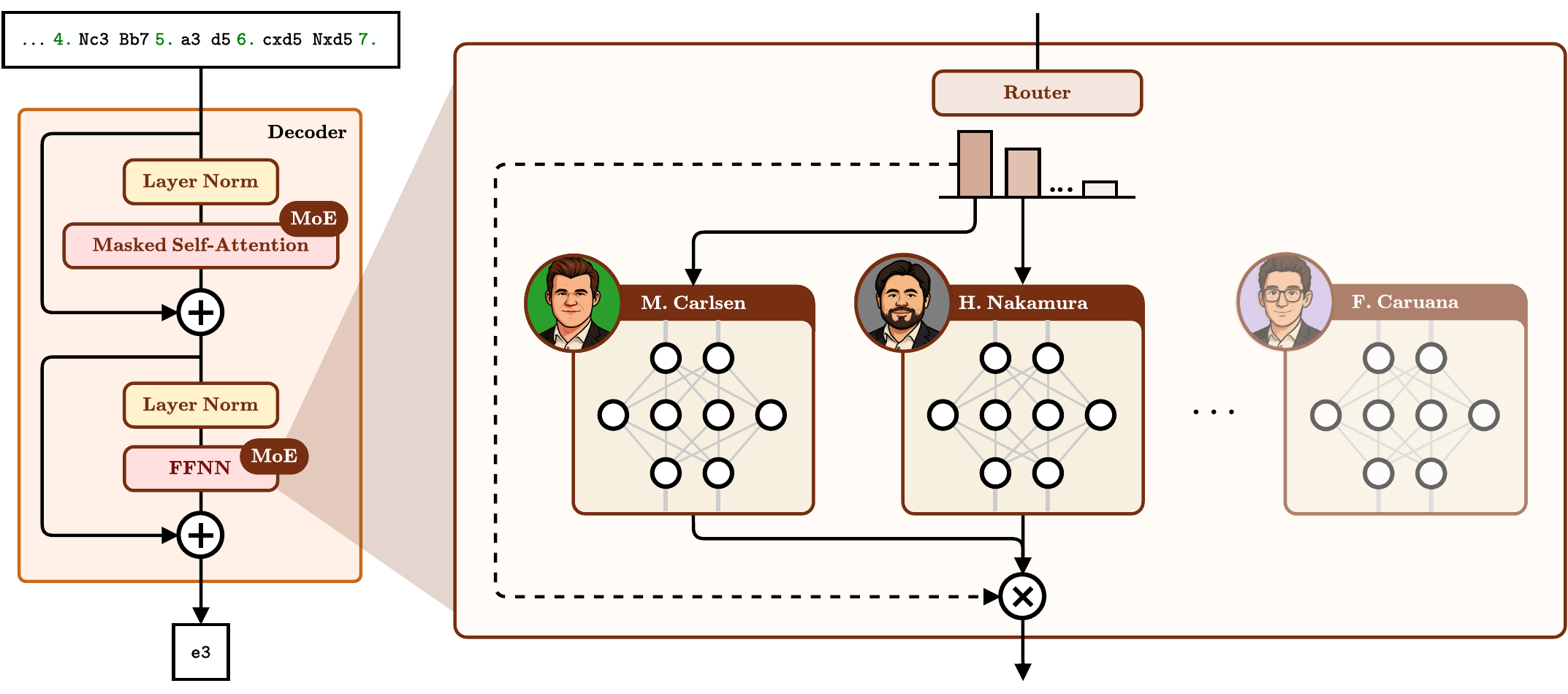}
    \caption{\textbf{Illustration of \textsc{\mom}.} First, multiple decoder-only chess language models are trained to emulate the game decisions of specific grandmasters. Then, their layers are combined into a sparse language model by alternating uniform weight merging and top-$k$ routing for next move prediction.}
    \label{fig:mom}
\end{figure}

Inspired by~\cite{DBLP:journals/corr/abs-2208-03306,DBLP:conf/emnlp/ZhangBBCFFKSSDGL25}, our sparse \mom model---depicted in Figure~\ref{fig:mom}---is constructed in three stages.
\begin{enumerate}[noitemsep,topsep=0pt]
    \item \textbf{Branch.} We create $P$ replicas $\mathcal{E} =(\varepsilon_{\phi_1},\ldots,\varepsilon_{\phi_P})$ of a $\phi$-parameterized seed model---a dense, decoder-only transformer pretrained on chess language modeling.
    \item \textbf{Train.} Each model copy $\varepsilon_{\phi_p}$ undergoes independent fine-tuning on moves made by a target \gm $p$, playing as either White or Black. The resulting models are referred to as \emph{experts}.
    \item \textbf{Stitch.} \mom is assembled from experts using a hybrid approach. At each layer, we either apply a weight-merging algorithm or implement a router to gate access to the original weights. Training is confined to the merged-weight and routing layers.
\end{enumerate}
In line with standard practice~\cite{DBLP:journals/corr/abs-2403-15498,DBLP:journals/corr/abs-2008-04057,DBLP:conf/nips/ZhangZSKETKM24}, \mom operates on games in Portable Game Notation (PGN) using a 32-character input-output vocabulary (see \cref{app:implementation_hw}), which ensures broad compatibility with existing seed models.

\subsubsection{Grandmaster experts}
\label{subsec:experts}

Each expert, $\varepsilon_{\phi_p}$, is derived by fine-tuning a copy of the seed model using self-supervised learning (SSL) to autoregressively predict the moves of its target player, $p$.
The objective is to refine the model's move distribution toward the stylistic tendencies of $p$.
To this end, the cross-entropy loss is computed exclusively on $p$'s tokens, as including opponent moves would introduce extraneous stylistic signals that dilute or contradict $p$'s profile.
Formally, let $\mathcal{D}_p=\left\{\left(s,m\right)\right\}$ denote the state--move pairs from the expert dataset, where $m$ is executed by $p$ starting from board state $s$.
Our \emph{player-side loss} is defined as: $\mathcal{L}_{\text{SSL}} = - \sum_{\left(s,m\right) \in \mathcal{D}_p} \log \varepsilon_{\phi_p}\left(m|s\right)$.
To ensure that stylometric evaluation is not confounded by unrelated incentives, we restrict the main paper to the described SSL; \cref{app:rl} investigates the effects of a supplementary reinforcement learning phase with legality rewards.

\subsubsection{Stitching}
\label{subsec:stitching}

\mom employs a hybrid parameter composition strategy: $\Phi_{\mathrm{MoM}} = \Phi_{\mathrm{gated}} \cup \Phi_{\mathrm{shared}}$.

\paragraph{Routing}

$\Phi_{\mathrm{gated}}$ comprises expert-specific layers subjected to dynamic routing.
Within each expert's decoder blocks, we isolate and parallelize all learned linear transformations: the $Q$-$K$-$V$ and output projection layers of masked self-attention, and the up- and down-projection layers of the feed-forward network.
Routing the full set of linear transformations in each block---rather than a subset---is consistent with recent expert upcycling work~\cite{DBLP:conf/nips/ZhangGG0CVFBRUL24}.
To regulate the information flow, a learnable linear gating network $\mathcal{G}_{\phi}$ is inserted prior to each parallelized module, mapping the current board state $s$ to a probability distribution over player experts: $\smash{P(p|s) = \text{softmax}(\mathcal{G}_{\phi}(s))}$.
During inference, only the top-$k$ experts with the highest routing probabilities are activated, aggregating their contributions via weighted sum pooling: $\smash{\sum_{p \in \text{top-}k(P(p|s))} P(p|s) \cdot \varepsilon_{\phi_p}(s)}$.
Training this router requires addressing two coupled challenges: making \gm selection differentiable, and preventing the collapse onto a single expert.
To enable end-to-end training through discrete selection, we replace hard top-$k$ with a Gumbel-Softmax relaxation~\cite{DBLP:conf/iclr/JangGP17}, which approximates discrete sampling via a continuous distribution parameterized by a temperature $\tau$.
Annealing $\tau$ from a high initial value to near zero over the course of training transitions the router from soft, exploratory blending toward sharp, near-discrete routing.
We additionally impose the auxiliary load balancing loss of~\cite{DBLP:journals/jmlr/FedusZS22}, which penalizes routing distributions that concentrate probability mass on a small subset of experts.

\paragraph{Merging}

All parameters not subject to gating are merged across experts into a shared backbone $\Phi_{\text{shared}}$ via uniform weight averaging~\cite{DBLP:conf/icml/WortsmanIGRLMNF22}.
Uniform averaging assigns equal importance to every expert, ensuring that the stylistic contribution of each \gm is preserved symmetrically.
More sophisticated consolidation strategies---including Fisher-weighted merging~\cite{DBLP:conf/nips/MatenaR22}, Task Arithmetic~\cite{DBLP:conf/iclr/IlharcoRWSHF23}, and KnOTS~\cite{DBLP:conf/iclr/StoicaRECH25}---are evaluated in \cref{app:merging}.
Although they can comparably high playing strength, they introduce weight disparities across experts, contradicting the democratic principle that motivates \mom: every \gm should contribute equally to the shared backbone, with selective activation delegated entirely to the router.

\subsection{Behavioral stylometry}
\label{subsec:behavioral_stylometry_metric}
Chess players exhibit stylistic preferences that surface as recurrent choices over openings, tactical motifs, risk, simplification, and conversion of favorable positions.
At the \gm level, these signatures are difficult to isolate because all players command a broad strategic repertoire and often converge toward high-precision moves.
We treat player identity as a source of systematic variation in sequential decision traces, and evaluate whether player-specialized models expose this variation through likelihoods, activation distributions, and routing patterns on held-out games.
Appendix~\ref{app:behavioral_stylometry} provides further details and baselines of previous work in chess stylometry analysis.

\paragraph{Distributional activation specialization}
We first identify which layers are most affected by player-specific fine-tuning.
Because all experts share the same initialization and architecture, their corresponding parameters and hidden states can be compared directly to uncover the mechanical differences that arise among structurally aligned models whose hidden representations are activated in player-specific ways.
More formally, for a held-out state--move pair $\smash{(s,m)}$, let $\smash{h^{(\ell)}_{p}(s)}$ denote the hidden state produced by expert $\varepsilon_{\phi_p}$ after the $\ell$-th decoder block parametrized by $\smash{\phi_p^{(\ell)}}$.
For every pair of experts, we compute and compare the Euclidean distance between their layer-wise weights $\smash{\Delta_\mathcal{W}=\Vert \phi_p^{(\ell)} - \phi_q^{(\ell)} \Vert_2}$ and their hidden states $\smash{\Delta_\mathcal{H}=\Vert h^{(\ell)}_{p}(s) - h^{(\ell)}_{q}(s) \Vert_2}$ evaluated on the same board positions $s$.
Comparing the plasticity of these metrics on own-master and other-master games allows us to disentangle parameter proximity from functional specialization.
In this setting, style emerges as a property of models that occupy a similar region of the parameter space yet produce distinct activation patterns when conditioned on their own master's data.

\paragraph{Likelihood and routing diagnostics}
We complement the activation analysis with output- and routing-level diagnostics.
At the output level, we compare each expert's negative log-likelihood (NLL) on held-out games from its own master with the NLL assigned to games from the remaining masters.
A positive own-master confidence gap indicates that the expert assigns a higher probability to the decisions of the player on whose games it was trained.
At the routing level, we analyze the expert probabilities induced by master-conditioned inputs in the stitched \mom model.
These probabilities provide an explicit trace of which player-specialized branches are recruited for each prediction.
For each routed module $\ell$, we retain $\smash{P^{(\ell)}(p|s)}$, the top activated experts, the top-$k$ probability mass, and the entropy of the routing distribution.
These quantities characterize how concentrated the router is on its selected experts and how sharply those experts separate from the remaining branches.
High top-$k$ probability mass and a large gap between selected and unselected experts indicate decisive expert composition.
Finally, comparison with a random-partition \mom control tests whether persona-aligned specialization yields sharper and more interpretable routing than generic sparse partitioning.
\section{Experiments}
\label{sec:experiments}


\subsection{Experimental setup}
\label{subsec:experimental_setup}

\paragraph{\faUsers\; Experts}

We anchor our research on $P{=}10$ \gms, selected for their high level and coverage in chess databases: \ding{182}~V. Anand, \ding{183}~L. Aronian, \ding{184}~M. Carlsen, \ding{185}~F. Caruana, \ding{186}~A. Firouzja, \ding{187}~A. Giri, \ding{188}~H. Nakamura, \ding{189}~I. Nepomniachtchi, \ding{190}~W. So, \ding{191}~M. Vachier-Lagrave.
To maximize the ensemble's efficacy and maintain a tractable scope for our experiments, we stitch only the five experts with superior playing strength.

\paragraph{\faDatabase\; Data}
For each \gm, we collect game records from PGNMentor, Chess.com, and Lichess and apply a curated filtering pipeline covering variant exclusion, time-control filtering, PGN normalization, deduplication, and color balancing (full details in \cref{app:data}).
The resulting per-\gm collections range from ${\sim}1{,}200$ to ${\sim}11{,}000$ games, spanning 1984--2025 with an average Elo of 2,817 and 91 moves per game.
Each collection is split 80:20 into training and held-out test sets via stratified sampling over color and game outcome.
The \emph{training data} serves three distinct roles.
(i) Expert creation: each expert $\varepsilon_{\phi_p}$ is trained on the full training split of its target \gm.
(ii) Router and merging calibration: following~\cite{DBLP:conf/emnlp/ZhangBBCFFKSSDGL25}, the stitching components are trained for 2M tokens on a 50:50 mixture of seed model pretraining games and a color-balanced, equal-size sample from the training split of each \gm.
(iii) Stylometry setup: training-dependent behavioral stylometry tools are fitted on a controlled subset of 1,000 color-balanced games per \gm drawn from the training split.
The held-out \emph{test data} is reserved exclusively for evaluating stylometry reliability on gold player-attributed data before applying the validated metrics to analyze expert- and \mom-generated moves.

\paragraph{\faGears\; Seed models}
Different seeds expose different levels of \quotes{plasticity.}
We examine four 50M-parameter nanoGPT models~\cite{karpathy2022} trained from scratch on Lichess records.
Three are drawn from the Transcendence collection~\cite{DBLP:conf/nips/ZhangZSKETKM24}: models $T_t$ trained on games by players rated below a Glicko-2 threshold $t$; our experiments involve $T_{1{,}000}$, $T_{1{,}300}$, and $T_{1{,}500}$.
The fourth is the largest model of Karvonen~\cite{DBLP:journals/corr/abs-2403-15498}, trained without rating restrictions.
All models have a maximum context length of 1,023 tokens, which accommodates $\sim$92 turns (184 moves).

\paragraph{\faTachometer\; Playing strength evaluation}
\roundedtag{Stockfish battle}
The model plays 100 games against Stockfish 16.1 at a specified skill level.
We constrain Stockfish to evaluate up to 100K nodes per move without a time cap; this operational mode reduces computational cost while eliminating inconsistencies arising from hardware and load variations~\cite{DBLP:journals/corr/abs-2403-15498}.
The game proceeds turn-by-turn: after each move by the model, the updated board state is passed to Stockfish, and vice versa.
Model moves are selected via greedy decoding.
The model operates under a strict no-retry policy; a single illegal move results in immediate forfeiture.
Games are capped at 90 turns, consistent with~\cite{DBLP:journals/corr/abs-2403-15498}, and outcomes beyond this horizon are determined by the centipawn evaluation of the final position.\footnote{$\text{Win\%} = 100 / (1 + \exp(-0.00368208 \times \text{centipawns}))$ from \url{https://lichess.org/page/accuracy}. The outcome is $>50\Rightarrow\text{win}$, $=50\Rightarrow\text{draw}$, $<50\Rightarrow\text{lose}$.}
In each match-up, the model and Stockfish swap seats to ensure fair White and Black exposure.
Stockfish's move selection is inherently stochastic: a randomized score perturbation is applied across top candidate moves, with the magnitude increasing as skill level decreases.
From the resulting outcomes, we compute three metrics: (i) \textit{Win Rate}, \% of games won, (ii) \textit{Draw Rate}, \% of games drawn, (iii) \textit{FIDEScore}, a novel aggregate scalar that awards 1 point for a win, 0.5 for a draw, and 0 for a loss---mirroring the scoring system adopted in official FIDE-rated competitions.\footnote{As codified in Article~11 of the FIDE Laws of Chess, this 1--$\tfrac{1}{2}$--0 system applies universally to all FIDE-rated events, from local Swiss tournaments to the World Championship; \url{https://handbook.fide.com/chapter/E012023}.}
\roundedtag{Rating}
The model plays a round-robin tournament against Stockfish lv.0--5, contributing 100 games per opponent for a total of 600 games.
Ratings are estimated via Glicko-2~\cite{glickman2012example}, which produces a point estimate $R$ of playing strength on an Elo-anchored scale and a rating deviation $RD$ quantifying residual uncertainty.
The model enters the tournament as an unrated player, initialized at the system defaults ($r{=}1{,}500$, $RD{=}350$), while Stockfish opponents are seeded at their officially calibrated Elo estimates.
We report the 95\% confidence interval $[R-2{\cdot}RD, R+2{\cdot}RD]$~\cite{DBLP:conf/nips/ZhangZSKETKM24}.
A comprehensive treatment of rating systems and practices across the literature are provided in \cref{app:play_strength_estimation}.
\roundedtag{Legality}
\% of games against Stockfish completed without the model generating an illegal move.

\paragraph{\faTachometer\; Stylometry evaluation}

\roundedtag{Emulation accuracy}
\% of positions where an expert's top-1 move prediction matches the move played by a \gm in held-out games.
\roundedtag{NLL}
Average NLL of an expert on held-out \gm games; lower values indicate higher confidence. We also report NLL advantage, computed as other-master NLL minus own-master NLL, where higher positive values indicate stronger player-specific confidence.
\roundedtag{Activation displacement}
For every pair of dense experts, we compare the static layer-wise parameter distance $\smash{\Delta_{\mathcal{W}}}$ with the corresponding activation distance $\smash{\Delta_{\mathcal{H}}}$ on held-out positions.
For each layer and game phase, we average $\smash{\Delta_{\mathcal{H}}}$ separately over \emph{own} inputs, whose games come from one of the two masters in the expert pair, and \emph{other} inputs, whose games come from the remaining masters.
The displacement score of \emph{own}/\emph{other} activation ratio,
$\smash{\Delta_{\mathcal{H},\mathrm{own}}^{(\ell)} /
\Delta_{\mathcal{H},\mathrm{other}}^{(\ell)}}$,
is averaged over expert pairs.
Values above one indicate that expert pairs become more functionally separated on their own reference master's games.
\roundedtag{MoE routing}
For \mom and the MoE controls, we record the full softmax routing distribution at every routed sublayer for master-side move tokens.
Each token contributes its top expert, unordered top-$k$ expert set, top probability, normalized entropy, and full expert-probability vector.
We summarize these records with routing concentration---the fraction of passes in which the top-$k$ experts jointly receive at least a fixed amount of probability mass---and routing sharpness, measured by the probability gap between the second- and third-ranked experts.
These metrics test whether \mom makes decisive sparse expert selections, and whether persona-aligned experts yield sharper routing than random expert partitions.

\paragraph{\faExchange\; Baselines}

All comparisons are conducted within a controlled family of models sharing the same base architecture, vocabulary, and evaluation protocol; differences in outcome are thus attributable solely to training strategy and composition design.
Engines outside this family---such as those relying on FEN representations, action-value supervision, or search---are not comparable on equal terms~\cite{DBLP:conf/nips/RuossDMGLCRLVG24}.
We consider two categories of baselines.
\roundedtag{Single-model}
(i) the \textit{best seed model}, which serves as the pre-specialization reference and quantifies the contribution of \gm-specific fine-tuning; (ii) a \textit{mixed-\gm model}, obtained by fine-tuning the seed on the pooled training data of all \gms.
\roundedtag{Composite}
(iii) \textit{model soup}~\cite{DBLP:conf/icml/WortsmanIGRLMNF22}, a dense model derived by uniformly averaging all expert and seed weights.
(iv) \textit{random-partition MoM},  where experts are fine-tuned on randomly shuffled, disjoint subsets of the same pooled GM data rather than player-specific collections---isolating the contribution of persona-aligned specialization over generic data partitioning; to account for variance due to random assignment, we instantiate three independent partitions.

Implementation details and hardware setup are provided in \cref{app:implementation_hw}.


\subsection{Results}
\label{subsec:results}

We organize our experimental results around a series of research questions (RQ1-RQ5) that interrogate different dimensions of our approach, from the impact of design decisions to \mom capabilities.

\paragraph{\rqtag{RQ1} Which seed model is the optimal foundation for training \gm experts?}

\begin{wrapfigure}{r}{0.32\textwidth}
    \centering
    \vspace{-4mm}
    \begin{tikzpicture}
        \begin{axis}[
            width=\linewidth, height=4.5cm,
            grid=major,
            xtick={1,2,3,4},
            xticklabels={$T_{1,000}$,$T_{1,300}$,$T_{1,500}$,Karv.},
            xticklabel style={
                xshift={ifthenelse(\ticknum==3,-2mm,-1.5mm)},
                yshift={ifthenelse(\ticknum==3,2mm,1.5mm)},
            },
            x dir=reverse,
            enlarge x limits=0.1,
            ylabel={Experts},
            ytick={1,2,3,4,5,6,7,8,9,10},
            yticklabels={\ding{182},\ding{183},\ding{184},\ding{185},\ding{186},\ding{187},\ding{188},\ding{189},\ding{190},\ding{191}},
            ylabel style={rotate=30},
            zlabel={Avg\\FIDEScore},
            xticklabel style={font=\fontsize{7}{7}\selectfont},
            yticklabel style={font=\fontsize{7}{7}\selectfont},
            zticklabel style={font=\fontsize{7}{7}\selectfont},
            xlabel style={font=\fontsize{7}{7}\selectfont},
            ylabel style={font=\fontsize{7}{7}\selectfont},
            zlabel style={font=\fontsize{7}{7}\selectfont, align=center, xshift=1mm},
            view={50}{60},
            colormap name=chess,
        ]
        \addplot3[
            surf,
            mesh/rows=10,
            mesh/cols=4,
            shader=flat,
            draw=black!80,
            mesh/ordering=x varies
        ] table {
            x      y      z
            1      1      0.29
            2      1      0.42
            3      1      0.46
            4      1      0.59
            1      2      0.25
            2      2      0.4
            3      2      0.47
            4      2      0.6
            1      3      0.28
            2      3      0.43
            3      3      0.48
            4      3      0.63
            1      4      0.26
            2      4      0.41
            3      4      0.49
            4      4      0.63
            1      5      0.27
            2      5      0.42
            3      5      0.46
            4      5      0.59
            1      6      0.28
            2      6      0.42
            3      6      0.48
            4      6      0.64
            1      7      0.26
            2      7      0.38
            3      7      0.50
            4      7      0.64
            1      8      0.26
            2      8      0.42
            3      8      0.49
            4      8      0.65
            1      9      0.26
            2      9      0.43
            3      9      0.45
            4      9      0.66
            1      10     0.28
            2      10     0.4
            3      10     0.42
            4      10     0.58
        };
        \addplot3[red, thick, dashed] coordinates {
            (1, 10, 0.4) (1, 1, 0.4) (4, 1, 0.4)
        };
        \addplot3[green, thick, dashed] coordinates {
            (1, 10, 0.6) (1, 1, 0.6) (4, 1, 0.6)
        };
        \end{axis}
    \end{tikzpicture}
    \caption{\textbf{Effect of seed model on expert FIDEScore}. Stockfish lv.0, pooled over 10 runs.}
    \label{fig:seed_expert_fidescore}
    \vspace{-5mm}
\end{wrapfigure}

While $T_{t}$ models show incremental gains as the rating threshold $t$ increases, the Karvonen seed yields a significant performance jump, consistently pushing experts toward the $0.6$ FIDEScore threshold (Figure~\ref{fig:seed_expert_fidescore}).
This suggests that the Karvonen model's representation space is more amenable to stylistic shifting without triggering catastrophic forgetting of the underlying game manifold.
Mechanistically, this superiority likely stems from a more expressive internal world model.
Whereas the $T_{t}$ series is exposed to the mode-averaged noise of lower-rated amateur play, the Karvonen seed's training on general Lichess records provides a more diverse strategic baseline.
Consequently, we adopt the Karvonen model as the backbone for \mom.

\paragraph{\rqtag{RQ2} Which \gm experts achieve the highest playing strength?}

Every expert surpasses the seed model (Figure~\ref{tab:expert_results}), confirming that
persona-specific fine-tuning consistently improves performance against Stockfish.
FIDEScores span a 7.8-point range--from 57.8 (\ding{191}) to 65.6 (\ding{190})--a gap exceeding the within-expert standard deviation for eight out of ten models, indicating genuine performance differences rather than sampling noise; the bottom of the ranking coincides with the smallest training sets, implicating data volume as a primary limiting factor.
Figure~\ref{fig:legality_ssl} reveals a native legality varying from $\sim$78\% to $\sim$87\%.
We select the five experts with jointly superior FIDEScore and native legality (\ding{184}, \ding{187}, \ding{188}, \ding{189}, \ding{190}) as the components of \mom.
Figure~\ref{fig:legality_cd} further quantifies the strength recoverable by eliminating illegal moves via constrained decoding~(CD) at inference time: FIDEScore improves by $+11$--$+15$ percentage points across all five experts.
Unless otherwise stated, all evaluations reported hereafter are conducted \emph{without} CD, so that all results reflect the native competence of the models.

\begin{table*}[!htb]
    \caption{\textbf{Playing strength of \gm experts.} Stockfish lv.0, pooled over 10 runs. Game metrics are Avg$\pm$Std (\%).}
    \centering
    \begin{adjustbox}{width=\linewidth}
    \begin{tabular}{l !{\vrule width 0.8pt} c !{\vrule width 0.8pt} cccccccccc}
        \hline
        \rowcolor{colHd}
        \textbf{Metric}
            & \textbf{Seed}
            & {\Large\ding{182}} & {\Large\ding{183}} & {\Large\ding{184}} & {\Large\ding{185}}
            & {\Large\ding{186}} & {\Large\ding{187}} & {\Large\ding{188}} & {\Large\ding{189}}
            & {\Large\ding{190}} & {\Large\ding{191}} \\
        \hline
        Draw Rate
            & 24.0$_{\pm2.4}$
            & 14.7$_{\pm3.0}$ & 14.6$_{\pm3.7}$ & 15.2$_{\pm3.7}$ & 14.4$_{\pm3.1}$
            & 15.8$_{\pm5.3}$ & 16.2$_{\pm3.1}$ & 16.1$_{\pm2.6}$ & 13.5$_{\pm4.3}$
            & 18.5$_{\pm4.9}$ & 16.3$_{\pm1.8}$ \\
        Win Rate
            & 42.1$_{\pm4.0}$
            & 52.0$_{\pm4.5}$ & 52.6$_{\pm4.2}$ & 55.0$_{\pm5.4}$ & 55.3$_{\pm4.1}$
            & 51.2$_{\pm6.5}$ & 55.6$_{\pm5.2}$ & 55.4$_{\pm3.1}$ & 58.5$_{\pm5.5}$
            & 56.4$_{\pm5.0}$ & 49.6$_{\pm4.1}$ \\
        FIDEScore
            & \cellcolor{fidescorecolor!5}54.1$_{\pm4.1}$
            & \cellcolor{fidescorecolor!24}59.4$_{\pm3.8}$ & \cellcolor{fidescorecolor!28}59.9$_{\pm4.6}$ & \cellcolor{fidescorecolor!54}62.6$_{\pm5.3}$ & \cellcolor{fidescorecolor!53}62.5$_{\pm4.6}$
            & \cellcolor{fidescorecolor!22}59.1$_{\pm5.5}$ & \cellcolor{fidescorecolor!68}63.7$_{\pm4.4}$ & \cellcolor{fidescorecolor!65}63.5$_{\pm4.0}$ & \cellcolor{fidescorecolor!90}65.3$_{\pm4.1}$
            & \cellcolor{fidescorecolor!95}65.6$_{\pm4.4}$ & \cellcolor{fidescorecolor!14}57.8$_{\pm4.1}$ \\
        \hline
    \end{tabular}
    \end{adjustbox}
    \label{tab:expert_results}
\end{table*}

\begin{figure*}[!h]
\centering
\colorlet{zoomcol}{orange!70!brown}
%
\subcaptionbox{%
    Legality distributions across \gm experts. Stockfish lv.0, pooled over 10 runs.%
    \label{fig:legality_ssl}}[0.4\textwidth]{%
\begin{tikzpicture}
\begin{axis}[
    width=\linewidth,
    height=3.4cm,
    grid=both,
    grid style={gray!30, dashed},
    axis background/.style={fill=plotbackground},
    ymin=65, ymax=100,
    boxplot/draw direction=y,
    ylabel=Legality (\%),
    yticklabel style={font=\fontsize{7}{7}\selectfont},
    ylabel style={font=\fontsize{9}{9}\selectfont},
    xlabel=Experts,
    xlabel style={font=\fontsize{9}{9}\selectfont},
    xtick={1,2,3,4,5,6,7,8,9,10},
    xticklabels={%
        \ding{182},\ding{183},\ding{184},\ding{185},\ding{186},
        \ding{187},\ding{188},\ding{189},\ding{190},\ding{191}},
    xticklabel style={font=\fontsize{9}{9}\selectfont},
    xmin=0.2, xmax=10.8,
]
\addplot+[boxplot prepared={
    lower whisker=70, lower quartile=73.3, median=78,
    upper quartile=81.8, upper whisker=85, box extend=0.35},
    draw=black, line width=0.5pt,
    fill=ssl-color, fill opacity=0.7, solid,
    boxplot/draw position=1] coordinates {};
\addplot+[boxplot prepared={
    lower whisker=72, lower quartile=76, median=79,
    upper quartile=81.5, upper whisker=86, box extend=0.35},
    draw=black, line width=0.5pt,
    fill=ssl-color, fill opacity=0.7, solid,
    boxplot/draw position=2] coordinates {};
\addplot+[boxplot prepared={
    lower whisker=71, lower quartile=80, median=83.5,
    upper quartile=84, upper whisker=88, box extend=0.35},
    draw=black, line width=0.5pt,
    fill=ssl-color, fill opacity=0.7, solid,
    boxplot/draw position=3] coordinates {};
\addplot+[boxplot prepared={
    lower whisker=76, lower quartile=81.3, median=84,
    upper quartile=85.5, upper whisker=88, box extend=0.35},
    draw=black, line width=0.5pt,
    fill=ssl-color, fill opacity=0.7, solid,
    boxplot/draw position=4] coordinates {};
\addplot+[boxplot prepared={
    lower whisker=74, lower quartile=78.25, median=79.5,
    upper quartile=83.5, upper whisker=88, box extend=0.35},
    draw=black, line width=0.5pt,
    fill=ssl-color, fill opacity=0.7, solid,
    boxplot/draw position=5] coordinates {};
\addplot+[boxplot prepared={
    lower whisker=78, lower quartile=82.5, median=84,
    upper quartile=85.8, upper whisker=87, box extend=0.35},
    draw=black, line width=0.5pt,
    fill=ssl-color, fill opacity=0.7, solid,
    boxplot/draw position=6] coordinates {};
\addplot+[boxplot prepared={
    lower whisker=79, lower quartile=82.3, median=85,
    upper quartile=88, upper whisker=89, box extend=0.35},
    draw=black, line width=0.5pt,
    fill=ssl-color, fill opacity=0.7, solid,
    boxplot/draw position=7] coordinates {};
\addplot+[boxplot prepared={
    lower whisker=75, lower quartile=84, median=86,
    upper quartile=89.5, upper whisker=91, box extend=0.35},
    draw=black, line width=0.5pt,
    fill=ssl-color, fill opacity=0.7, solid,
    boxplot/draw position=8] coordinates {};
\addplot+[boxplot prepared={
    lower whisker=80, lower quartile=83.3, median=85,
    upper quartile=88.5, upper whisker=93, box extend=0.35},
    draw=black, line width=0.5pt,
    fill=ssl-color, fill opacity=0.7, solid,
    boxplot/draw position=9] coordinates {};
\addplot+[boxplot prepared={
    lower whisker=75, lower quartile=78, median=79,
    upper quartile=80, upper whisker=86, box extend=0.35},
    draw=black, line width=0.5pt,
    fill=ssl-color, fill opacity=0.7, solid,
    boxplot/draw position=10] coordinates {};
\end{axis}
\end{tikzpicture}}%
%
\hfill
\raisebox{-0.9cm}{{\color{panSep}\rule{0.5pt}{3.8cm}}}%
\hfill
%
\subcaptionbox{%
    FIDEScore gains from constrained decoding (CD), i.e., 100\% legality. Top-5 experts. Stockfish lv.0.
    \label{fig:cd_fidescore}}[0.56\textwidth]{%
\resizebox{\linewidth}{!}{%
\begin{tikzpicture}[
    every node/.style={font=\small},
    inner sep=0pt
]

\node[inner sep=0pt] (maintable) {%
    \renewcommand{\arraystretch}{1.25}%
    \setlength{\tabcolsep}{8pt}%
    \begin{tabular}{lcc}
    \hline
    \rowcolor{colHd}
    \textbf{Exp.} & \textbf{\xmark\; CD} & \textbf{\cmark\; CD} \\
    \hline
    \ding{184} & 62.6 & \cellcolor{zoomcol!15}76.4 \\
    \ding{187} & 63.7 & \cellcolor{zoomcol!15}79.0 \\
    \ding{188} & 63.5 & \cellcolor{zoomcol!15}77.0 \\
    \ding{189} & 65.3 & \cellcolor{zoomcol!15}76.8 \\
    \ding{190} & 65.6 & \cellcolor{zoomcol!15}76.4 \\
    \hline
    \rowcolor{gray!8}
    Avg.  & 64.1 & \cellcolor{zoomcol!15}77.1 \\
    \hline
    \end{tabular}%
};

\draw[zoomcol, line width=1.1pt, rounded corners=2pt]
    ([xshift=-1.45cm, yshift=0cm]maintable.north east)
    rectangle
    ([xshift= 0cm, yshift=0cm]maintable.south east);

\coordinate (zoomTR) at
    ([xshift=0cm, yshift=0cm]maintable.north east);
\coordinate (zoomBR) at
    ([xshift=0cm, yshift=0cm]maintable.south east);

\node[draw=zoomcol, line width=1.2pt, rounded corners=5pt,
      inner sep=8pt, fill=zoomcol!4,
      anchor=west] (zoombox) at
      ([xshift=0.35cm]maintable.east) {%
    \begin{minipage}{5.5cm}
    \centering
    {\small\textcolor{zoomcol}{\textbf{Why does CD improve?}}}\\[4pt]
    \renewcommand{\arraystretch}{1.25}%
    \setlength{\tabcolsep}{5pt}%
    \begin{tabular}{lcc}
    \hline
    \rowcolor{colHd}
    \textbf{Exp.}
        & $\bm{\Delta}$\,\textbf{Win Rate}
        & $\bm{\Delta}$\,\textbf{Draw Rate} \\
    \hline
    \ding{184} & $+11.6\%$ & $-6.0\%$ \\
    \ding{187} & $+15.0\%$ & $-7.6\%$ \\
    \ding{188} & $+13.3\%$ & $-8.3\%$ \\
    \ding{189} & $+13.4\%$ & $-8.6\%$ \\
    \ding{190} & $+11.8\%$ & $-6.4\%$ \\
    \hline
    \end{tabular}
    \end{minipage}%
};

\draw[zoomcol, dashed, line width=0.9pt]
    (zoomTR) -- (zoombox.north west);
\draw[zoomcol, dashed, line width=0.9pt]
    (zoomBR) -- (zoombox.south west);

\end{tikzpicture}}}%
\caption{\textbf{Expert legality and its effect on playing strength.}}
\label{fig:legality_cd}
\end{figure*}

\paragraph{\rqtag{RQ3} Does \mom outperform single-model and composite baselines in playing strength?}

Prior to the main comparison, we ablate the number of active experts $k$ in \mom (Figure~\ref{fig:topk_ablation}): FIDEScore peaks at $k{=}2$ and degrades monotonically beyond it, indicating that pairwise persona routing is a right balance between specialization and ensemble dilution; we fix $k{=}2$ throughout.
\mom consistently achieves the highest FIDEScore at every difficulty level, establishing it as the most well-rounded model across the full evaluation range.
Against the Karvonen seed, \mom improves FIDEScore by $+15.0$ points at lv.0---a margin that exceeds the gain achieved by even the strongest individual expert (\ding{190}, $+11.5$), and that persists, albeit compressed in absolute terms, as Stockfish difficulty increases.
\mom further outperforms the expert soup baseline by up to $+3$ FIDEScore points, demonstrating that learned, input-dependent gating yields measurably better outcomes than full naive uniform parameter averaging over the same set of expert weights.
Critically, \mom also surpasses random-partition \mom baselines---models that share identical routing capacity, parameter count, and training budget, but whose experts are fine-tuned on randomly shuffled, player-agnostic data partitions rather than GM-specific collections.
Persona-aligned specialization, not ensemble diversity, drives the gain.
These findings are corroborated by Glicko-2 ratings; we caution that these ratings should be interpreted comparatively within our evaluation protocol rather than mapped directly onto online platform scales or ratings estimated by other publications under different tournament settings~\cite{DBLP:conf/nips/RuossDMGLCRLVG24}.

\begin{figure*}[!t]
    \centering
    \begin{subfigure}[t]{\linewidth}
    \centering
    \begin{tikzpicture}
        \fontsize{8}{8}\selectfont
        \node[draw=none, inner sep=1pt, align=center] {
        \begin{tabular}{llll}
            \makecell[tl]{\textbf{Composite:}}
                & \makecell[tl]{\raisebox{.5ex}{\tikz{\draw[momline, line width=1.2pt] (0,0) -- (0.3,0);}} \mom \textit{(Ours)} \\ \roundedtag{$1,557{\pm}12$}}
                & \makecell[tl]{\raisebox{.5ex}{\tikz{\draw[modelsoupline, line width=1pt] (0,0) -- (0.3,0);}} Expert soup~\cite{DBLP:conf/icml/WortsmanIGRLMNF22} \\ \roundedtag{$1,553{\pm}12$}}
                & \makecell[tl]{\raisebox{.2ex}{\tikz[opacity=0.6]{\draw[randompartitionarea, line width=1.2pt] (0,0) -- (0.3,0);\draw[randompartitionarea, line width=1.2pt] (0,-0.05) -- (0,0.05);\draw[randompartitionarea, line width=1.2pt] (0.3,-0.05) -- (0.3,0.05);}} Random-partitioned \mom \\ \roundedtag{Avg. $1,549{\pm}7$}} \\[4mm]
            \makecell[tl]{\textbf{Single-model:}}
                & \makecell[tl]{\raisebox{.2ex}{\tikz[opacity=0.6]{\draw[expertarea, line width=1.2pt] (0,0) -- (0.3,0);\draw[expertarea, line width=1.2pt] (0,-0.05) -- (0,0.05);\draw[expertarea, line width=1.2pt] (0.3,-0.05) -- (0.3,0.05);}} Individual experts \\ \roundedtag{Avg. $1,498{\pm}12$}}
                & \makecell[tl]{\raisebox{.5ex}{\tikz{\draw[mixedgmline, line width=1pt] (0,0) -- (0.3,0);}} Mixed-GM \\ \roundedtag{$1,540{\pm}12$}}
                & \makecell[tl]{\raisebox{.5ex}{\tikz{\draw[karvonenline, line width=1pt] (0,0) -- (0.3,0);}} Karvonen~\cite{DBLP:journals/corr/abs-2403-15498} \\ \roundedtag{$1,423{\pm}11$}} \\
        \end{tabular}
        };
    \end{tikzpicture}
    \end{subfigure}\\[1mm]
    \begin{minipage}[t]{.03\linewidth}
        \centering
        \raisebox{.7cm}{\rotatebox{90}{\scriptsize Avg FIDEScore}}
    \end{minipage}
    \hfill
    \begin{subfigure}[t]{.15\linewidth}
    \begin{tikzpicture}
    \centering
        \begin{axis}[
            ybar=3pt, 
            bar width=3pt,
            width=.7\linewidth, height=1.8cm,
            scale only axis,
            ymajorgrids=true,
            grid=both,
            grid style=dashed,
            axis background/.style={fill=plotbackground},
            enlarge x limits=0.02,
            enlarge y limits=0.1,
            xtick={0},
            ymin=50,
            every tick label/.append style={font=\fontsize{8}{8}\selectfont},
            xlabel style={font=\fontsize{8}{8}\selectfont},
            ylabel style={font=\fontsize{8}{8}\selectfont},
        ]
        \addplot [fill=momline, draw=none, error bars/.cd, y dir=both, y explicit] coordinates {
            (0,69.1)
        };\label{plot:mom}
        \addplot [fill=randompartitionarea, draw=none, error bars/.cd, y dir=both, y explicit] coordinates {
            (0,67.65) +- (0, 0.5)
        };\label{plot:random_partitioned}
        \addplot [fill=modelsoupline, draw=none, error bars/.cd, y dir=both, y explicit] coordinates {
            (0,65.4)
        };\label{plot:model_soup}
        \addplot [fill=expertarea, draw=none, error bars/.cd, y dir=both, y explicit] coordinates {
            (0,64.1) +- (0, 1.5)
        };\label{plot:individual_expert}
        \addplot [fill=mixedgmline, draw=none, error bars/.cd, y dir=both, y explicit] coordinates {
            (0,65.45)
        };
        \addplot [fill=karvonenline, draw=none, error bars/.cd, y dir=both, y explicit] coordinates {
            (0,54)
        };\label{plot:karvonen}
        \end{axis}
    \end{tikzpicture}
    \end{subfigure}
    \hfill
    \begin{subfigure}[t]{.15\linewidth}
    \begin{tikzpicture}
    \centering
        \begin{axis}[
            ybar=3pt, 
            bar width=3pt,
            width=.7\linewidth, height=1.8cm,
            scale only axis,
            ymajorgrids=true,
            grid=both,
            grid style=dashed,
            axis background/.style={fill=plotbackground},
            enlarge x limits=0.02,
            enlarge y limits=0.1,
            xtick={1},
            ymin=50,
            every tick label/.append style={font=\fontsize{8}{8}\selectfont},
            xlabel style={font=\fontsize{8}{8}\selectfont},
            ylabel style={font=\fontsize{8}{8}\selectfont},
        ]
        \addplot [fill=momline, draw=none, error bars/.cd, y dir=both, y explicit] coordinates {
            (1,59.3)
        };
        \addplot [fill=randompartitionarea, draw=none, error bars/.cd, y dir=both, y explicit] coordinates {
            (1,58.5) +- (0, 0.5)
        };
        \addplot [fill=modelsoupline, draw=none, error bars/.cd, y dir=both, y explicit] coordinates {
            (1,58.1)
        };
        \addplot [fill=expertarea, draw=none, error bars/.cd, y dir=both, y explicit] coordinates {
            (1,57.5) +- (0, 0.5)
        };
        \addplot [fill=mixedgmline, draw=none, error bars/.cd, y dir=both, y explicit] coordinates {
            (1,57.9)
        };
        \addplot [fill=karvonenline, draw=none, error bars/.cd, y dir=both, y explicit] coordinates {
            (1,50)
        };
        \end{axis}
    \end{tikzpicture}
    \end{subfigure}
    \hfill
    \begin{subfigure}[t]{.15\linewidth}
    \begin{tikzpicture}
    \centering
        \begin{axis}[
            ybar=3pt, 
            bar width=3pt,
            width=.7\linewidth, height=1.8cm,
            scale only axis,
            ymajorgrids=true,
            grid=both,
            grid style=dashed,
            axis background/.style={fill=plotbackground},
            enlarge x limits=0.02,
            enlarge y limits=0.1,
            xtick={2},
            ymin=40,
            every tick label/.append style={font=\fontsize{8}{8}\selectfont},
            xlabel style={font=\fontsize{8}{8}\selectfont},
            ylabel style={font=\fontsize{8}{8}\selectfont},
        ]
        \addplot [fill=momline, draw=none, error bars/.cd, y dir=both, y explicit] coordinates {
            (2,50.1)
        };
        \addplot [fill=randompartitionarea, draw=none, error bars/.cd, y dir=both, y explicit] coordinates {
            (2,49.8) +- (0, 0.25)
        };
        \addplot [fill=modelsoupline, draw=none, error bars/.cd, y dir=both, y explicit] coordinates {
            (2,49.6)
        };
        \addplot [fill=expertarea, draw=none, error bars/.cd, y dir=both, y explicit] coordinates {
            (2,48.25) +- (0, 0.25)
        };
        \addplot [fill=mixedgmline, draw=none, error bars/.cd, y dir=both, y explicit] coordinates {
            (2,48.6)
        };
        \addplot [fill=karvonenline, draw=none, error bars/.cd, y dir=both, y explicit] coordinates {
            (2,40.7)
        };
        \end{axis}
    \end{tikzpicture}
    \end{subfigure}
    \hfill
    \begin{subfigure}[t]{.15\linewidth}
    \begin{tikzpicture}
    \centering
        \begin{axis}[
            ybar=3pt, 
            bar width=3pt,
            width=.7\linewidth, height=1.8cm,
            scale only axis,
            ymajorgrids=true,
            grid=both,
            grid style=dashed,
            axis background/.style={fill=plotbackground},
            enlarge x limits=0.02,
            enlarge y limits=0.1,
            xtick={3},
            ymin=20,
            every tick label/.append style={font=\fontsize{8}{8}\selectfont},
            xlabel style={font=\fontsize{8}{8}\selectfont},
            ylabel style={font=\fontsize{8}{8}\selectfont},
        ]
        \addplot [fill=momline, draw=none, error bars/.cd, y dir=both, y explicit] coordinates {
            (3,34.4)
        };
        \addplot [fill=randompartitionarea, draw=none, error bars/.cd, y dir=both, y explicit] coordinates {
            (3,33.4) +- (0, 0.35)
        };
        \addplot [fill=modelsoupline, draw=none, error bars/.cd, y dir=both, y explicit] coordinates {
            (3,31.9)
        };
        \addplot [fill=expertarea, draw=none, error bars/.cd, y dir=both, y explicit] coordinates {
            (3,31.85) +- (0, 0.35)
        };
        \addplot [fill=mixedgmline, draw=none, error bars/.cd, y dir=both, y explicit] coordinates {
            (3,31.65)
        };
        \addplot [fill=karvonenline, draw=none, error bars/.cd, y dir=both, y explicit] coordinates {
            (3,25)
        };
        \end{axis}
    \end{tikzpicture}
    \end{subfigure}
    \hfill
    \begin{subfigure}[t]{.15\linewidth}
    \begin{tikzpicture}
    \centering
        \begin{axis}[
            ybar=3pt, 
            bar width=3pt,
            width=.7\linewidth, height=1.8cm,
            scale only axis,
            ymajorgrids=true,
            grid=both,
            grid style=dashed,
            axis background/.style={fill=plotbackground},
            enlarge x limits=0.02,
            enlarge y limits=0.1,
            xtick={4},
            ymin=15,
            every tick label/.append style={font=\fontsize{8}{8}\selectfont},
            xlabel style={font=\fontsize{8}{8}\selectfont},
            ylabel style={font=\fontsize{8}{8}\selectfont},
        ]
        \addplot [fill=momline, draw=none, error bars/.cd, y dir=both, y explicit] coordinates {
            (4,24)
        };
        \addplot [fill=randompartitionarea, draw=none, error bars/.cd, y dir=both, y explicit] coordinates {
            (4,22.5) +- (0, 0.5)
        };
        \addplot [fill=modelsoupline, draw=none, error bars/.cd, y dir=both, y explicit] coordinates {
            (4,21.25)
        };
        \addplot [fill=expertarea, draw=none, error bars/.cd, y dir=both, y explicit] coordinates {
            (4,20.5) +- (0, 0.5)
        };
        \addplot [fill=mixedgmline, draw=none, error bars/.cd, y dir=both, y explicit] coordinates {
            (4,21.1)
        };
        \addplot [fill=karvonenline, draw=none, error bars/.cd, y dir=both, y explicit] coordinates {
            (4,18)
        };
        \end{axis}
    \end{tikzpicture}
    \end{subfigure}
    \hfill
    \begin{subfigure}[t]{.15\linewidth}
    \begin{tikzpicture}
    \centering
        \begin{axis}[
            ybar=3pt, 
            bar width=3pt,
            width=.7\linewidth, height=1.8cm,
            scale only axis,
            ymajorgrids=true,
            grid=both,
            grid style=dashed,
            axis background/.style={fill=plotbackground},
            enlarge x limits=0.02,
            enlarge y limits=0.1,
            xtick={5},
            ymin=10,
            every tick label/.append style={font=\fontsize{8}{8}\selectfont},
            xlabel style={font=\fontsize{8}{8}\selectfont},
            ylabel style={font=\fontsize{8}{8}\selectfont},
        ]
        \addplot [fill=momline, draw=none, error bars/.cd, y dir=both, y explicit] coordinates {
            (5,18)
        };
        \addplot [fill=randompartitionarea, draw=none, error bars/.cd, y dir=both, y explicit] coordinates {
            (5,17.05) +- (0, 0.05)
        };
        \addplot [fill=modelsoupline, draw=none, error bars/.cd, y dir=both, y explicit] coordinates {
            (5,16.5)
        };
        \addplot [fill=expertarea, draw=none, error bars/.cd, y dir=both, y explicit] coordinates {
            (5,15.45) +- (0, 0.1)
        };
        \addplot [fill=mixedgmline, draw=none, error bars/.cd, y dir=both, y explicit] coordinates {
            (5,15.8)
        };
        \addplot [fill=karvonenline, draw=none, error bars/.cd, y dir=both, y explicit] coordinates {
            (5,14)
        };
        \end{axis}
    \end{tikzpicture}
    \end{subfigure}\\[-1mm]
    \begin{minipage}[t]{\linewidth}
        \centering
        {\scriptsize Stockfish level}
    \end{minipage}
    \caption{\textbf{Playing strength of \mom against all baselines.} Bars report average FIDEScore at each Stockfish skill level (0--5); Glicko-2 ratings estimated from individual round-robin tournaments against all six levels are reported in the legend.}
    \label{fig:overall_winrate}
\end{figure*}

\paragraph{\rqtag{RQ4} Do \gm experts internalize persona-specific stylistic signatures?}

Experts exhibit distinctive player structure in both their hidden representations and likelihood assignments.
Figure~\ref{fig:weight_activation_displacement} reveals that while weight distances from the seed increase monotonically with depth---reflecting progressive stylistic divergence during fine-tuning---the most pronounced functional separation emerges in the activation space.
Own-master games systematically induce larger expert-pair activation displacement than other-master games, with ratios peaking in layers~10--14.
This indicates that persona specialization is not uniformly distributed across the network; rather, it is most visible in the upper-middle decoder blocks, where structurally similar experts produce increasingly distinct hidden states when evaluated on positions from their reference masters.
The NLL analysis in Figure~\ref{fig:act_nll_phase_bars_b} corroborates this picture with a complementary behavioral signal.
Across all game phases, each expert assigns lower NLL to its own master's held-out games than to those of the remaining nine (higher advantage)---a positive specialization gap that is largest in the opening, where individual repertoires are most explicit, but remains consistently positive through the middlegame and endgame.
The persistence of the advantage beyond the opening rules out the confound that specialization is merely a byproduct of memorized early-game sequences, and instead implicates genuine strategic and positional preferences.
The color-stratified decomposition further reveals that own-master confidence is stronger on White-side positions across most experts, consistent with the greater expressive latitude afforded to the proactive side, where a player's preferred pawn structures and piece configurations are more directly imposed.
Persona-specific fine-tuning improves emulation accuracy over the seed for nearly all experts (Figure~\ref{tab:stylometry_summary}); moreover, each expert's accuracy on its own master's held-out games consistently exceeds the average accuracy of all other experts on the same positions---demonstrating that the acquired preferences are not merely a generic improvement in move prediction, but are directionally aligned with the target \gm's decision-making. 
NLL ranking converges on the same conclusion from an independent angle, identifying the own master as the top-1 most-likely player in 9 out of 10 cases.
These results establish that \gm experts do not merely overfit to surface move statistics but acquire player-specific behavioral preferences that manifest in the internal geometry of their representations, in their probabilistic confidence, and in their final move predictions.

\begin{figure*}[!t]
\centering
%
\begin{subfigure}[b]{0.50\textwidth}
\centering
    \colorlet{masterAnand}{blue!80!white}
    \colorlet{masterAronian}{red!80!white}
    \colorlet{masterCarlsen}{green!60!black}
    \colorlet{masterCaruana}{violet!80!white}
    \colorlet{masterFirouzja}{brown!80!white}
    \colorlet{masterGiri}{magenta!80!white}
    \colorlet{masterNakamura}{gray!80!white}
    \colorlet{masterNepomniachtchi}{lime!60!black}
    \colorlet{masterSo}{cyan!80!black}
    \colorlet{masterVachierLagrave}{orange!80!white}
    \colorlet{weightdistcolor}{black!88}
    \newcommand{\actlegend}[3]{%
        \mbox{%
        \tikz[baseline=-0.55ex]{
            \draw[#1,line width=1.1pt] (0,0) -- (0.22,0);
            \draw[#1, fill=white, line width=0.8pt] (0.11,0) circle (0.75pt);
        }\,{\fontsize{7}{7}\selectfont\ding{#2}}~#3%
        }%
    }
    \resizebox{\linewidth}{!}{%
    \begin{tikzpicture}
    \begin{axis}[
        name=weightaxis,
        width=.96\linewidth, height=1.2cm,
        scale only axis,
        grid=both, grid style={gray!25!white,dashed},
        ymajorgrids=true, xmajorgrids=true,
        axis background/.style={fill=plotbackground},
        xmin=-0.3, xmax=15.3,
        xtick={0,1,2,3,4,5,6,7,8,9,10,11,12,13,14,15},
        xticklabels={},
        ymin=0.00016, ymax=0.00033,
        ytick={0.00018,0.00024,0.00030},
        yticklabels={$1.8$,$2.4$,$3.0$},
        scaled y ticks=false,
        yticklabel style={font=\fontsize{6}{6}\selectfont, text width=1.3em, align=right},
        ylabel={$\Delta_{\mathcal{W}}^{(\ell)}$},
        ylabel style={font=\scriptsize},
    ]
    \addplot [name path=WeightUpper, draw=none] coordinates {
        (0,0.000217366)(1,0.000212620)(2,0.000215624)(3,0.000258746)
        (4,0.000272587)(5,0.000282461)(6,0.000296393)(7,0.000296719)
        (8,0.000298098)(9,0.000302202)(10,0.000312395)(11,0.000315649)
        (12,0.000320789)(13,0.000322787)(14,0.000321726)(15,0.000302812)};
    \addplot [name path=WeightLower, draw=none] coordinates {
        (0,0.000169499)(1,0.000167612)(2,0.000166611)(3,0.000197157)
        (4,0.000209212)(5,0.000216220)(6,0.000225054)(7,0.000224312)
        (8,0.000224239)(9,0.000227982)(10,0.000234516)(11,0.000235782)
        (12,0.000239534)(13,0.000243577)(14,0.000244123)(15,0.000235311)};
    \addplot [fill=weightdistcolor, fill opacity=0.13, draw=none]
        fill between[of=WeightUpper and WeightLower];
    \addplot [color=weightdistcolor, line width=1.2pt, mark=diamond*,
              mark size=1.15pt, mark options={fill=white}] coordinates {
        (0,0.000193433)(1,0.000190116)(2,0.000191118)(3,0.000227951)
        (4,0.000240900)(5,0.000249341)(6,0.000260723)(7,0.000260516)
        (8,0.000261169)(9,0.000265092)(10,0.000273456)(11,0.000275715)
        (12,0.000280162)(13,0.000283182)(14,0.000282925)(15,0.000269061)};
    \end{axis}
    \node[font=\scriptsize, anchor=south west]
        at ($(weightaxis.north west)+(0.02cm,0.02cm)$) {$\cdot 10^{-4}$};
    \begin{axis}[
        name=actaxis,
        at={($(weightaxis.south west)+(0,-0.2cm)$)},
        anchor=north west,
        width=.96\linewidth, height=4.25cm,
        scale only axis,
        grid=both, grid style={gray!25!white,dashed},
        ymajorgrids=true, xmajorgrids=true,
        axis background/.style={fill=plotbackground},
        xmin=-0.3, xmax=15.3,
        xtick={0,1,2,3,4,5,6,7,8,9,10,11,12,13,14,15},
        xlabel={Layer}, xlabel style={font=\footnotesize},
        ymin=0.8452, ymax=1.3801,
        ytick={0.850,0.900,0.950,1.000,1.050,1.100,1.150,1.200,1.250,1.300,1.350},
        ylabel={$\Delta_{\mathcal{H},\mathrm{own}}^{(\ell)} / \Delta_{\mathcal{H},\mathrm{other}}^{(\ell)}$},
        ylabel style={font=\footnotesize},
        xticklabel style={font=\fontsize{7}{7}\selectfont},
        yticklabel style={font=\fontsize{7}{7}\selectfont, text width=1.3em, align=right},
        extra y ticks={1.0}, extra y tick labels={},
        extra y tick style={grid=major, grid style={gray!80!white, dashed}},
    ]
    \addplot [name path=Carlsenupper, draw=none] coordinates {(0,1.014833)(1,1.040372)(2,1.027531)(3,1.046506)(4,1.070991)(5,1.105889)(6,1.110414)(7,1.128790)(8,1.137195)(9,1.219603)(10,1.284819)(11,1.262641)(12,1.237045)(13,1.221244)(14,1.202833)(15,1.129392)};
    \addplot [name path=Carlsenlower, draw=none] coordinates {(0,0.994721)(1,0.990288)(2,0.991913)(3,1.001318)(4,0.998002)(5,1.005793)(6,1.023771)(7,1.026392)(8,1.028719)(9,1.061282)(10,1.110681)(11,1.103124)(12,1.072824)(13,1.078387)(14,1.081739)(15,0.873465)};
    \addplot [fill=masterCarlsen, fill opacity=0.15, draw=none] fill between[of=Carlsenupper and Carlsenlower];
    \addplot [color=masterCarlsen, line width=1.2pt, mark=*, mark size=1.1pt, mark options={fill=white}]
        coordinates {(0,1.004777)(1,1.015330)(2,1.009722)(3,1.023912)(4,1.034497)(5,1.055841)(6,1.067092)(7,1.077591)(8,1.082957)(9,1.140443)(10,1.197750)(11,1.182883)(12,1.154935)(13,1.149815)(14,1.142286)(15,1.001429)};
    \addplot [name path=Nakamuraupper, draw=none] coordinates {(0,1.007255)(1,1.027106)(2,1.013476)(3,1.027745)(4,1.052011)(5,1.079560)(6,1.098203)(7,1.105425)(8,1.117175)(9,1.197271)(10,1.285669)(11,1.251061)(12,1.230826)(13,1.202119)(14,1.176438)(15,1.105055)};
    \addplot [name path=Nakamuralower, draw=none] coordinates {(0,1.000058)(1,0.986282)(2,0.988091)(3,0.995991)(4,0.995271)(5,0.988396)(6,0.997291)(7,0.990616)(8,0.989019)(9,1.012842)(10,1.078950)(11,1.055445)(12,1.045400)(13,1.026003)(14,1.018773)(15,0.958200)};
    \addplot [fill=masterNakamura, fill opacity=0.15, draw=none] fill between[of=Nakamuraupper and Nakamuralower];
    \addplot [color=masterNakamura, line width=1.2pt, mark=*, mark size=1.1pt, mark options={fill=white}]
        coordinates {(0,1.003656)(1,1.006694)(2,1.000784)(3,1.011868)(4,1.023641)(5,1.033978)(6,1.047747)(7,1.048021)(8,1.053097)(9,1.105057)(10,1.182310)(11,1.153253)(12,1.138113)(13,1.114061)(14,1.097605)(15,1.031627)};
    \addplot [name path=Anandupper, draw=none] coordinates {(0,1.010423)(1,1.037043)(2,1.033143)(3,1.031774)(4,1.026117)(5,1.026773)(6,1.043643)(7,1.038482)(8,1.039985)(9,1.079899)(10,1.137312)(11,1.136793)(12,1.123551)(13,1.139669)(14,1.139501)(15,1.227507)};
    \addplot [name path=Anandlower, draw=none] coordinates {(0,1.000119)(1,0.986730)(2,0.998384)(3,0.999974)(4,0.991276)(5,0.980525)(6,0.994819)(7,0.983816)(8,0.982021)(9,0.998790)(10,1.052533)(11,1.030128)(12,1.018533)(13,1.023328)(14,1.029950)(15,1.070426)};
    \addplot [fill=masterAnand, fill opacity=0.15, draw=none] fill between[of=Anandupper and Anandlower];
    \addplot [color=masterAnand, line width=1.2pt, mark=*, mark size=1.1pt, mark options={fill=white}]
        coordinates {(0,1.005271)(1,1.011886)(2,1.015764)(3,1.015874)(4,1.008696)(5,1.003649)(6,1.019231)(7,1.011149)(8,1.011003)(9,1.039345)(10,1.094923)(11,1.083460)(12,1.071042)(13,1.081499)(14,1.084725)(15,1.148966)};
    \addplot [name path=Soupper, draw=none] coordinates {(0,1.020624)(1,1.060970)(2,1.053908)(3,1.048275)(4,1.060123)(5,1.096721)(6,1.118455)(7,1.135673)(8,1.153954)(9,1.241478)(10,1.300185)(11,1.273035)(12,1.265341)(13,1.259500)(14,1.247164)(15,1.097315)};
    \addplot [name path=Solower, draw=none] coordinates {(0,0.997024)(1,0.983406)(2,0.993936)(3,0.996253)(4,1.000096)(5,1.001090)(6,1.016669)(7,1.016459)(8,1.022351)(9,1.051903)(10,1.114084)(11,1.096145)(12,1.080770)(13,1.066682)(14,1.063166)(15,0.922439)};
    \addplot [fill=masterSo, fill opacity=0.15, draw=none] fill between[of=Soupper and Solower];
    \addplot [color=masterSo, line width=1.2pt, mark=*, mark size=1.1pt, mark options={fill=white}]
        coordinates {(0,1.008824)(1,1.022188)(2,1.023922)(3,1.022264)(4,1.030110)(5,1.048905)(6,1.067562)(7,1.076066)(8,1.088152)(9,1.146691)(10,1.207135)(11,1.184590)(12,1.173056)(13,1.163091)(14,1.155165)(15,1.009877)};
    \addplot [name path=Aronianupper, draw=none] coordinates {(0,1.009626)(1,1.025653)(2,1.022122)(3,1.027240)(4,1.033347)(5,1.048424)(6,1.065472)(7,1.069571)(8,1.071376)(9,1.126185)(10,1.216332)(11,1.186425)(12,1.166704)(13,1.158944)(14,1.153937)(15,1.159682)};
    \addplot [name path=Aronianlower, draw=none] coordinates {(0,1.000205)(1,1.000073)(2,1.000550)(3,0.996632)(4,1.001373)(5,1.006142)(6,1.011709)(7,1.013198)(8,1.012055)(9,1.050418)(10,1.108686)(11,1.089186)(12,1.066639)(13,1.042333)(14,1.037755)(15,0.950920)};
    \addplot [fill=masterAronian, fill opacity=0.15, draw=none] fill between[of=Aronianupper and Aronianlower];
    \addplot [color=masterAronian, line width=1.2pt, mark=*, mark size=1.1pt, mark options={fill=white}]
        coordinates {(0,1.004916)(1,1.012863)(2,1.011336)(3,1.011936)(4,1.017360)(5,1.027283)(6,1.038591)(7,1.041385)(8,1.041715)(9,1.088301)(10,1.162509)(11,1.137806)(12,1.116671)(13,1.100639)(14,1.095846)(15,1.055301)};
    \addplot [name path=Caruanaupper, draw=none] coordinates {(0,1.012847)(1,1.083534)(2,1.058695)(3,1.056779)(4,1.057754)(5,1.073864)(6,1.079631)(7,1.082543)(8,1.099323)(9,1.149076)(10,1.182612)(11,1.185951)(12,1.181035)(13,1.179723)(14,1.170071)(15,1.131749)};
    \addplot [name path=Caruanalower, draw=none] coordinates {(0,0.998413)(1,0.999710)(2,0.991327)(3,0.993941)(4,1.004837)(5,1.018472)(6,1.024811)(7,1.024174)(8,1.027665)(9,1.058545)(10,1.112995)(11,1.100662)(12,1.087262)(13,1.071188)(14,1.063946)(15,0.906742)};
    \addplot [fill=masterCaruana, fill opacity=0.15, draw=none] fill between[of=Caruanaupper and Caruanalower];
    \addplot [color=masterCaruana, line width=1.2pt, mark=*, mark size=1.1pt, mark options={fill=white}]
        coordinates {(0,1.005630)(1,1.041622)(2,1.025011)(3,1.025360)(4,1.031296)(5,1.046168)(6,1.052221)(7,1.053359)(8,1.063494)(9,1.103810)(10,1.147803)(11,1.143306)(12,1.134148)(13,1.125455)(14,1.117008)(15,1.019245)};
    \addplot [name path=Firouzjaupper, draw=none] coordinates {(0,1.017852)(1,1.058771)(2,1.052124)(3,1.039210)(4,1.051829)(5,1.069421)(6,1.094629)(7,1.089550)(8,1.100944)(9,1.159149)(10,1.207641)(11,1.222908)(12,1.209871)(13,1.247274)(14,1.236242)(15,1.125912)};
    \addplot [name path=Firouzjalower, draw=none] coordinates {(0,0.998490)(1,0.992171)(2,0.993674)(3,0.998303)(4,0.996675)(5,0.992054)(6,1.031553)(7,1.027253)(8,1.021487)(9,1.054794)(10,1.087404)(11,1.111800)(12,1.081263)(13,1.106052)(14,1.091449)(15,0.848190)};
    \addplot [fill=masterFirouzja, fill opacity=0.15, draw=none] fill between[of=Firouzjaupper and Firouzjalower];
    \addplot [color=masterFirouzja, line width=1.2pt, mark=*, mark size=1.1pt, mark options={fill=white}]
        coordinates {(0,1.008171)(1,1.025471)(2,1.022899)(3,1.018757)(4,1.024252)(5,1.030737)(6,1.063091)(7,1.058401)(8,1.061216)(9,1.106972)(10,1.147522)(11,1.167354)(12,1.145567)(13,1.176663)(14,1.163846)(15,0.987051)};
    \addplot [name path=Giriupper, draw=none] coordinates {(0,1.033972)(1,1.108719)(2,1.099494)(3,1.085201)(4,1.126757)(5,1.183565)(6,1.196945)(7,1.216223)(8,1.231524)(9,1.324419)(10,1.377109)(11,1.357425)(12,1.364372)(13,1.342012)(14,1.305504)(15,1.192872)};
    \addplot [name path=Girilower, draw=none] coordinates {(0,1.008748)(1,1.017015)(2,1.011427)(3,1.000052)(4,1.016961)(5,1.045252)(6,1.045330)(7,1.040184)(8,1.053225)(9,1.122417)(10,1.206138)(11,1.201429)(12,1.199386)(13,1.180939)(14,1.153840)(15,0.923777)};
    \addplot [fill=masterGiri, fill opacity=0.15, draw=none] fill between[of=Giriupper and Girilower];
    \addplot [color=masterGiri, line width=1.2pt, mark=*, mark size=1.1pt, mark options={fill=white}]
        coordinates {(0,1.021360)(1,1.062867)(2,1.055460)(3,1.042627)(4,1.071859)(5,1.114408)(6,1.121138)(7,1.128203)(8,1.142374)(9,1.223418)(10,1.291623)(11,1.279427)(12,1.281879)(13,1.261475)(14,1.229672)(15,1.058324)};
    \addplot [name path=Nepomniachtchiupper, draw=none] coordinates {(0,1.015280)(1,1.067472)(2,1.060949)(3,1.043620)(4,1.070737)(5,1.099908)(6,1.119184)(7,1.130959)(8,1.139244)(9,1.218980)(10,1.314179)(11,1.293976)(12,1.281333)(13,1.260673)(14,1.239112)(15,1.152627)};
    \addplot [name path=Nepomniachtchilower, draw=none] coordinates {(0,0.997036)(1,0.989640)(2,0.988940)(3,1.003670)(4,1.007747)(5,1.006668)(6,1.027431)(7,1.024142)(8,1.022381)(9,1.063540)(10,1.097492)(11,1.095191)(12,1.075300)(13,1.081321)(14,1.082745)(15,0.947656)};
    \addplot [fill=masterNepomniachtchi, fill opacity=0.15, draw=none] fill between[of=Nepomniachtchiupper and Nepomniachtchilower];
    \addplot [color=masterNepomniachtchi, line width=1.2pt, mark=*, mark size=1.1pt, mark options={fill=white}]
        coordinates {(0,1.006158)(1,1.028556)(2,1.024945)(3,1.023645)(4,1.039242)(5,1.053288)(6,1.073307)(7,1.077551)(8,1.080812)(9,1.141260)(10,1.205836)(11,1.194583)(12,1.178316)(13,1.170997)(14,1.160928)(15,1.050142)};
    \addplot [name path=VachierLagraveupper, draw=none] coordinates {(0,1.017794)(1,1.073036)(2,1.055548)(3,1.069938)(4,1.100431)(5,1.136851)(6,1.144620)(7,1.156504)(8,1.165101)(9,1.214944)(10,1.250853)(11,1.244024)(12,1.241317)(13,1.196807)(14,1.166375)(15,1.073632)};
    \addplot [name path=VachierLagravelower, draw=none] coordinates {(0,0.997687)(1,0.990583)(2,0.990117)(3,1.002494)(4,1.001152)(5,1.001646)(6,1.016805)(7,1.019299)(8,1.030423)(9,1.054875)(10,1.097538)(11,1.096151)(12,1.073381)(13,1.064162)(14,1.056316)(15,0.934426)};
    \addplot [fill=masterVachierLagrave, fill opacity=0.15, draw=none] fill between[of=VachierLagraveupper and VachierLagravelower];
    \addplot [color=masterVachierLagrave, line width=1.2pt, mark=*, mark size=1.1pt, mark options={fill=white}]
        coordinates {(0,1.007741)(1,1.031810)(2,1.022833)(3,1.036216)(4,1.050792)(5,1.069248)(6,1.080713)(7,1.087902)(8,1.097762)(9,1.134909)(10,1.174196)(11,1.170087)(12,1.157349)(13,1.130485)(14,1.111345)(15,1.004029)};
    \end{axis}
    \end{tikzpicture}%
    }
    \vspace{1.0mm}
    \makebox[\linewidth][c]{%
    {\fontsize{5.1}{5.1}\selectfont
    \setlength{\tabcolsep}{2.5pt}
    \renewcommand{\arraystretch}{0.95}
    \begin{tabular}{@{}cccc@{}}
        \actlegend{masterAnand}{182}{Anand} &
        \actlegend{masterAronian}{183}{Aronian} &
        \actlegend{masterCarlsen}{184}{Carlsen} &
        \actlegend{masterCaruana}{185}{Caruana} \\[-1pt]
        \actlegend{masterFirouzja}{186}{Firouzja} &
        \actlegend{masterGiri}{187}{Giri} &
        \actlegend{masterNakamura}{188}{Nakamura} &
        \actlegend{masterNepomniachtchi}{189}{Nepomniachtchi} \\[-1pt]
        \multicolumn{4}{c}{%
            \actlegend{masterSo}{190}{So}
            \hspace{14pt}
            \actlegend{masterVachierLagrave}{191}{Vachier-Lagrave}
        }
    \end{tabular}%
    }%
    }
    \caption{%
        \textbf{Expert latent analysis.}
        ($\uparrow$)~Normalized Euclidean weight distance.
        ($\downarrow$)~Activation displacement ratio on own and other
        held-out games over all master pairs.}
    \label{fig:weight_activation_displacement}
\end{subfigure}%
\hfill
%
\begin{subfigure}[b]{0.46\textwidth}
\centering
    \resizebox{\linewidth}{!}{%
    \begin{tikzpicture}
      \def\avgOpening{0.06380}
      \def\avgMiddlegame{0.03229}
      \def\avgEndgame{0.01987}
      \def\avgFullgame{0.02927}
      \begin{groupplot}[
          group style={
              group size=2 by 2,
              horizontal sep=0.65cm,
              vertical sep=0.85cm,
          },
          width=0.4\linewidth, height=2.5cm,
          scale only axis,
          axis background/.style={fill=plotbackground},
          grid=both, grid style={gray!20!white,dashed},
          xmin=0.35, xmax=10.65,
          xtick={1,2,3,4,5,6,7,8,9,10},
          xticklabels={
              {\scriptsize\ding{182}},{\scriptsize\ding{183}},
              {\scriptsize\ding{184}},{\scriptsize\ding{185}},
              {\scriptsize\ding{186}},{\scriptsize\ding{187}},
              {\scriptsize\ding{188}},{\scriptsize\ding{189}},
              {\scriptsize\ding{190}},{\scriptsize\ding{191}}},
          ymin=-0.08, ymax=0.20,
          ytick={-0.08,-0.04,0.00,0.04,0.08,0.12,0.16,0.20},
          yticklabel style={
              /pgf/number format/fixed,
              /pgf/number format/fixed zerofill,
              /pgf/number format/precision=2,
              font=\fontsize{5.3}{5.3}\selectfont},
          every axis plot/.append style={forget plot},
          axis on top=true,
      ]
      \nextgroupplot[
          title={Opening},
          title style={font=\fontsize{7}{7}\selectfont, yshift=-1.2ex}]
      \addplot[color=black!35, line width=0.5pt, mark=none]
          coordinates {(0.35,0)(10.65,0)};
      \addplot[ybar,bar width=3.0pt,bar shift=-1.7pt,draw=black,fill=black!72]
          coordinates{(1,0.087754)(2,0.076679)(3,0.069564)(4,0.048759)(5,0.036424)
                      (6,0.089498)(7,0.098413)(8,0.072943)(9,0.077590)(10,-0.023466)};
      \addplot[ybar,bar width=3.0pt,bar shift=1.7pt,draw=black,fill=white]
          coordinates{(1,0.072030)(2,0.080891)(3,0.033677)(4,0.042300)(5,0.050545)
                      (6,0.061458)(7,0.166752)(8,0.071467)(9,0.069599)(10,-0.006870)};
      \addplot[color=red!80!white, dashed, line width=0.8pt, mark=none]
          coordinates {(0.35,\avgOpening)(10.65,\avgOpening)};
      \nextgroupplot[
          title={Middlegame},
          title style={font=\fontsize{7}{7}\selectfont, yshift=-1.2ex}]
      \addplot[color=black!35, line width=0.5pt, mark=none]
          coordinates {(0.35,0)(10.65,0)};
      \addplot[ybar,bar width=3.0pt,bar shift=-1.7pt,draw=black,fill=black!72]
          coordinates{(1,0.019713)(2,0.041072)(3,0.023714)(4,0.018433)(5,0.009388)
                      (6,0.048752)(7,0.028139)(8,0.040602)(9,0.049745)(10,0.001179)};
      \addplot[ybar,bar width=3.0pt,bar shift=1.7pt,draw=black,fill=white]
          coordinates{(1,0.020012)(2,0.041221)(3,0.037720)(4,0.022866)(5,0.049878)
                      (6,0.048441)(7,0.048329)(8,0.039637)(9,0.050206)(10,0.006769)};
      \addplot[color=red!80!white, dashed, line width=0.8pt, mark=none]
          coordinates {(0.35,\avgMiddlegame)(10.65,\avgMiddlegame)};
      \nextgroupplot[
          title={Endgame},
          title style={font=\fontsize{7}{7}\selectfont, yshift=-1.2ex},
          xlabel style={font=\fontsize{6.5}{6.5}\selectfont, yshift=0.5ex}]
      \addplot[color=black!35, line width=0.5pt, mark=none]
          coordinates {(0.35,0)(10.65,0)};
      \addplot[ybar,bar width=3.0pt,bar shift=-1.7pt,draw=black,fill=black!72]
          coordinates{(1,-0.015894)(2,0.018001)(3,0.017853)(4,0.004683)(5,-0.009004)
                      (6,0.043537)(7,0.012824)(8,0.028490)(9,0.033472)(10,0.007719)};
      \addplot[ybar,bar width=3.0pt,bar shift=1.7pt,draw=black,fill=white]
          coordinates{(1,-0.023691)(2,0.029091)(3,0.037222)(4,0.020328)(5,0.043807)
                      (6,0.051343)(7,0.023115)(8,0.026212)(9,0.039067)(10,0.009289)};
      \addplot[color=red!80!white, dashed, line width=0.8pt, mark=none]
          coordinates {(0.35,\avgEndgame)(10.65,\avgEndgame)};
      \nextgroupplot[
          title={Full game},
          title style={font=\fontsize{7}{7}\selectfont, yshift=-1.2ex},
          xlabel style={font=\fontsize{6.5}{6.5}\selectfont, yshift=0.5ex}]
      \addplot[color=black!35, line width=0.5pt, mark=none]
          coordinates {(0.35,0)(10.65,0)};
      \addplot[ybar,bar width=3.0pt,bar shift=-1.7pt,draw=black,fill=black!72]
          coordinates{(1,0.015889)(2,0.034001)(3,0.022751)(4,0.013685)(5,0.005064)
                      (6,0.045443)(7,0.026212)(8,0.038059)(9,0.045531)(10,0.003956)};
      \addplot[ybar,bar width=3.0pt,bar shift=1.7pt,draw=black,fill=white]
          coordinates{(1,0.011077)(2,0.037179)(3,0.036022)(4,0.023021)(5,0.045717)
                      (6,0.047115)(7,0.042821)(8,0.038825)(9,0.046348)(10,0.006720)};
      \addplot[color=red!80!white, dashed, line width=0.8pt, mark=none]
          coordinates {(0.35,\avgFullgame)(10.65,\avgFullgame)};
      \end{groupplot}
      \node[rotate=90, anchor=center, font=\fontsize{7}{7}\selectfont]
          at ($(current bounding box.west)+(-0.10cm,0)$)
          {NLL advantage ($\uparrow$)};
    \end{tikzpicture}%
    }
    \vspace{1.5mm}
    \noindent\makebox[\linewidth][c]{%
        {\hspace{0.9cm}\fontsize{6.5}{7}\selectfont
        \tikz[baseline=0.15ex]{\fill[black!72] (0,0) rectangle (0.24,0.14);}\enspace Black-side games
        \hspace{10pt}
        \tikz[baseline=0.15ex]{\draw[black] (0,0) rectangle (0.24,0.14);}\enspace White-side games}%
    }
    \caption{%
        \textbf{Own-master NLL advantage.}
        NLL advantage on 1,000 held-out games. The higher the better.
        The red dashed line is the average NLL across experts.}
    \label{fig:act_nll_phase_bars_b}
\end{subfigure}
%
\\[4mm]
\begin{subfigure}[t]{\linewidth}
\centering
    \begin{adjustbox}{width=\linewidth}
    \begin{tabular}{l !{\vrule width 0.8pt} l !{\vrule width 0.8pt} cccccccccc}
        \hline
        \rowcolor{colHd}
        \textbf{Setting}
            & \textbf{Metric}
            & {\Large\ding{182}} & {\Large\ding{183}} & {\Large\ding{184}} & {\Large\ding{185}}
            & {\Large\ding{186}} & {\Large\ding{187}} & {\Large\ding{188}} & {\Large\ding{189}}
            & {\Large\ding{190}} & {\Large\ding{191}} \\
        \hline
        Seed
            & Emulation Acc.
            & 43.43 & 45.41 & 47.24 & 46.92 & 48.96
            & 44.64 & 47.67 & 45.50 & 46.96 & 46.19 \\
        Expert
            & Emulation Acc.
            & 45.20 & 48.18 & 49.33 & 49.13 & 48.97
            & 49.29 & 48.82 & 47.67 & 49.86 & 46.49 \\
        Expert
            & Others' Emulation Acc.
            & 43.69$_{\pm0.50}$
            & 45.69$_{\pm0.41}$
            & 48.08$_{\pm0.59}$
            & 47.34$_{\pm0.80}$
            & 48.63$_{\pm0.66}$
            & 45.35$_{\pm0.49}$
            & 47.98$_{\pm0.55}$
            & 45.77$_{\pm0.40}$
            & 47.24$_{\pm0.42}$
            & 46.61$_{\pm0.44}$ \\
        \hline
    \end{tabular}
    \end{adjustbox}
    \caption{%
        \textbf{Emulation accuracy before and after persona-specific fine-tuning.}
        Own-master accuracy is compared against the average accuracy obtained by each expert on held-out games.}
    \label{tab:stylometry_summary}
\end{subfigure}
\caption{%
    \textbf{Expert specialization diagnostics.}
    \textit{(a)}~Layer-wise weight distance and activation displacement ratio across experts, measuring the degree to which persona fine-tuning produces functionally
    distinct internal representations.
    \textit{(b)}~Own-master NLL advantage by game phase and side,
    quantifying each expert's predictive alignment with its target \gm
    on held-out games.
    \textit{(c)}~Emulation accuracy.}
\label{fig:expert_diagnostics}
\end{figure*}

\paragraph{\rqtag{RQ5} Do \mom activation patterns reflect interpretable and meaningful style transitions?}

\begin{figure*}[!t]
    \centering
    \begin{subfigure}[c]{6.55cm}
        \includegraphics[width=\textwidth]{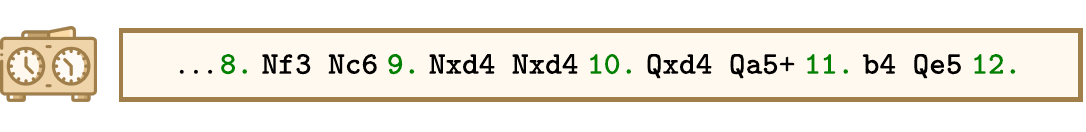}
    \end{subfigure}
    \hfill
    \begin{subfigure}[c]{6.55cm}
        \hspace{-0.1mm}
        \includegraphics[width=\textwidth]{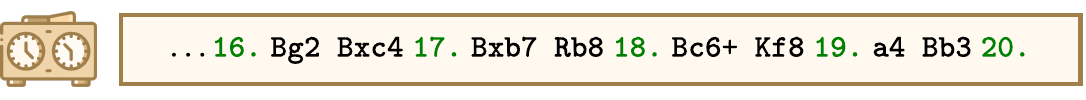}
    \end{subfigure}
    \\[-1mm]
    \begin{subfigure}[c]{2.5cm}
        \centering
        \includegraphics[width=\textwidth]{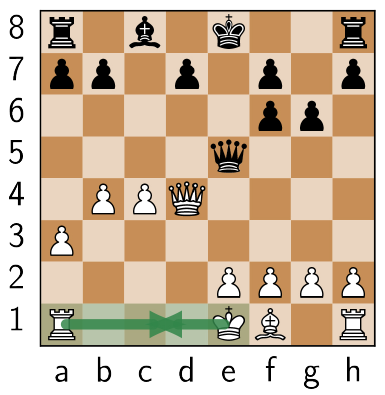}
    \end{subfigure}
    \hfill
    \begin{subfigure}[c]{3.9cm}
        \centering
        \vspace{1mm}
        \includegraphics[width=\textwidth]{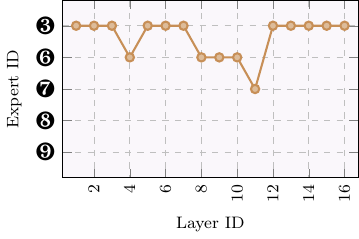}
    \end{subfigure}
    \hspace{8mm}
    \begin{subfigure}[c]{2.5cm}
        \centering
        \includegraphics[width=\textwidth]{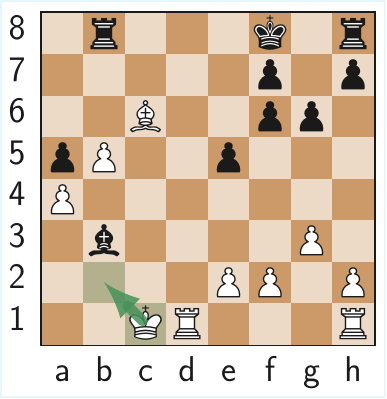}
    \end{subfigure}
    \hfill
    \begin{subfigure}[c]{3.9cm}
        \centering
        \vspace{1mm}
        \includegraphics[width=\textwidth]{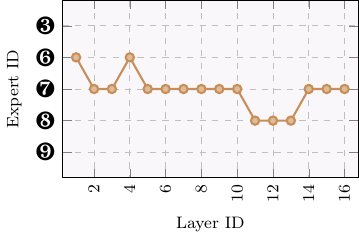}
    \end{subfigure}
    \caption{\textbf{Visualization of how \mom activated experts vary when playing a game at test time against Stockfish.} Decoder block top-$1$ routing paths for two distinct board states. \mom (White) dynamically adjusts expert utilization in response to the evolving position.}
    \label{fig:expert_activation}
\end{figure*}

The stitched \mom model is never explicitly trained to activate specific experts for specific positions, since routing emerges entirely from end-to-end optimization with no supervision on which expert to select.
We evaluate this behavior on samples balanced by reference master and routed-expert activation, preventing the routing statistics from being driven by uneven master coverage or repeated selection of the same branch.
Despite this, Figure~\ref{tab:routing_analysis} shows that \mom spontaneously concentrates its routing mass on two experts rather than spreading it uniformly across all five.
In $99.9\%$ of forward passes the top-2 experts jointly hold at least $50\%$ of routing mass, and in $84.0\%$ of passes they hold at least $60\%$.
The boundary between selected and unselected experts is sharp, with the gap between the second- and third-ranked expert exceeding five percentage points in $65.1\%$ of passes.
Random-partition \mom controls trained with identical architecture and recipe reach only $18$--$21\%$ at the $50\%$ concentration threshold and collapse to near zero at stricter levels.
Because the only difference between \mom and the random-partition baselines is whether experts encode \gm identity or arbitrary data partitions, the emergent discretization is a direct consequence of persona-grounded specialization rather than repeated collapse onto the same branch.
This tendency to make sharp yet balanced routing decisions also translates into stronger play, as discussed in RQ3.
The qualitative routing traces in Figure~\ref{fig:expert_activation} make this mechanism visible during play: as the position evolves, the top-routed expert changes across decoder blocks and board states, suggesting that \mom conditionally recruits different stylistic biases in specific situations rather than uniformly averaging all experts.
In the example of Figure~\ref{fig:expert_activation}, the early middlegame is dominated by experts associated with dynamic attacking play, consistent with \mom's decision to castle queenside with \texttt{12.O-O-O} despite Black's counterplay.
Later, the routing mass shifts toward more tactical experts before the sacrifice on \texttt{20.Kb2}, where \mom gives material to sustain the initiative.

\begin{figure*}[!t]
\centering
\setlength{\aboverulesep}{0pt}
\setlength{\belowrulesep}{0pt}
\setlength{\extrarowheight}{0pt}
\setlength{\tabcolsep}{3.5pt}
\renewcommand{\arraystretch}{1.22}
%
\begin{subfigure}[b]{0.265\linewidth}
    \begin{tikzpicture}
        \begin{axis}[
            width=4cm, height=3.2cm,
            ymajorgrids=true, grid=both, grid style={dashed,gray!25},
            axis background/.style={fill=plotbackground},
            boxplot/draw direction=y, boxplot={box extend=0.38},
            enlarge x limits=0.18,
            xtick={1,2,3,4,5}, xticklabels={1,2,3,4,5},
            xlabel={$k$ (active experts)},
            xlabel style={font=\fontsize{8}{8}\selectfont},
            ymin=52, ymax=83, ylabel={FIDEScore},
            ylabel style={font=\fontsize{8}{8}\selectfont},
            every tick label/.append style={font=\fontsize{7}{7}\selectfont},
        ]
        \addplot+[boxplot prepared={median=68,upper quartile=70.5,lower quartile=66.8,
            upper whisker=74.5,lower whisker=58},
            draw=black,fill=trainingbar!70] coordinates {};
        \addplot+[boxplot prepared={median=68.8,upper quartile=71.9,lower quartile=66.3,
            upper whisker=77.5,lower whisker=61},
            draw=topkborder,line width=1.1pt,fill=topkfill] coordinates {};
        \addplot+[boxplot prepared={median=66.8,upper quartile=72,lower quartile=64.6,
            upper whisker=76.5,lower whisker=57.5},
            draw=black,fill=trainingbar!70] coordinates {};
        \addplot+[boxplot prepared={median=68.7,upper quartile=70,lower quartile=64.7,
            upper whisker=73.5,lower whisker=58},
            draw=black,fill=trainingbar!70] coordinates {};
        \addplot+[boxplot prepared={median=64.3,upper quartile=69.4,lower quartile=63.6,
            upper whisker=71.5,lower whisker=56},
            draw=black,fill=trainingbar!70] coordinates {};
        \addplot[bellcolor,smooth,domain=0.5:5.5,samples=120]
            {64 + 5.5*exp(-0.35*(x-1.9)^2)};
        \node[font=\fontsize{11}{11}\selectfont,text=topkborder,anchor=south]
            at (axis cs:2,76.5) {$\boldsymbol{\star}$};
        \end{axis}
    \end{tikzpicture}
    \caption{\textbf{FIDEScore vs.\ number of active experts (top-$k$)}. \mom. Stockfish lv.\,0, pooled over 10 runs.}
    \label{fig:topk_ablation}
\end{subfigure}%
\hfill
{\color{panSep}\rule{0.5pt}{4.15cm}}%
\hfill
%
\begin{subfigure}[b]{0.69\linewidth}
    \centering\footnotesize
    \resizebox{\linewidth}{!}{%
    \begin{tabular}{lcccccccccc}
        \toprule
        \rowcolor{colHd}
        \textbf{Model} &
        \multicolumn{5}{c}{\textbf{Concentration: top-2 mass $\geq\!X$ (\%)}} &
        \multicolumn{5}{c}{\textbf{Sharpness: \#2 vs.\ \#3 gap (pp)}} \\
        \cmidrule(lr){2-6}\cmidrule(l){7-11}
        \rowcolor{colHd}
        &
        \makebox[2.8em][c]{${\geq}50$} &
        \makebox[2.8em][c]{${\geq}55$} &
        \makebox[2.8em][c]{${\geq}60$} &
        \makebox[2.8em][c]{${\geq}65$} &
        \makebox[2.8em][c]{${\geq}70$} &
        \makebox[2.8em][c]{Mean} &
        \makebox[2.8em][c]{Std} &
        \makebox[2.8em][c]{${>}5$} &
        \makebox[2.8em][c]{${>}10$} &
        \makebox[2.8em][c]{${>}15$} \\
        \midrule
        \mom &
        \cellcolor{concentrationcolor!100}\textcolor{white}{99.9} &
        \cellcolor{concentrationcolor!96}\textcolor{white}{96.1} &
        \cellcolor{concentrationcolor!84}\textcolor{white}{84.0} &
        \cellcolor{concentrationcolor!59}\textcolor{white}{58.9} &
        \cellcolor{concentrationcolor!22}{22.1} &
        \cellcolor{sharpnesscolor!8}{7.83} &
        \cellcolor{sharpnesscolor!5}{5.48} &
        \cellcolor{sharpnesscolor!65}\textcolor{white}{65.1} &
        \cellcolor{sharpnesscolor!33}{33.8} &
        \cellcolor{sharpnesscolor!11}{11.4} \\
        \midrule
        P1 &
        \cellcolor{concentrationcolor!21}21.2 & \cellcolor{concentrationcolor!2}2.1 &
        \cellcolor{concentrationcolor!0}0.2  & 0.0 & 0.0 &
        \cellcolor{sharpnesscolor!2}2.14 & \cellcolor{sharpnesscolor!2}1.93 &
        \cellcolor{sharpnesscolor!10}10.2 & \cellcolor{sharpnesscolor!0}0.2 & 0.0 \\
        P2 &
        \cellcolor{concentrationcolor!18}18.4 & \cellcolor{concentrationcolor!3}2.8 &
        \cellcolor{concentrationcolor!1}0.6  & 0.0 & 0.0 &
        \cellcolor{sharpnesscolor!2}2.12 & \cellcolor{sharpnesscolor!2}2.05 &
        \cellcolor{sharpnesscolor!10}9.8  & \cellcolor{sharpnesscolor!1}0.8 & 0.0 \\
        P3 &
        \cellcolor{concentrationcolor!19}18.8 & \cellcolor{concentrationcolor!1}1.2 &
        0.0 & 0.0 & 0.0 &
        \cellcolor{sharpnesscolor!2}2.11 & \cellcolor{sharpnesscolor!2}1.84 &
        \cellcolor{sharpnesscolor!10}9.9  & \cellcolor{sharpnesscolor!0}0.2 & 0.0 \\
        \bottomrule
    \end{tabular}%
    }%
    \caption{\textbf{Concentration and sharpness of the routing distribution} (softmax over all 5 experts). Stockfish lv.0. \textit{Left}: fraction of forward passes where the top-2 experts jointly hold $\geq\!X\%$ of routing mass. \textit{Right}: activation gap (pp) between the 2nd- and 3rd-ranked expert per pass. \mom vs.\ random-partition \mom baselines (P1--P3).}
    \label{tab:routing_analysis}
\end{subfigure}
\caption{\textbf{Router behaviour at inference time}. \textit{(a)}~Playing strength under top-$k$ routing for \mom. \textit{(b)}~Emergent routing discretization in \mom: \gm partition vs. random partitions.}
\label{fig:routing_full}
\end{figure*}

\section{Conclusions}

This paper challenges the conventional practice of training dense chess language models on aggregated, player-undistinguished datasets.
We introduce \mom, the first chess MoE that combines independently trained \gm networks through weight merging and lightweight player routing.
The resulting model improves over individual experts and controlled baselines in games against Stockfish, while retaining the behavioral diversity that dense aggregation tends to suppress.

Our stylometric analyses further show that the expert branches preserve measurable player-specific structure.
Although the router is not trained with player-identity labels or direct stylistic objectives, it develops sharp expert-selection patterns that align with the specialized behavior of the underlying branches.
This suggests that the performance gains of \mom are not only a consequence of increased
capacity, but also of meaningful specialization emerging from sparse composition.

More broadly, chess provides a controlled testbed for studying specialization in sparse language models.
\mom shows that, when expert branches encode coherent behavioral modes, MoE architectures can exploit their differences without collapsing them into a single averaged model.
This supports a view of specialization as useful not only for efficiency or scaling, but also for preserving complementary behaviors that improve downstream decision-making.

Collectively, our results validate persona-specialized sparse modeling as a promising direction for chess language models and compositional AI systems, offering a path toward models that are stronger, more controllable, more inspectable, and behaviorally plural.

\clearpage
\newpage



{
\small
\bibliographystyle{ieee}
\bibliography{bibliography}
}

\clearpage
\newpage



\appendix

\begin{appendices}

\startcontents[sections]
\printcontents[sections]{l}{1}{\setcounter{tocdepth}{2}}

\clearpage
\newpage

\section{Survey}
\label{app:survey}

In parallel to the methodological and resource contributions presented in the main paper, we designed and administered a survey aimed at clarifying long-standing open questions in the chess community that directly underpin our modeling approach.
In particular, the survey seeks to explore the viewpoint of participants on four fundamental dimensions: (i) the perceived possibility of identifying professional players through the sole observation of their games, (ii) the existence and definition of the notion of \quotes{playing style,} (iii) the practical feasibility of assigning coherent style categories to \gms, and (iv) the extent to which chess engines and AI models influence human play.
Since \mom presupposes assumptions about style and player recognition---concepts that remain contested even among experts---this empirical complement serves as a critical validation step.

We selected the \textit{Alma Mater Studiorum Chess Tournament 2025}\footnote[4]{\url{https://events.unibo.it/alma-mater-university-chess-tournament}} as primary venue for data collection.
An international academic competition organized by the University of Bologna and held behind closed doors from September 12 to 14, 2025, at the Biblioteca Universitaria di Bologna, Italy.
The event convened 72 mixed-gender players, grouped into 18 teams of four members each, representing some of the world's most prestigious universities from 10 countries across three continents.
The selection process for these teams was notably rigorous, as each institution was responsible for fielding its most talented representatives, often through internal qualification tournaments.
Consequently, the participant pool---composed entirely of adult English-speaking students (from bachelor to Ph.D. level) and faculty members---included players of exceptional caliber, among them national champions.
The tournament structure consisted of five playing sessions governed by the Swiss system with a time control of 45 minutes plus a 10-second increment per move.
It was overseen by arbiters from the Italian Chess Federation and was not rated by FIDE to preserve its inclusive and collegial character, prioritizing cultural exchange and sportsmanship.
The event received live commentary on Chess.com channels\footnote[5]{\url{https://www.chess.com/it/events/alma-mater-chess-tournament-2025}} and featured an AI analysis room sponsored by Intel.

The decision to anchor our study in this specific tournament was deliberate to ensure the collection of high-quality and reliable data from a culturally diverse participant base.
In stark contrast to large-scale online surveys, where participant veracity and expertise can be difficult to ascertain, this setting provided a controlled environment with a verified cohort of competent players.
The context also provided an atmosphere of intellectual openness and reflection, well suited for a survey.

\paragraph{Data collection}
Our data collection protocol was executed in two distinct phases.
The first phase took place in person during the three days of the tournament.
This direct interaction encouraged thoughtful, authentic responses, collected in an environment free from external distractions.
The closed-door format of the event allowed us to engage not only players but also arbiters and AI experts, thereby broadening the scope of informed perspectives. 
Recognizing that the demanding tournament schedule could limit participation, we initiated a second phase post-event.
An online version of the survey was made available for a limited period to allow contributions from individuals who were unable to complete it on-site, as well as to include additional voices from the broader chess community, such as members of chess clubs who did not attend the tournament.
Throughout both data collection phases, strict ethical and procedural standards were maintained.
We ensured all respondents were over the age of 18 and obtained their informed consent.
The submission of responses was strictly voluntary, without financial or other incentives.
Survey users were not shown their previous answers and aggregate results during or after the collection process, a measure implemented to mitigate potential conformity biases.
On average, completing the survey required approximately eight minutes.
To guarantee participant privacy, the survey was designed to be fully anonymous.
No personally identifiable information---such as names, email addresses, or IP addresses---nor any other sensitive data was gathered.

\subsection{Participant geography and demographics}

\begin{figure}[!htb]
    \centering
    \includegraphics[width=1.0\linewidth]{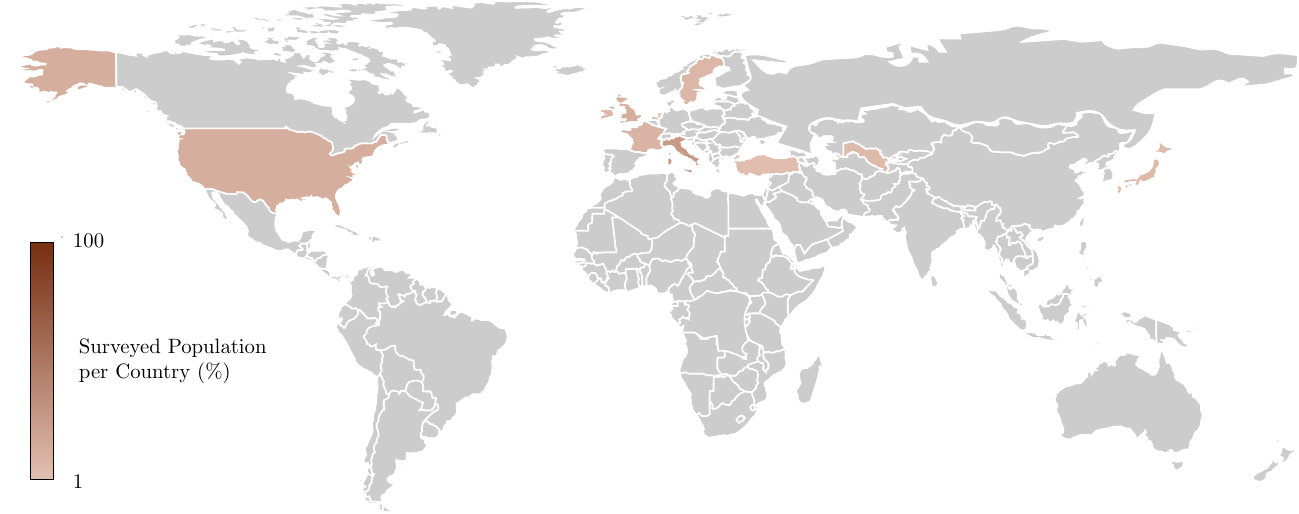}
    \caption{\textbf{Geographic distribution of survey participants by affiliation country.}}
    \label{fig:survey_participants_geographies}
\end{figure}

The survey solicited non-identifying, high-level demographic information: affiliation name, affiliation country, and current Elo rating.
For each of these items, a \textit{\quotes{Prefer not to say}} option was supplied to respect respondent privacy.
Our sample covers a broad heterogeneity in both demographic and geographic terms, with 50 responses obtained from all the 18 competing universities, arbiters and independent experts.
This diversity was crucial, allowing us to capture opinions across multiple chess traditions and educational backgrounds.
Simultaneously, the shared academic context delivered sufficient common ground to ensure meaningful comparability of responses.
Our respondents included members of teams from Yale and Harvard in the United States, and from Oxford and Cambridge in the United Kingdom---pairs of institutions whose chess rivalries trace back more than 150 years.
Geographic breakdowns are presented in Figure~\ref{fig:survey_participants_geographies}, while demographic characteristics are summarized in Table~\ref{tab:survey_participants_stats}.

\begin{table}[!htb]
    \centering
    \caption{\textbf{Demographics of survey participants ($N=50$).} Overall distribution of Elo ratings. Counts and percentages of participants by affiliation.}
    \label{tab:survey_participants_stats}
    \resizebox{\textwidth}{!}{
    \begin{threeparttable}
    \begin{tabular}{p{1.6cm}lllrl}
        \toprule
        \textbf{Elo} & \textbf{Continent} & \textbf{Country} & \textbf{Affiliation}\tnote{$\dagger$} & \multicolumn{2}{l}{\textbf{Participants}} \\
        \toprule
        \multirow{24}{*}{\includegraphics[width=1.35cm]{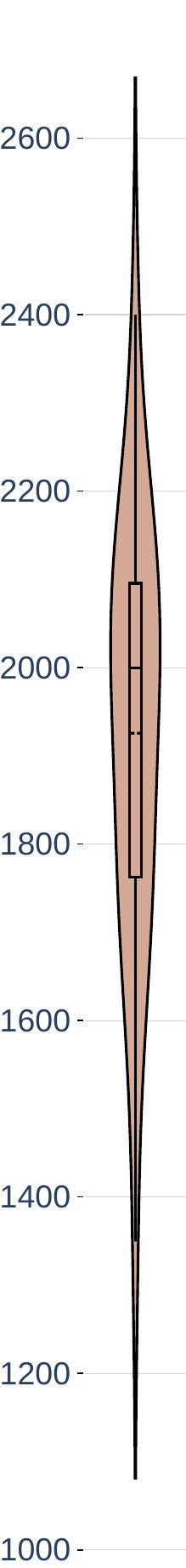}} & \multirow{22}{*}{\textbf{Europe}} & \multirow{6}{*}{\textbf{Italy}} & Alma Mater Studiorum -- Università di Bologna & 4 & \pcb{8} \% \\
        & & & {\cellcolor{tablehighlight}Università di Pisa} & {\cellcolor{tablehighlight}2} & {\cellcolor{tablehighlight}\pcb{4} \%} \\
        & & & Università degli Studi di Padova & 2 & \pcb{4} \% \\
        & & & {\cellcolor{tablehighlight}Università degli Studi di Milano Bicocca} & {\cellcolor{tablehighlight}1} & {\cellcolor{tablehighlight}\pcb{2} \%} \\
        & & & Università degli Studi di Napoli Federico II & 1 & \pcb{2} \% \\
        & & & {\cellcolor{tablehighlight}Other} & {\cellcolor{tablehighlight}3} & {\cellcolor{tablehighlight}\pcb{6} \%} \\
        \cmidrule{3-6}
        & & \multirow{3}{*}{\textbf{United Kingdom}} & University of Oxford & 2 & \pcb{4} \% \\
        & & & {\cellcolor{tablehighlight}University of Cambridge} & {\cellcolor{tablehighlight}2} & {\cellcolor{tablehighlight}\pcb{4} \%} \\
        & & & Other & 2 & \pcb{4} \% \\
        \cmidrule{3-6}
        & & \multirow{3}{*}{\textbf{Ireland}} & {\cellcolor{tablehighlight}Trinity College Dublin} & {\cellcolor{tablehighlight}2} & {\cellcolor{tablehighlight}\pcb{4} \%} \\
        & & & University College Dublin & 2 & \pcb{4} \% \\
        & & & {\cellcolor{tablehighlight}Other} & {\cellcolor{tablehighlight}1} & {\cellcolor{tablehighlight}\pcb{2} \%} \\
        \cmidrule{3-6}
        & & \multirow{2}{*}{\textbf{France}} & Université Paris 1 Panthéon Sorbonne & 2 & \pcb{4} \% \\
        & & & {\cellcolor{tablehighlight}Other} & {\cellcolor{tablehighlight}2} & {\cellcolor{tablehighlight}\pcb{4} \%} \\
        \cmidrule{3-6}
        & & \multirow{2}{*}{\textbf{Netherlands}} & Maastricht University & 2 & \pcb{4} \% \\
        & & & {\cellcolor{tablehighlight}Eindhoven University of Technology} & {\cellcolor{tablehighlight}1} & {\cellcolor{tablehighlight}\pcb{2} \%} \\
        \cmidrule{3-6}
        & & \multirow{2}{*}{\textbf{Sweden}} & Lund University & 2 & \pcb{4} \% \\
        & & & {\cellcolor{tablehighlight}Other} & {\cellcolor{tablehighlight}1} & {\cellcolor{tablehighlight}\pcb{2} \%} \\
        \cmidrule{2-6}
        & \multirow{3}{*}{\shortstack[l]{\textbf{North}\\\textbf{America}}} & \multirow{3}{*}{\shortstack[l]{\textbf{United States}\\\textbf{of America}}} & Harvard University & 2 & \pcb{4} \% \\
        & & & {\cellcolor{tablehighlight}Yale University} & {\cellcolor{tablehighlight}2} & {\cellcolor{tablehighlight}\pcb{4} \%} \\
        & & & Other & 2 & \pcb{4} \% \\
        \cmidrule{2-6}
        & \multirow{4}{*}{\textbf{Asia}} & \textbf{Japan} & {\cellcolor{tablehighlight}Keio University} & {\cellcolor{tablehighlight}2} & {\cellcolor{tablehighlight}\pcb{4} \%} \\
        \cmidrule{3-6}
        & & \textbf{Turkey} & Bogazici University & 1 & \pcb{2} \% \\
        \cmidrule{3-6}
        & & \textbf{Uzbekistan} & {\cellcolor{tablehighlight}Samarkand State University} & {\cellcolor{tablehighlight}1} & {\cellcolor{tablehighlight}\pcb{2} \%} \\
        \cmidrule{2-6}
        & \multicolumn{3}{r}{\textit{Prefer not to say}} & 6 & \pcb{12} \% \\
        \bottomrule
    \end{tabular}
    \begin{tablenotes}
    \item[$\dagger$] Other = non-university participants.
    \end{tablenotes}
    \end{threeparttable}
    }
\end{table}

\subsection{Player recognizability}

Some experts argue that professional players can indeed be recognized from their moves alone, pointing to recent machine learning studies that achieve high accuracy in attributing games even when results and openings are excluded~\cite{DBLP:conf/nips/McIlroy-YoungWS21}, suggesting that mid- and late-game decisions carry individual traces.
Others, however, caution that such recognizability diminishes among elite grandmasters, whose choices converge toward objective best play, making distinctions far less clear.
The debate therefore hinges on whether the residual patterns left in high-level games are strong enough to constitute a reliable identity marker, or whether recognizability is largely an artifact of broader repertoires and tendencies observable outside the very top tier.
To explore how this issue is perceived in practice, we sought to probe the opinion of our sample by submitting the following question:

\begin{surveyquestion}
There is a longstanding discussion in chess literature as to whether a player's identity can be inferred from the moves alone. Classical commentators and modern machine learning studies suggest that players exhibit \quotes{fingerprints} in their decision making. This raises the question whether recognizability through move patterns is accepted among experts.

To what extent do you agree with the statement:\\
\textit{\quotes{Professional chess players can be recognized by the moves they play, independently of the final result.}}
\end{surveyquestion}

\begin{figure}[!htb]
    \centering
    \includegraphics[width=.8\textwidth]{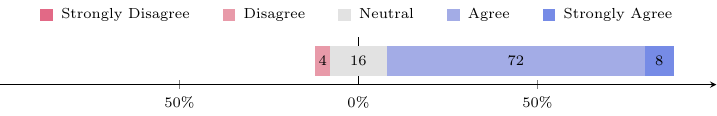}
    \caption{\textbf{Distribution of responses to the statement that professional chess players are recognizable by their moves alone.} The horizontal stacked bar represents the proportion of respondents on a five-point Likert scale (from Strongly Disagree to Strongly Agree).}
    \label{fig:survey_recognizability}
\end{figure}

The question was framed to omit any mention of the high accuracy rates achieved by prior AI studies, ensuring that responses would reflect participants' genuine beliefs rather than being primed by this information.
The distribution of responses in Figure~\ref{fig:survey_recognizability} shows a positive-skewed distribution.
A clear majority (80\%) agree or strongly agree that professional players can be recognized from their moves.
This level of endorsement is considerably higher than expected, given the persistent debate in the community and the presence of skeptical positions regarding the reliability of such recognizability at the elite level.
Although a minority of respondents expressed reservations, the overall pattern provides strong evidence-based support for our behavioral stylometry model-based metrics.

The strong consensus on the existence of player recognizability motivates a deeper inquiry into its nature.
We therefore posed a follow-up question designed to identify the specific factors that practitioners believe constitute a player's identity.
This question was administered to the entire cohort to understand which factors contribute to the definition of a chess persona, independent of whether those factors are ultimately considered strong enough for reliable identification.

\begin{surveyquestion}
Although many agree that players can be recognized from their games, it is far less clear what exactly makes them recognizable. E.g., what makes Kasparov \quotes{Kasparov}? The identity of a chess player appears to be multidimensional, and even experts often disagree on which aspects are most decisive. Understanding which dimensions practitioners themselves consider relevant is crucial for clarifying the concept of \quotes{chess persona.}

\textit{Which of the following factors, in your opinion, most contribute to making a player recognizable?}
\end{surveyquestion}

\begin{figure}[!htb]
    \begin{subfigure}[b]{0.24\textwidth}
        \centering
        \begin{minipage}{\textwidth}
            \centering
            \parbox[c][20pt][c]{\textwidth}{%
                \centering\scriptsize
                Preferred openings and early move repertoire
            }\\[2mm]
            \begin{tikzpicture}[font=\scriptsize]
                \def\yes{60}
                \wheelchart[
                    data=,
                    middle={\yes\%},
                    radius={0.5}{1},
                ]{
                \yes/yes-color/Yes,
                {100-\yes}/no-color/No
                }
            \end{tikzpicture}
        \end{minipage}
    \end{subfigure}
    \begin{subfigure}[b]{0.24\textwidth}
        \centering
        \begin{minipage}{\textwidth}
            \centering
            \parbox[c][20pt][c]{\textwidth}{%
                \centering\scriptsize
                Inclination toward aggressive\\play (sacrifices, king attacks)\\or defensive solidity
            }\\[2mm]
            \begin{tikzpicture}[font=\scriptsize]
                \def\yes{88}
                \wheelchart[
                    data=,
                    middle={\yes\%},
                    radius={0.5}{1},
                ]{
                \yes/yes-color/Yes,
                {100-\yes}/no-color/No
                }
            \end{tikzpicture}
        \end{minipage}
    \end{subfigure}
    \begin{subfigure}[b]{0.24\textwidth}
        \centering
        \begin{minipage}{\textwidth}
            \centering
            \parbox[c][20pt][c]{\textwidth}{%
                \centering\scriptsize
                Choices when faced with\\objectively equivalent\\alternatives
            }\\[2mm]
            \begin{tikzpicture}[font=\scriptsize]
                \def\yes{36}
                \wheelchart[
                    data=,
                    middle={\yes\%},
                    radius={0.5}{1},
                ]{
                \yes/yes-color/Yes,
                {100-\yes}/no-color/No
                }
            \end{tikzpicture}
        \end{minipage}
    \end{subfigure}
    \begin{subfigure}[b]{0.24\textwidth}
        \centering
        \begin{minipage}{\textwidth}
            \centering
            \parbox[c][20pt][c]{\textwidth}{%
                \centering\scriptsize
                Characteristic handling\\of the endgame
            }\\[2mm]
            \begin{tikzpicture}[font=\scriptsize]
                \def\yes{28}
                \wheelchart[
                    data=,
                    middle={\yes\%},
                    radius={0.5}{1},
                ]{
                \yes/yes-color/Yes,
                {100-\yes}/no-color/No
                }
            \end{tikzpicture}
        \end{minipage}
    \end{subfigure}\\[2mm]
    \begin{subfigure}[b]{0.24\textwidth}
        \centering
        \begin{minipage}{\textwidth}
            \centering
            \parbox[c][20pt][c]{\textwidth}{%
                \centering\scriptsize
                Typical risk management\\(readiness to sacrifice material\\vs tendency to simplify)
            }\\[2mm]
            \begin{tikzpicture}[font=\scriptsize]
                \def\yes{68}
                \wheelchart[
                    data=,
                    middle={\yes\%},
                    radius={0.5}{1},
                ]{
                \yes/yes-color/Yes,
                {100-\yes}/no-color/No
                }
            \end{tikzpicture}
        \end{minipage}
    \end{subfigure}
    \begin{subfigure}[b]{0.24\textwidth}
        \centering
        \begin{minipage}{\textwidth}
            \centering
            \parbox[c][20pt][c]{\textwidth}{%
                \centering\scriptsize
                Other
            }\\[2mm]
            \begin{tikzpicture}[font=\scriptsize]
                \def\yes{12}
                \wheelchart[
                    data=,
                    middle={\yes\%},
                    radius={0.5}{1},
                ]{
                \yes/yes-color/Yes,
                {100-\yes}/no-color/No
                }
            \end{tikzpicture}
        \end{minipage}
    \end{subfigure}
    \caption{\textbf{Perceived contribution of gameplay attributes to player recognizability.} The donut charts display the percentage of respondents who selected each given factor.}
    \label{fig:survey_recognizability_dims}
\end{figure}

The results in Figure~\ref{fig:survey_recognizability_dims} indicate that an inclination toward aggressive or defensive play is perceived as the most defining characteristics, being selected by 88\% of participants, respectively.
Typical risk management and preferred opening repertoires are also ranked highly, cited by 68\% and 60\% of the sample, respectively.
In contrast, other attributes were considered less significant; characteristic handling of the endgame was endorsed by only 28\%, while support for choices between objectively equivalent alternatives (36\%) was notably lower than expected (see~\cref{app:survey_style}).
The \quotes{Other} category, selected by 12\% of participants, captured a range of insightful points.
Some respondents leveraged this option to register a premise reject, arguing that recognizability is exceedingly difficult among today's universal top players.
Others pointed to more granular factors such as preferences for specific pawn structures, weaknesses in opening, middle, and end game.
Notably, some participants highlighted time management.
This last point is particularly salient; while we concur that decision speed is a powerful discriminative signal, the absence of move-timing information in the PGN datasets used in this work precluded its inclusion in our stylometry model.

\subsection{Existence and definition of style}
\label{app:survey_style}

Following the question of recognizability, we delve into the related but more fundamental concept of playing style, the existence of which remains a subject of controversy within the chess community.
One mindset posits that as players approach optimality, individual style dissolves into a universal pursuit of the objectively best move.
A telling case is Anatoly Karpov, who provocatly declared \quotes{\textit{Style? I have no style!},} a statement intended to underscore a commitment to pure objectivity.
Conversely, the opposing view argues that style is not a deviation from correct play, but rather a discernible pattern of preferences that emerges in complex positions where multiple viable continuations exist.
In forced positions with a single correct move, style has no space to manifest; it is in the majority of positions with multiple viable continuations that a player’s individuality comes to the fore.
Garry Kasparov, offered a paradoxical rebuttal to Karpov's claim, joking that \quotes{\textit{His style is precisely to have no style: his essence is to accept only those positions in which there are neither risks nor doubts.}}
Even Karpov, indeed, was nicknamed the \quotes{boa constrictor} for his recurring board states, and admitted to systematically favoring clear positional lines over tactical complications.
This aligns with the long-held idea that style is an expression of personality, as champion Rudolf Spielmann noted in the 1930s: \quotes{\textit{Show me your strategic principles in a game and I will tell you who you are.}}
This expression is nevertheless constrained by a player's practical abilities and shaped by subjective factors like personal taste (e.g., a kingside attack vs a central buildup) and psychological attitude toward risk.
The tension between style as an illusion negated by objective truth and style as a valid construct revealed through subjective choice is critical to our work.

\begin{surveyquestion}
There is no consensus on whether \quotes{style} truly exists in modern chess. Some grandmasters (e.g., Karpov) have claimed they had no style, only the pursuit of objectively best moves; others are consistently described as emblematic of a style. The debate revolves around whether style is an illusion or a legitimate construct. It is also important to distinguish between playing style---broad categories such as \quotes{attacking} or \quotes{positional}---and persona, the individual identity of a specific player. Two grandmasters may have very distinct personas while still being classified under the same style.

\textit{Do you acknowledge the existence of a playing style in chess, defined as a recurrent pattern of preferences in move selection, or do you believe only the search for the objectively best move matters?}
\end{surveyquestion}

\begin{figure}[!htb]
    \centering
    \begin{subfigure}[b]{0.3\textwidth}
        \centering
        \begin{minipage}{\textwidth}
            \centering
            \parbox[c][15pt][c]{\textwidth}{%
                \centering\scriptsize
                Yes,\\style exists and is identifiable
            }\\[2mm]
            \begin{tikzpicture}[font=\scriptsize]
                \def\yes{92}
                \wheelchart[
                    data=,
                    middle={\yes\%},
                    radius={0.5}{1},
                ]{
                \yes/yes-color/Yes,
                {100-\yes}/no-color/No
                }
            \end{tikzpicture}
        \end{minipage}
    \end{subfigure}
    \begin{subfigure}[b]{0.3\textwidth}
        \centering
        \begin{minipage}{\textwidth}
            \centering
            \parbox[c][15pt][c]{\textwidth}{%
                \centering\scriptsize
                No,\\only objective correctness matters
            }\\[2mm]
            \begin{tikzpicture}[font=\scriptsize]
                \def\yes{8}
                \wheelchart[
                    data=,
                    middle={\yes\%},
                    radius={0.5}{1},
                ]{
                \yes/yes-color/Yes,
                {100-\yes}/no-color/No
                }
            \end{tikzpicture}
        \end{minipage}
    \end{subfigure}
    \caption{\textbf{Expert consensus on the existence of playing style.} Binary question.}
    \label{fig:survey_style_existence}
\end{figure}

The results in Figure~\ref{fig:survey_style_existence} indicate a near-unanimous agreement among the sampled experts. 
An overwhelming 92\% of participants affirmed that style exists and is identifiable, recognizing differences in preferences and approaches between various players.

We proceed from the premise that style categories are not rigid, mutually exclusive labels but rather useful archetypes for characterizing a player's predominant tendencies.
Therefore, we presented our expert sample with a list of the most commonly accepted categories in chess literature.
Our goal was to test which of these are broadly considered valid, and to identify whether, in the perception of our respondents, any crucial descriptors were missing from our conventional taxonomy.

\begin{surveyquestion}
If one accepts that style exists in chess, the next challenge is defining and categorizing it. This is not straightforward: styles may overlap, and manifest differently across contexts.

\textit{Which of the following playing styles do you consider valid and useful categories?}
\end{surveyquestion}

\begin{figure}[!htb]
    \begin{subfigure}[b]{0.24\textwidth}
        \centering
        \begin{minipage}{\textwidth}
            \centering
            \parbox[c][30pt][c]{\textwidth}{%
                \centering\scriptsize
                Attacking/Tactical:\\Prefers dynamic complications, combinations, sacrifices, and direct assaults
            }\\[2mm]
            \begin{tikzpicture}[font=\scriptsize]
                \def\yes{79}
                \wheelchart[
                    data=,
                    middle={\yes\%},
                    radius={0.5}{1},
                ]{
                \yes/attacking-color/Yes,
                {100-\yes}/no-color/No
                }
            \end{tikzpicture}
        \end{minipage}
    \end{subfigure}
    \begin{subfigure}[b]{0.24\textwidth}
        \centering
        \begin{minipage}{\textwidth}
            \centering
            \parbox[c][30pt][c]{\textwidth}{%
                \centering\scriptsize
                Positional/Strategic:\\Prefers long-term accumulation of small advantages, spatial control, maneuvering
            }\\[2mm]
            \begin{tikzpicture}[font=\scriptsize]
                \def\yes{88}
                \wheelchart[
                    data=,
                    middle={\yes\%},
                    radius={0.5}{1},
                ]{
                \yes/positional-color/Yes,
                {100-\yes}/no-color/No
                }
            \end{tikzpicture}
        \end{minipage}
    \end{subfigure}
    \begin{subfigure}[b]{0.24\textwidth}
        \centering
        \begin{minipage}{\textwidth}
            \centering
            \parbox[c][30pt][c]{\textwidth}{%
                \centering\scriptsize
                Solid/Defensive:\\Avoids risk, emphasizes safety and prophylaxis, often aims to neutralize the opponent
            }\\[2mm]
            \begin{tikzpicture}[font=\scriptsize]
                \def\yes{75}
                \wheelchart[
                    data=,
                    middle={\yes\%},
                    radius={0.5}{1},
                ]{
                \yes/solid-color/Yes,
                {100-\yes}/no-color/No
                }
            \end{tikzpicture}
        \end{minipage}
    \end{subfigure}
    \begin{subfigure}[b]{0.24\textwidth}
        \centering
        \begin{minipage}{\textwidth}
            \centering
            \parbox[c][30pt][c]{\textwidth}{%
                \centering\scriptsize
                Creative/Unorthodox:\\Prefers surprising, original, or non-standard continuations even in quiet positions
            }\\[2mm]
            \begin{tikzpicture}[font=\scriptsize]
                \def\yes{71}
                \wheelchart[
                    data=,
                    middle={\yes\%},
                    radius={0.5}{1},
                ]{
                \yes/creative-color/Yes,
                {100-\yes}/no-color/No
                }
            \end{tikzpicture}
        \end{minipage}
    \end{subfigure}\\[2mm]
    \begin{subfigure}[b]{0.24\textwidth}
        \centering
        \begin{minipage}{\textwidth}
            \centering
            \parbox[c][10pt][c]{\textwidth}{%
                \centering\scriptsize
                Other
            }\\[2mm]
            \begin{tikzpicture}[font=\scriptsize]
                \def\yes{4}
                \wheelchart[
                    data=,
                    middle={\yes\%},
                    radius={0.5}{1},
                ]{
                \yes/other-color/Yes,
                {100-\yes}/no-color/No
                }
            \end{tikzpicture}
        \end{minipage}
    \end{subfigure}
    \caption{\textbf{Validation of conventional playing style categories.} The donut charts show the percentage of respondents who endorsed each of the proposed style categories as valid and useful.}
    \label{fig:survey_style_dims}
\end{figure}

Figure~\ref{fig:survey_style_dims} visually summarizes the results.
A strong consensus emerged on the validity of conventional style categories, with all four proposed archetypes---Attacking/Tactical (79\%), Positional/Strategic (88\%), Solid/Defensive (75\%), and Creative/Unorthodox (71\%)---being widely acknowledged as useful descriptors.
Although this confirms the utility of the conventional taxonomy, qualitative feedback from the \quotes{Other} category (4\%) offered a more nuanced argument.
This feedback suggested that more weight should be on the decision-making process rather than the outcomes, arguing that while any strong player can adopt any of the aforementioned styles given the necessity of the position, the true variation arises in how decisions are made.
This viewpoint suggests a shift from outcome-based categories to process-oriented ones, framing a player's identity in terms of their characteristic cognitive weighting.
If the game process is seen as a product of intuition and calculation, the source of difference between players is the respective weight given to each of these two components.
As illustrative examples, Gukesh and Ding Liren were cited as players who rely intensely on calculation, while Magnus Carlsen and Ian Nepomniachtchi were seen as relying more heavily on intuition.
This emphasis on the cognitive process directly echoes our earlier point regarding time management as a key dimension for player identification.
As noted previously, the time a player allocates to a move is a strong external indicator of their internal decision-making process—crucial information that, while unfortunately unavailable in common PGN datasets, should be a central consideration for future work in this area.

Expert human players demonstrate a remarkable ability to assess complex positions by recognizing abstract visual cues and harmonious piece structures, a skill closely linked to what is often termed \quotes{chess beauty.}
This form of pattern recognition operates on a different level than tactical calculation, relying on an intuitive grasp of a position's strategic potential which is hard to derive from symbolic move notations only.
Accordingly, beyond move sequences, a player's identity is often thought to manifest in the visual patterns they characteristically create on the board.
A positional player, for instance, might consistently produce games with harmonious piece structures and solid pawn chains, while a tactical player's games may be visually defined by dynamic imbalances and asymmetric configurations. 
To assess how salient this visual dimension is for our expert sample, we posed the following direct question.

\begin{surveyquestion}
\textit{Do you believe that visual patterns are important to recognize style?}
\end{surveyquestion}

\begin{figure}[!htb]
    \centering
    \begin{subfigure}[b]{0.25\textwidth}
        \centering
        \begin{minipage}{\textwidth}
            \centering
            \parbox[c][10pt][c]{\textwidth}{%
                \centering\scriptsize
                Yes
            }\\[2mm]
            \begin{tikzpicture}[font=\scriptsize]
                \def\yes{78}
                \wheelchart[
                    data=,
                    middle={\yes\%},
                    radius={0.5}{1},
                ]{
                \yes/yes-color/Yes,
                {100-\yes}/no-color/No
                }
            \end{tikzpicture}
        \end{minipage}
    \end{subfigure}
    \begin{subfigure}[b]{0.25\textwidth}
        \centering
        \begin{minipage}{\textwidth}
            \centering
            \parbox[c][10pt][c]{\textwidth}{%
                \centering\scriptsize
                No
            }\\[2mm]
            \begin{tikzpicture}[font=\scriptsize]
                \def\yes{22}
                \wheelchart[
                    data=,
                    middle={\yes\%},
                    radius={0.5}{1},
                ]{
                \yes/yes-color/Yes,
                {100-\yes}/no-color/No
                }
            \end{tikzpicture}
        \end{minipage}
    \end{subfigure}
    \caption{\textbf{Perceived importance of visual patterns in style recognition.} Binary question.}
    \label{fig:survey_style_visual}
\end{figure}

As shown in Figure~\ref{fig:survey_style_visual}, the majority (78\%) of respondents affirmed that visual patterns are salient for style recognition.
This finding suggests that for the expert community, a player's identity is not solely encoded in symbolic move sequences, but is also tangibly reflected in the characteristic board states and piece configurations they produce.
The reported agreement also offers a solid empirical justification for our decision to pioneer a vision-based behavioral stylometry model for chess.

\subsection{Style in grandmasters}

In chess culture, it is common to attribute a dominant style to the great champions of the past: consider the classic contrast between Mikhail Tal, the archetype of the tactical genius who created sacrificial attacks, and Tigran Petrosian, the emblem of prophylactic and defensive play.
However, such characterizations, while illustrative, are a simplification.
Elite players of any era possess a very broad repertoire, and as experts argue, speaking of \quotes{style} at the master level often amounts to highlighting a player’s preferences or strengths, but by no means implies they are incapable of excelling in other aspects of the game.
This complexity is further deepened by the fact that style is not necessarily a fixed trait.
Like any human characteristic, style can evolve with experience: some change their style during their career, others maintain their trademark.
This raises a particularly critical question about today's grandmasters, who are often described as \quotes{universal.}
We therefore sought to determine if our expert sample believes that even within this modern paradigm of all-around excellence, it is still possible to attribute a predominant style to a modern elite player.

\begin{surveyquestion}
At the elite level, players are often described as \quotes{universal,} capable of playing any type of position well. Yet many analysts argue that even such players retain a dominant style, recognizable across their careers, though it may evolve.

To what extent do you agree with the statement:\\
\textit{\quotes{Even a modern elite grandmaster, while being nearly universal, still exhibits a dominant playing style.}}
\end{surveyquestion}

\begin{figure}[!htb]
    \centering
    \includegraphics[width=.8\textwidth]{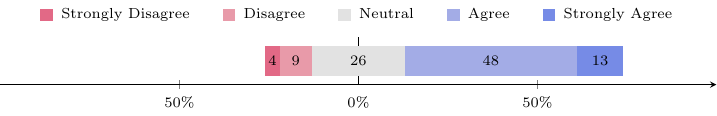}
    \caption{\textbf{Distribution of responses to the statement that modern elite grandmasters exhibit a dominant playing style.} The horizontal stacked bar represents the proportion of respondents on a five-point Likert scale (from Strongly Disagree to Strongly Agree).}
    \label{fig:survey_gm_has_style}
\end{figure}

The expert sample’s response to this question, detailed in Figure~\ref{fig:survey_gm_has_style}, indicates a prevailing, albeit not unanimous, belief in the persistence of a dominant style.
A 61\% majority of respondents affirmed this view, supporting the idea that a player's core tendencies remain identifiable even within a universal skill set.
The significant 26\% of neutral responses, with only a 13\% minority in outright disagreement, suggests that the primary source of contention is not whether a dominant style exists, but how to reconcile this concept with the acknowledged versatility of modern players.

To empirically test the practical implications of these beliefs, we transitioned from abstract opinion to a concrete labeling task.
We sought to determine whether the majority view---that dominant styles persist in modern grandmasters—is matched by a consistent ability among experts to apply such labels in practice.
Participants were therefore asked to assign the previously discussed style categories to each of the \gms who are the subjects of our computational analysis in the main paper.
This exercise serves to ground the theoretical discussion, allowing us to measure the degree of consensus that emerges when experts perform this practical classification.

\begin{surveyquestion}
To empirically ground the discussion, we ask respondents to attempt labeling specific contemporary grandmasters using the style categories introduced above. This helps test whether such labels are perceived as meaningful or not.

\textit{Please assign a dominant style to each of the following grandmasters. If you do not know the player well, or cannot attribute a dominant style, select the appropriate option.}
\end{surveyquestion}

\begin{figure}[!ht]
    \centering
    \begin{subfigure}[b]{3.2cm}
        \centering
        \begin{minipage}{3.2cm}
            \centering
            {\scriptsize\ding{182} V. Anand}\\[2mm]
            \includegraphics[width=1cm]{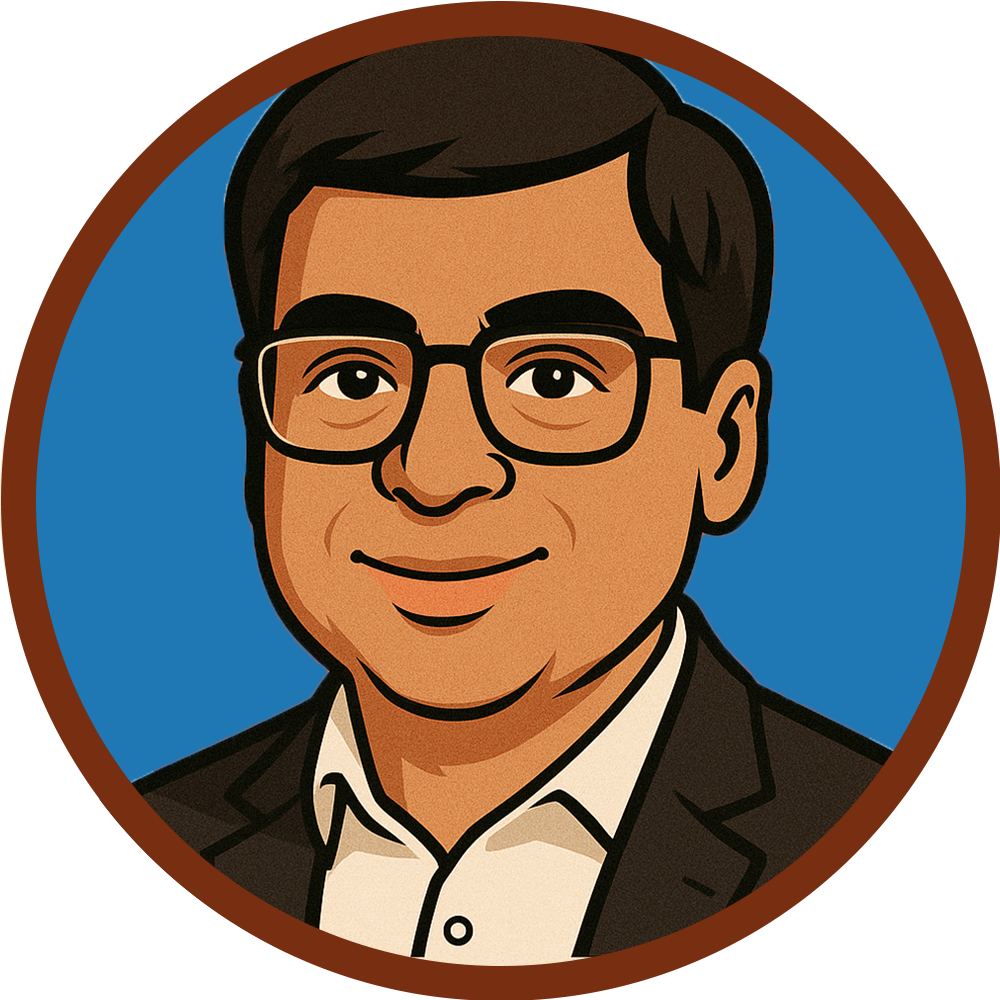}
        \end{minipage}\\[2mm]
        \begin{minipage}{3.2cm}
            \centering

        \end{minipage}
    \end{subfigure}
    \begin{subfigure}[b]{3.2cm}
        \centering
        \begin{minipage}{3.2cm}
            \centering
            {\scriptsize\ding{183} L. Aronian}\\[2mm]
            \includegraphics[width=1cm]{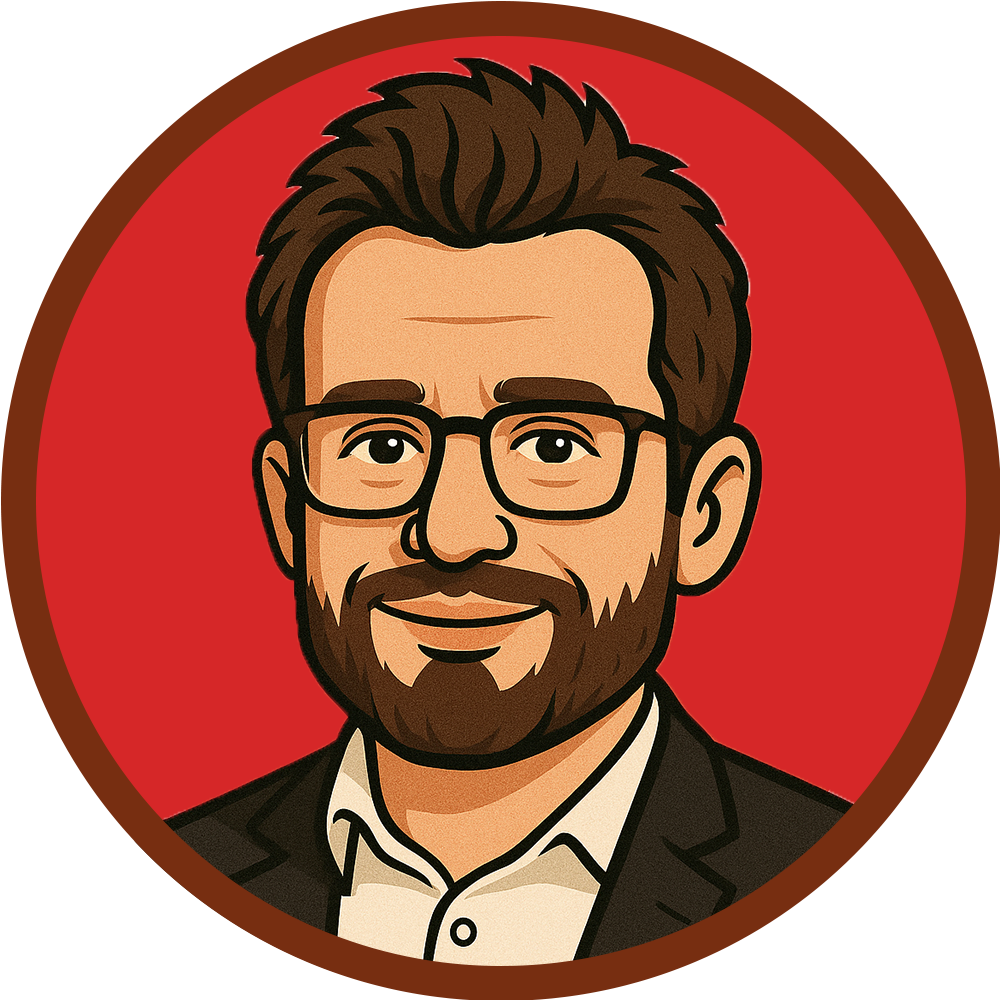}
        \end{minipage}\\[2mm]
        \begin{minipage}{3.2cm}
            \centering
            %
        \end{minipage}
    \end{subfigure}
    \begin{subfigure}[b]{3.2cm}
        \centering
        \begin{minipage}{3.2cm}
            \centering
            {\scriptsize\ding{184} M. Carlsen}\\[2mm]
            \includegraphics[width=1cm]{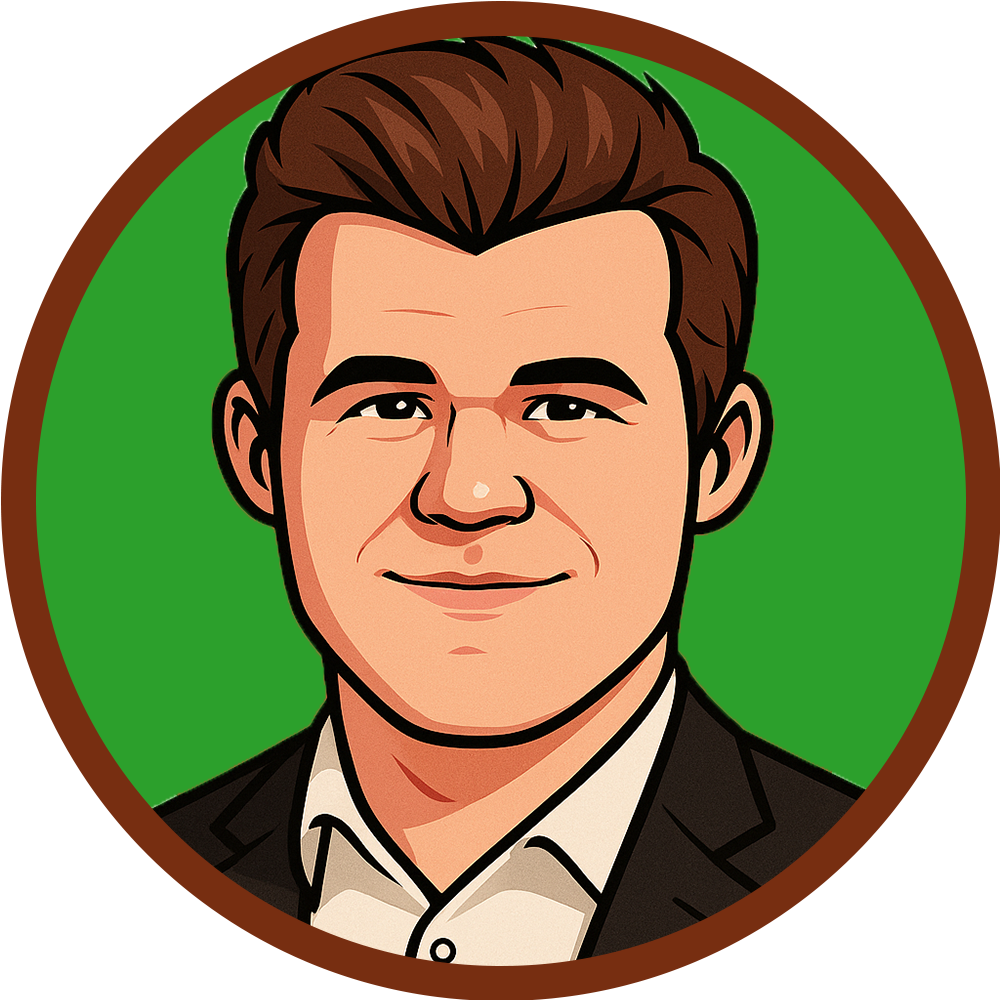}
        \end{minipage}\\[2mm]
        \begin{minipage}{3.2cm}
            \centering
            %
        \end{minipage}
    \end{subfigure}
    \begin{subfigure}[b]{3.2cm}
        \centering
        \begin{minipage}{3.2cm}
            \centering
            {\scriptsize\ding{185} F. Caruana}\\[2mm]
            \includegraphics[width=1cm]{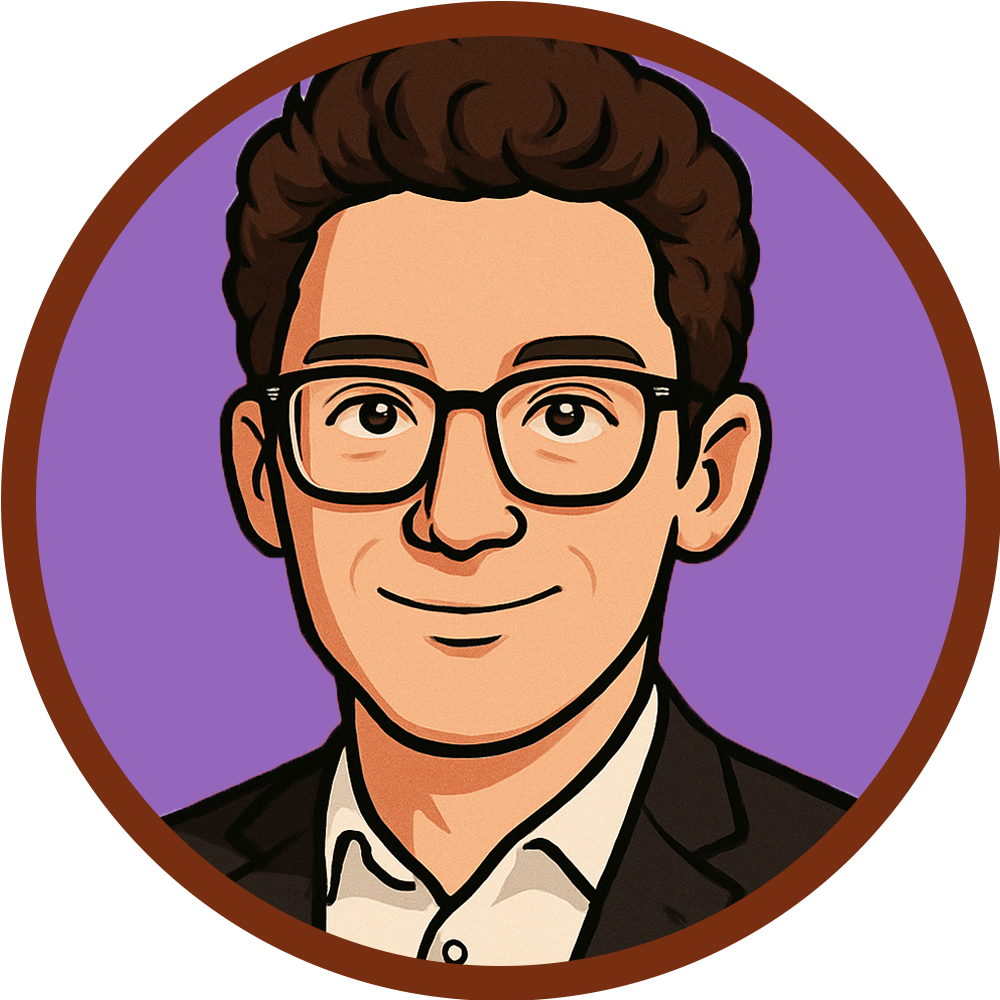}
        \end{minipage}\\[2mm]
        \begin{minipage}{3.2cm}
            \centering
            %
        \end{minipage}
    \end{subfigure}\\[3mm]
    \begin{subfigure}[b]{3.2cm}
        \centering
        \begin{minipage}{3.2cm}
            \centering
            {\scriptsize\ding{186} A. Firouzja}\\[2mm]
            \includegraphics[width=1cm]{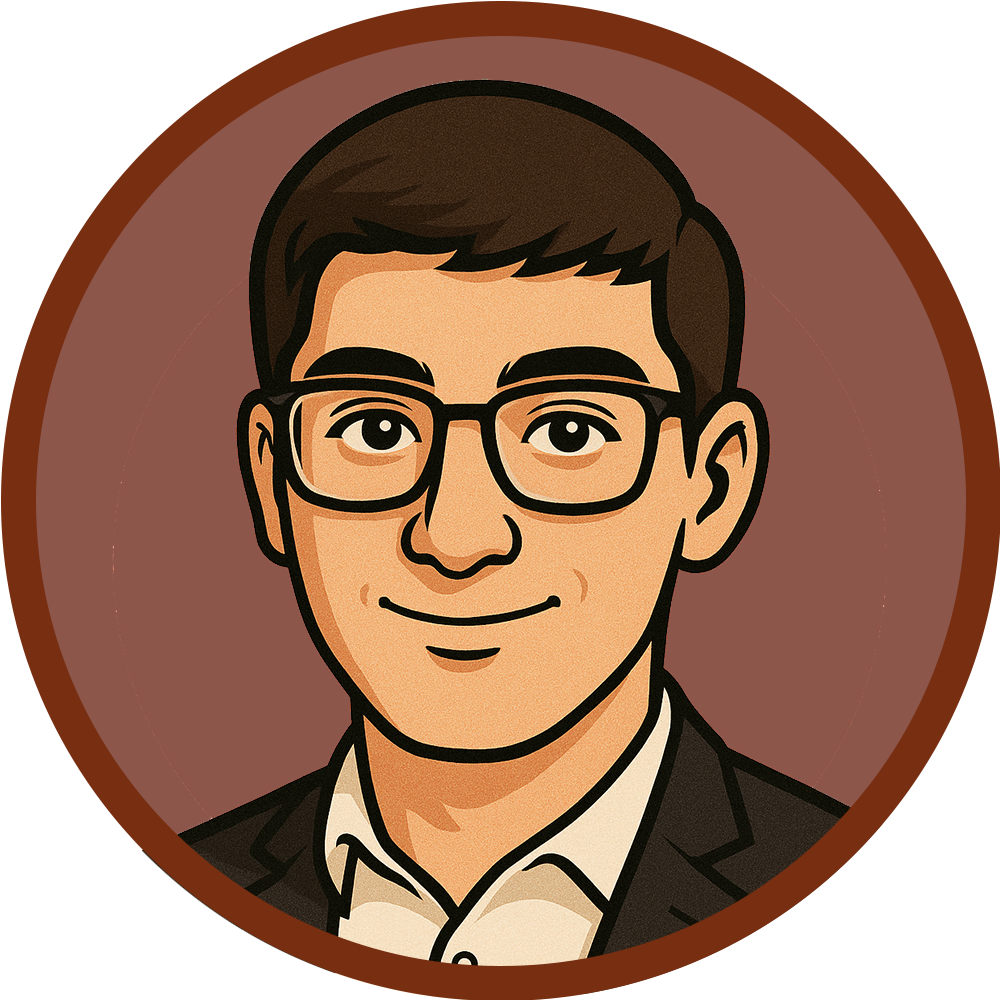}
        \end{minipage}\\[2mm]
        \begin{minipage}{3.2cm}
            \centering
            %
        \end{minipage}
    \end{subfigure}
    \begin{subfigure}[b]{3.2cm}
        \centering
        \begin{minipage}{3.2cm}
            \centering
            {\scriptsize\ding{187} A. Giri}\\[2mm]
            \includegraphics[width=1cm]{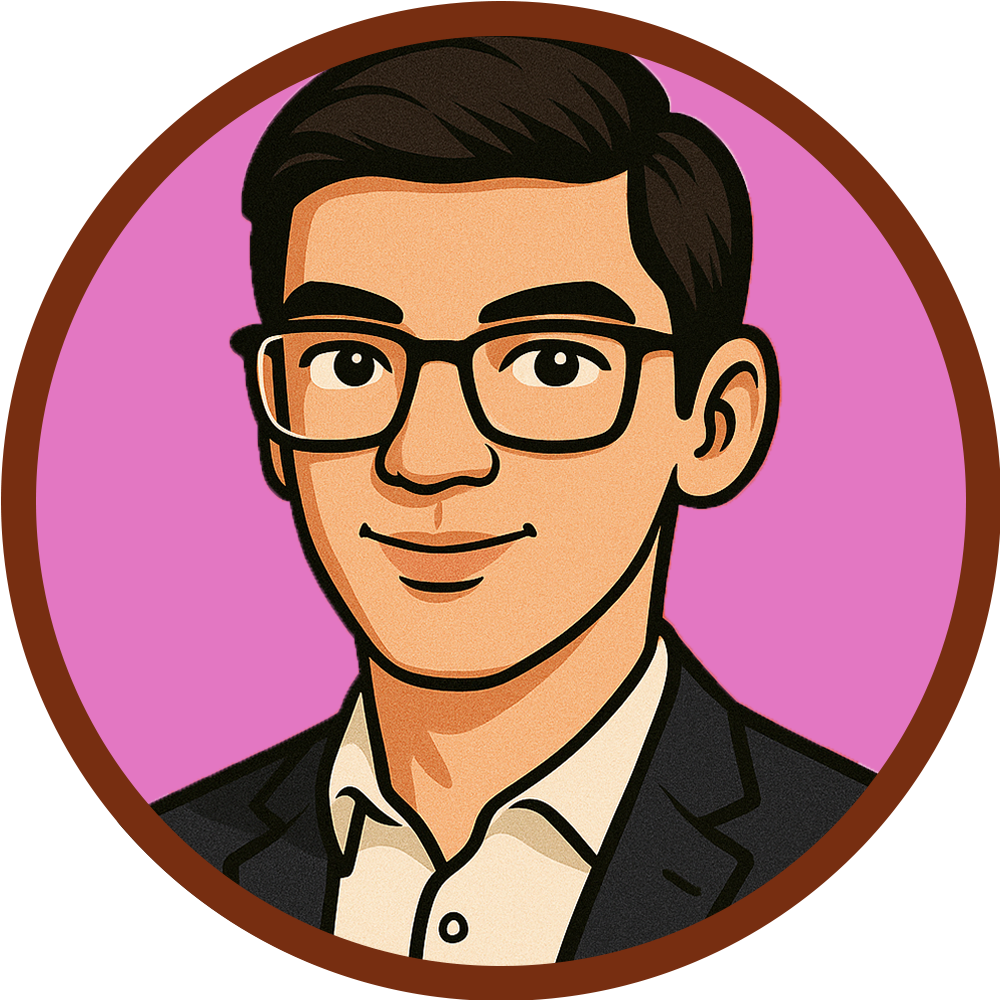}
        \end{minipage}\\[2mm]
        \begin{minipage}{3.2cm}
            \centering
            %
        \end{minipage}
    \end{subfigure}
    \begin{subfigure}[b]{3.2cm}
        \centering
        \begin{minipage}{3.2cm}
            \centering
            {\scriptsize\ding{188} H. Nakamura}\\[2mm]
            \includegraphics[width=1cm]{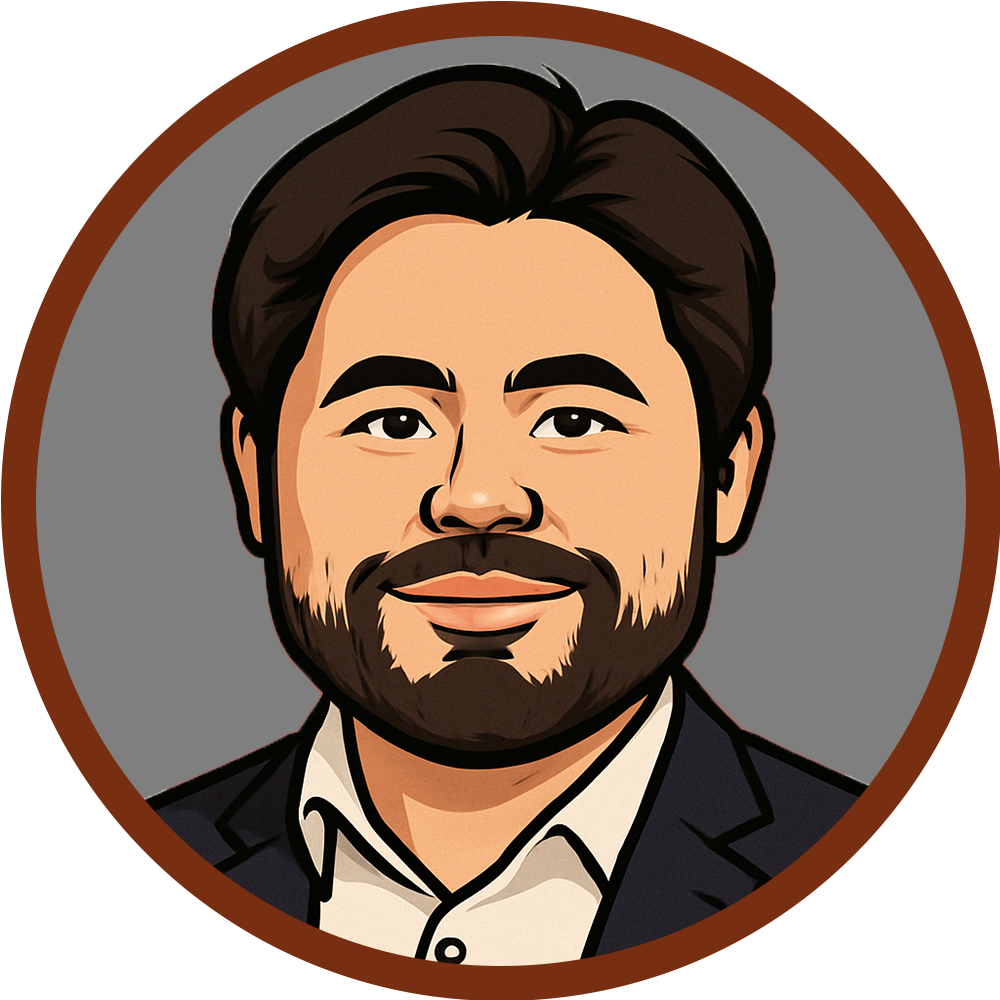}
        \end{minipage}\\[2mm]
        \begin{minipage}{3.2cm}
            \centering
            %
        \end{minipage}
    \end{subfigure}
    \begin{subfigure}[b]{3.2cm}
        \centering
        \begin{minipage}{3.2cm}
            \centering
            {\scriptsize\ding{189} I. Nepomniachtchi}\\[2mm]
            \includegraphics[width=1cm]{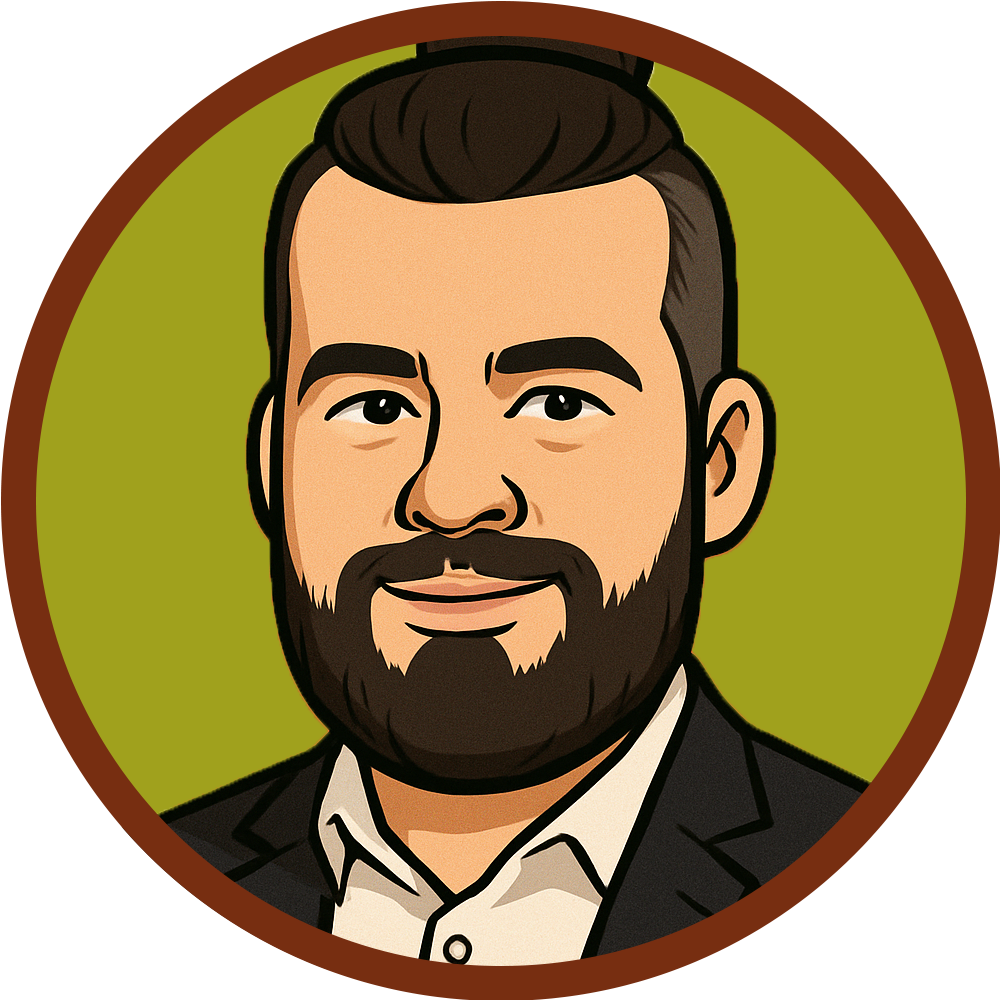}
        \end{minipage}\\[2mm]
        \begin{minipage}{3.2cm}
            \centering
            %
        \end{minipage}
    \end{subfigure}\\[3mm]
    \begin{subfigure}[b]{3.2cm}
        \centering
        \begin{minipage}{3.2cm}
            \centering
            {\scriptsize\ding{190} W. So}\\[2mm]
            \includegraphics[width=1cm]{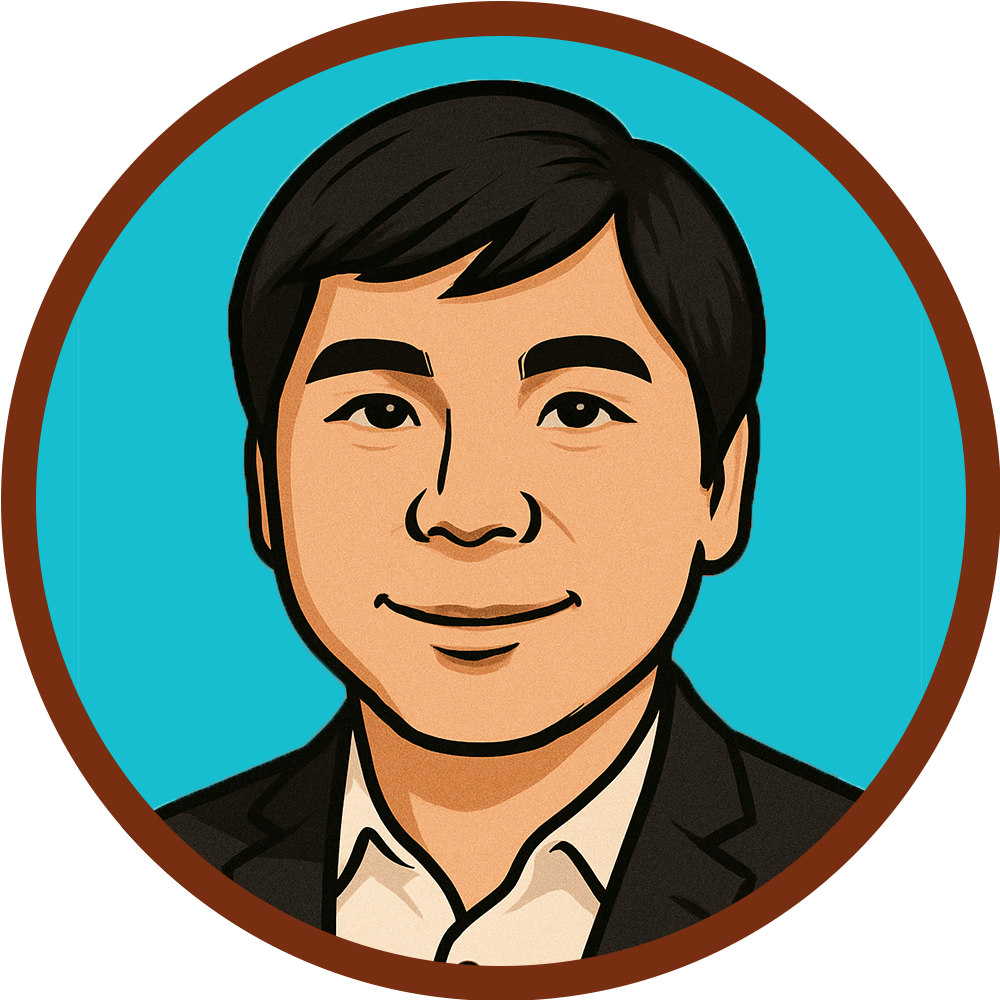}
        \end{minipage}\\[2mm]
        \begin{minipage}{3.2cm}
            \centering
            %
        \end{minipage}
    \end{subfigure}
    \begin{subfigure}[b]{3.2cm}
        \centering
        \begin{minipage}{3.2cm}
            \centering
            {\scriptsize\ding{191} M. Vachier-Lagrave}\\[2mm]
            \includegraphics[width=1cm]{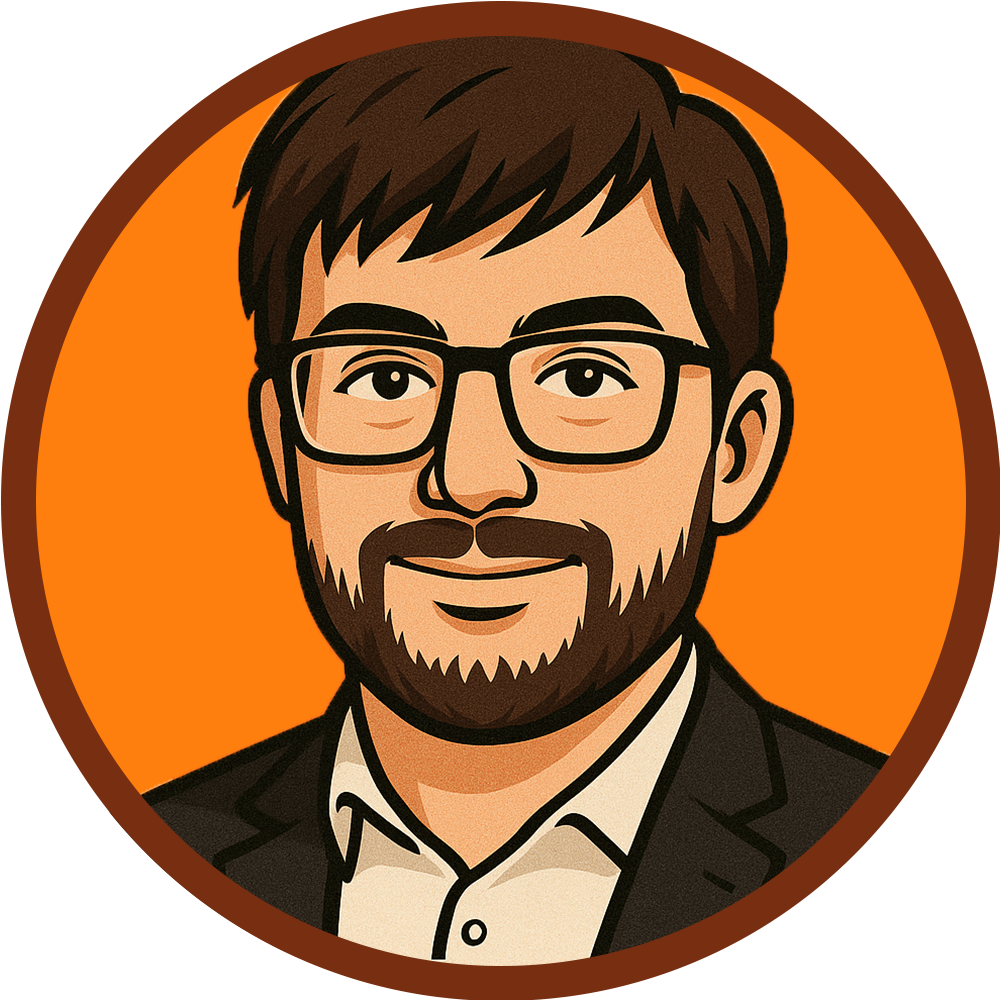}
        \end{minipage}\\[2mm]
        \begin{minipage}{3.2cm}
            \centering
            %
        \end{minipage}
    \end{subfigure}
    \caption{\textbf{Distribution of style category assignments for the ten grandmasters featured in this study.} Once choice per grandmaster. Each subplot displays the percentage of respondents assigning a dominant style category to a specific grandmaster.}
    \label{fig:survey_gm_style}
\end{figure}

The results of this practical labeling task, presented in Figure~\ref{fig:survey_gm_style}, underscore the inherent difficulty of assigning singular style categories to \gms.
This challenge is immediately apparent from the \quotes{Don't know (cannot assign)} option; it was selected for every grandmaster, representing 19\% of responses on average and peaking for Aronian (30\%).
When a choice was made within the four main style categories, high inter-annotator agreement was observed for only 4 out of the 10 \gms: Anand (attacking), Caruana (positional), Nakamura (attacking), and Vachier-Lagrave (attacking).
The remaining \gms received more fragmented and contrasting votes.
The \quotes{Other} option was leveraged by respondents to provide more specific characterizations.
For instance, Carlsen was described as \quotes{universal,} a label that transcends the given styles.
Similarly, Nepomniachtchi was defined through this option as a blend of \quotes{creative and aggressive.}
The use of this free-form option for such prominent players suggests that the conventional taxonomy, while broadly accepted, is sometimes perceived as insufficient to capture the identity of certain top \gms.

\subsection{Impact of chess engines and AI}

Our survey concludes by addressing a critical external determinant shaping the concepts discussed thus far: the impact of AI on human play.
The perceived rise of the \quotes{universal} player and the issues in applying stable style categories are often attributed to the ubiquitous use of chess engines in modern preparation.
A central concern within the community is that this technological reliance may be fostering a homogenization of play, eroding the expressive diversity that once defined different eras.
We therefore sought to determine whether our expert sample believes such a homogenization is occurring.

\begin{surveyquestion}
In contemporary practice, the systematic use of chess engines has become virtually indispensable for training, preparation, and post-game refinement. This has raised concern among players and scholars that such reliance may lead to a homogenization of playing behavior: players increasingly converge on the same engine-approved continuations, reducing the expressive diversity once observed across grandmasters of different schools or eras. This risk is considered even more pronounced in chess language models (CLMs). Unlike traditional search-based engines, which aim to compute the objectively best move, CLMs are trained to predict the most statistically likely next move from large corpora of historical games. In other words, they optimize for probability of occurrence rather than chess-theoretical correctness. By reflecting the aggregated tendencies of thousands of players of varied strength, such models might exacerbate stylistic flattening, reproducing the “average” move rather than preserving distinctive personas.

\textit{To what extent do you agree with the following statements:}
\end{surveyquestion}

\begin{figure}[!htb]
    \begin{tikzpicture}
    \begin{axis}[
      xbar stacked,
      width=7.3cm,
      height=3cm,
      scale only axis,
      trim axis left,
      trim axis right,
      xmin=-100,
      xmax=100,
      axis x line=bottom,
      xtick={-50, 50},
      xticklabels={50\%, 50\%},
      xticklabel style={font=\scriptsize},
      ytick=data,
      yticklabels={
        {Surprise and variability remain central to human play. The capacity to vary, to be unpredictable, or to switch between styles is considered crucial for practical and psychological reasons.},
        {In a hypothetical world where chess were solved and every player always chose the perfect move, the concept of style would lose its meaning.},
        {Intensive reliance on chess AI models has caused a flattening of stylistic diversity among players.},  
      },
      yticklabel style={align=right, font=\scriptsize, text width=6.4cm}, 
      enlarge y limits=0.22,
      bar width=14pt,
      legend style={
        at={(0.05,1.25)},
        anchor=north,
        legend columns=-1,        
        draw=none,                
        fill=none,
        font=\scriptsize,         
        /tikz/every odd column/.append style={column sep=2pt}, 
        /tikz/every even column/.append style={column sep=1.4em} 
      },
      legend image code/.code={
        \path[#1, draw=none] (-3pt,-3pt) rectangle (3pt,3pt);   
      },
      extra x ticks={0},
      extra x tick labels={0\%},
      extra x tick style={grid=major, tick style={draw=none}, major grid style={line width=0.9pt, draw=black}},
      nodes near coords,
      nodes near coords align={horizontal},
      nodes near coords style={anchor=center, yshift=0pt, font=\scriptsize, text=black},
      point meta=explicit symbolic,
    ]
    
    \addplot+[xbar, draw=none, fill=neutral-color, forget plot] coordinates {(0,0) (0,1) (0,2)};  
    \addplot+[xbar, draw=none, fill=disagree-color, forget plot] coordinates {(0,0) (-10,1)[10] (-20,2)[20]};   
    \addplot+[xbar, draw=none, fill=strongly-disagree-color, forget plot] coordinates {(0,0) (-10,1)[10] (-10,2)[10]};  
    
    \addplot+[xbar, draw=none, fill=neutral-color, forget plot] coordinates {(0,0) (0,1) (0,2)}; 
    \addplot+[xbar, draw=none, fill=agree-color, forget plot] coordinates {(50,0)[50] (10,1)[10] (60,2)[60]};   
    \addplot+[xbar, draw=none, fill=strongly-agree-color, forget plot] coordinates {(50,0)[50] (70,1)[70] (0,2)[10]};  
    
    \addlegendimage{fill=strongly-disagree-color}
    \addlegendentry{Strongly Disagree}
    \addlegendimage{fill=disagree-color}
    \addlegendentry{Disagree}
    \addlegendimage{fill=neutral-color}
    \addlegendentry{Neutral}
    \addlegendimage{fill=agree-color}
    \addlegendentry{Agree}
    \addlegendimage{fill=strongly-agree-color}
    \addlegendentry{Strongly Agree}
    
    \end{axis}
    \end{tikzpicture}
    \caption{\textbf{Perceptions of AI's impact on stylistic diversity and the enduring importance of human variability.} Each horizontal stacked bar represents the proportion of respondents on a five-point Likert scale (from Strongly Disagree to Strongly Agree).}
    \label{fig:survey_chess_ai}
\end{figure}

The results in Figure~\ref{fig:survey_chess_ai} confirm the widely held concern that motivates our paper: a strong majority (60\%) of respondents agree that intensive reliance on AI has caused a flattening of stylistic diversity among players.
This perceived homogenization is particularly noteworthy when contextualized by the second finding, where 70\% of respondents affirmed that style would lose its meaning in a hypothetically \quotes{solved} version of chess.
This result conceptually tethers style to the existence of meaningful human choice and imperfection.
In a powerful counterpoint, however, there was unanimous agreement (100\%) that surprise and variability remain crucial for practical and psychological reasons in human-to-human play.

\clearpage
\newpage

\section{Extended related work}
\label{app:related_work}

\subsection{Chess and AI}
Abstract games serve as a proxy for real-world skills, allowing the evaluation of a model's capacities in strategic planning and reasoning, memory, and adaptive learning, as well as theory of mind through the inference of an opponent's intent.
Chess is a landmark planning problem in AI research, distinguished by a rich history, extensive data corpora, and active community participation.

\paragraph{Traditional engines}
Early computer chess relies on heuristic-based techniques, as exemplified by Turing's initial explorations~\cite{burt1955faster} and implementations like NeuroChess~\cite{DBLP:conf/nips/Thrun94a}.
This paradigm culminates in Deep Blue~\cite{DBLP:journals/ai/CampbellHH02} and early versions of Stockfish~\cite{romstad2008stockfish}, which combine large-scale tree search with manually designed evaluation functions.
Specifically, these engines employ minimax search with alpha-beta pruning, exploring millions of positions per move to maximize winning probability through exhaustive lookahead.
Although most modern engines retain this search-and-evaluation framework, they have progressively replaced static evaluation with neural networks.
In this context, AlphaZero~\cite{DBLP:journals/corr/abs-1712-01815} marks a major milestone.
It learns to play chess solely through RL from repeated self-play, using a convolution-based residual network to evaluate board positions and to guide Monte Carlo Tree Search (MCTS) in selecting moves.
The open-source Leela Chess Zero~\cite{lczero2018leelachesszero} recreates this approach, incorporating improvements over time, including support for multiple hardware backends, opening rule variants, and transformer-based architectures.
Crucially, AlphaZero and Leela Chess Zero assume access to an exact simulator of the environment (i.e., the full game rules), which is used to generate future states during search.
This line of work is further generalized by MuZero~\cite{DBLP:journals/nature/SchrittwieserAH20}, which removes this assumption by learning a latent model of environment dynamics directly from experience, and is successfully applied beyond chess to a wider range of domains including Go, Shogi, and Atari games.
Instead of relying on a known transition function, MuZero jointly learns representation, dynamics, and prediction components, enabling planning via MCTS in an internally constructed state space while still achieving strong performance.

\paragraph{Language models}
More recently, chess has been reformulated as a sequence modeling problem because of its text-archived nature.
Unlike natural language, chess notation describes a simple, constrained, and deterministic domain with a well-defined transition function and legal move space.
A fundamental distinction arises between traditional chess engines and language models.
Search-based and RL engines are explicitly optimized to maximize game outcomes: their architectures incorporate deep lookahead, evaluation functions, pruning strategies, and transposition tables, all tuned to achieve optimal play.
In contrast, autoregressive language models are generally trained to maximize the likelihood of observed move sequences, rather than win per se.
Their architecture is comparatively simple, dispensing with search procedures and evaluation functions in favor of statistical pattern learning, which makes them easier to train and adapt.\\
\underline{Self-supervised models}.
\cite{DBLP:journals/corr/abs-2008-04057} are among the first to observe that fine-tuned GPT-2 models, under an SSL regime, can generate meaningful legal moves and strategically plausible continuations.
Subsequent work demonstrates that even a vanilla GPT model with 50M parameters only, trained from scratch on a few million game transcripts, can achieve a legal move rate of 99.8\% and $\sim$1,300 Elo---without signs of memorization~\cite{DBLP:journals/corr/abs-2403-15498}.
Notably, \cite{DBLP:conf/nips/ZhangZSKETKM24} provide evidence that self-supervised generative models can attain Elo ratings exceeding those of any player in their training corpus---a phenomenon referred to as transcendence.\\
\underline{Supervised value prediction models}.
To maximize performance, \cite{DBLP:conf/nips/RuossDMGLCRLVG24} depart from SSL and instead distill Stockfish into large-scale decoder-only transformers via supervised learning on engine-crafted annotations, reaching 2,895 Lichess blitz Elo against human opponents.
Their models (up to 270M parameters) are trained on 10M chess games annotated by Stockfish 16 to predict action-values given a board state, where each legal move is treated as an action and its value represents the expected return or quality of that move.
In parallel, \cite{DBLP:journals/corr/abs-2409-12272} (Leela Chess Zero team) introduce Chessformer, an encoder-only transformer architecture with chess-specific optimizations for action-value estimation.
After supervised training on AlphaZero self-play data, models (up to 240M) further surpass Ruoss et al. baselines while using fewer FLOPs.
\cite{DBLP:conf/naacl/ZhangHLCL25} fine-tune Open-LLaMA-3B to generate the best move in Standard Algebraic Notation (SAN) from a given board state in Forsyth-Edwards Notation (FEN).
By annotating training data with Stockfish and leveraging high-depth searches in its alpha-beta tree, they achieve an Elo rating of 1,788.\\
\underline{RL models}.
The intersection of chess and RL, particularly GRPO, remains under-explored, with most efforts centering on reasoning LLMs that output situation analyses other than suggested moves.
\cite{DBLP:journals/corr/abs-2507-12215} fine-tune Qwen-2.5-7B-Instruct with GRPO for Xiangqi (Chinese chess), using combined PGN and FEN inputs alongside multi-dimensional rewards designed to improve both output format and engine-evaluated quality.
\cite{DBLP:journals/corr/abs-2507-00726} apply GRPO to fine-tune Qwen-2.5 and LLaMA-3.1 models for chess puzzle solving, where the input representation is restricted to FEN and the reward signal comes solely from engine-derived post-move win probabilities.\\
\underline{Task diversification}.
Language models expand the scope of chess AI beyond next move prediction, powering rule induction~\cite{DBLP:journals/corr/abs-2209-11902,DBLP:conf/ranlp/Stockl21}, move quality assessment~\cite{DBLP:journals/corr/abs-1907-08321}, state tracking~\cite{DBLP:conf/icml/MerrillPS24,DBLP:conf/aaai/ToshniwalWLG22}, vision-based playing~\cite{DBLP:journals/corr/abs-2304-14918}, commentary generation~\cite{DBLP:journals/corr/abs-2212-08195}, and auxiliary generative tasks~\cite{DBLP:conf/nips/FengLWTYSM0W23}.
Chess-based tasks (e.g., find-legal-moves, checkmate-in-one) are incorporated into evaluation suites such as BIG-bench~\cite{DBLP:journals/tmlr/SrivastavaRRSAF23} to probe the zero-shot reasoning and planning capabilities of LLMs.
As expected, chess puzzle solving has become a widely adopted evaluation task, in which the objective is to identify the best move from a given position, typically specified in FEN.
In this setting, \cite{carlini2023playing} show that performance improves substantially when models are provided also with the full PGN history.
This result highlights that language models rely on trajectory-level context rather than purely on the current state.
Furthermore, models implicitly adapt to the quality of the observed move history: sequences containing weak or implausible moves bias predictions toward lower-quality continuations, whereas sequences reflecting strong play steer the model toward more accurate and principled decisions.\\
\underline{Input format}.
Exclusive use of FEN is typical only for engine distillation procedures~\cite{DBLP:journals/corr/abs-2409-12272,DBLP:conf/nips/RuossDMGLCRLVG24,DBLP:conf/naacl/ZhangHLCL25}, where the focus is on evaluating static states or targeting Chess960 puzzles that randomize the back-rank starting position.
When the goal is to model full games move by move, the progressive history in PGN format becomes essential.
Kaggle's Game Arena, in partnership with Google DeepMind, hosted a text-only chess competition in which commercial LLMs compete by generating moves from prompts that include both FEN and PGN represenations, without access to legal move lists or external engines.\footnote[7]{\url{https://www.kaggle.com/benchmarks/kaggle/chess-text/tournament}}
Fine-tuning on chess textbooks, commentary, and tactical calculations also proves effective~\cite{DBLP:journals/corr/abs-2310-20260,DBLP:conf/nips/FengLWTYSM0W23,DBLP:conf/naacl/WangJWZLHW25}, giving the model both move sequences and explanatory texts.

\paragraph{Mechanistic interpretability}
AI successes spark a research agenda centered on interpreting superhuman models by probing their internal representations.
For AlphaZero, \cite{DBLP:journals/corr/abs-2111-09259} show that a self-play agent develops internal representations aligned with human chess knowledge: linear probes can recover a broad range of chess concepts from network activations, and the emergence of these concepts can be tracked across training stages and layers.
Complementarily, \cite{DBLP:conf/nips/JennerKGAER24} provide evidence of learned look-ahead in Leela Chess Zero, showing that its policy network internally represents future optimal moves.
In language models, \cite{DBLP:journals/corr/abs-2403-15498} show that GPT-style models trained on PGN transcripts acquire latent board-state representations: simple linear probes can decode piece locations from activations, and the model also estimates latent variables such as player strength.
These findings suggest that strong chess models do not merely memorize surface-level move statistics, but develop structured internal representations of board state, tactical consequences, and player-dependent context.

Our work builds on pure autoregressive, small chess language models for next move prediction--taking PGN as input and operating without action-value estimation or external search mechanisms.
We focus on traditional SSL; in \cref{app:rl}, we provide an ablative training study investigating the capacity of GRPO to improve legal play without external supervision.
To our knowledge, \mom represents the strongest result reported within this model class and scale (i.e., under 200M parameters).

\subsection{Expert merging}
\paragraph{General-domain}
Growing evidence suggests that diversity can be more valuable than strength alone.
In modern language models, MoE architectures instantiate this principle through sparse routing: only a subset of experts is activated per input, allowing total capacity to scale without proportional compute.
These models have shown advantages in settings where distinct capabilities must be preserved.
In multi-task learning, \cite{DBLP:journals/corr/abs-2509-07945} show that routing enables the model to maintain specialized competencies by mitigating gradient interference, effectively preserving distinct behaviors within a single system.
However, experts are typically learned jointly from broad data mixtures, and specialization emerges implicitly from routing and optimization rather than from explicit semantic design.
Routing patterns are often analyzed post hoc and experts are rarely assigned a human-interpretable identity prior to training.
\cite{wang2026mythexpertspecializationmoes} argue that observed expert behavior often reflects structure in the latent representation space rather than coherent functional roles and domain expertise, raising questions about interpretability.
In response, recent work explores alternative ways to construct and combine experts.
\cite{DBLP:conf/icml/MuqeethLLR24} model experts as parameter-efficient adapters, enabling targeted specialization while maintaining a shared backbone, and reinforcing the view of experts as composable functional units rather than emergent partitions.
Branch-Train-Merge~\cite{DBLP:journals/corr/abs-2208-03306} departs from classical MoE by removing routing entirely: it trains independent expert language models on disjoint data partitions and merges them post hoc into a single dense model (i.e., inference-time ensembling).
Their results reveal that semantically meaningful partitioning (e.g., by domain) is critical, as random splits degrade performance.
Similarly, BTS~\cite{DBLP:conf/emnlp/ZhangBBCFFKSSDGL25} maintains a seed LLM and domain-specialized expert copies frozen, and trains only lightweight stitch layers between them, preserving modularity, flexibility, and interpretability.
\paragraph{Boardgames}
In sequential decision-making, a recurring strategy is to decompose a complex task into multiple specialized policies and combine them, rather than relying on a single monolithic model.
\cite{DBLP:conf/aaaiss/DobreL17} follow this approach in Settlers of Catan by training separate feed-forward policy networks on state--action data corresponding to different phases of the game.
Although the partitioning is predefined and does not rely on learned routing, the resulting experts capture complementary strategies associated with different game contexts, improving performance over a single model.
Similarly, heterogeneous teams of Go agents outperform both solitary agents~\cite{DBLP:conf/icml/KakadeL02} and homogeneous teams~\cite{DBLP:conf/aaai/MarcolinoXJTB14}, showing that combining diverse policies can improve robustness and strategic coverage.
In the chess domain, \cite{DBLP:journals/corr/abs-2401-16852} introduced a MoE framework combining phase-specialized neural networks (opening, middlegame, endgame) with MCTS, demonstrating that modular specialization can improve performance when integrated into search.

Our work explores the first MoE-based chess language model where experts are \gm personas.
In line with BTS, our experts are independently trained specialized copies of a shared GPT chess language model, rather than jointly optimized subnetworks.
We alternate weight merging with dynamic routing, enabling the model to both consolidate shared knowledge and selectively activate persona-specific experts at inference time.
The goal is to investigate whether persona-based specialization yields measurable and interpretable gains over both dense models and randomly partitioned sparse alternatives, while preserving stylistic diversity.

\subsection{Human-AI alignment in chess}
Humans engage with chess AI both as competitors and as training partners.
This motivates research aimed at predicting the moves \textit{some} humans are likely to make, rather than those that are strictly optimal.
Interacting with bots exhibiting contrasting styles helps users recognize and respond to diverse strategies, exercising cognitive flexibility.
Commercial products like Play Magnus and Chess.com's bots are player-personalized, though their methods remain undisclosed.
In open research, \cite{DBLP:conf/kdd/McIlroy-Young0K20} introduce Maia, a supervised adaptation of Leela Chess Zero that predicts moves of average human players at specific rating levels, with separate models trained per Elo band.
Maia-2 extends this approach to an efficient and unified model using skill-aware attention~\cite{DBLP:conf/nips/TangJMK0A24}.
This perspective naturally leads to the study of \textit{behavioral stylometry}, i.e., the identification of players based solely on their move patterns.
A central insight is that chess players exhibit behavioral fingerprints: consistent tendencies in how they approach similar positions, which persist beyond the objective strength of their moves.
These fingerprints can arise from a combination of factors, including opening repertoire preferences, risk tolerance, piece coordination habits, and characteristic responses to recurring positional structures.
This suggests that neural networks trained on game data can, in principle, learn to associate sequences of moves with their authors by capturing these latent behavioral signatures.
In \cite{DBLP:conf/kdd/McIlroy-YoungW022}, a population-level Maia-1 model (Maia-1900) is fine-tuned into 400 independent personalized models, each trained on the games of a specific Lichess player (Elo 1,000--2,000).
Each personalized model predicts moves in the style of its target player, and behavioral stylometry is enabled as a byproduct.
Given a set of query games, identification is performed by evaluating all candidate models and selecting the one that best predicts the observed moves, measured in terms of move prediction accuracy (i.e., likelihood of the played moves).
In practice, personalized models better match the move choices of the target player than the population-level model only when at least $\sim$5,000 games are available for that player, while smaller datasets (e.g., 1,000 games) are insufficient or even detrimental.
\cite{DBLP:journals/corr/abs-2502-14998} adopt a different generative formulation based on shared parameterization.
Instead of training one model per player, they train a single backbone model on all players' data and freeze it, then add a fixed inventory of shared LoRA adapters that are trained jointly across players.
These adapters are not assigned to specific players nor trained with separate objectives; rather, they are optimized collectively to improve move prediction over the full multi-player dataset, so they do not carry predefined semantic meaning.
Each player is represented by a style vector that defines a distribution over this shared adapter inventory, effectively selecting and combining low-rank weight updates during inference.
This contrasts with \cite{DBLP:conf/kdd/McIlroy-YoungW022}, where each player has a fully independent model trained on their own data, whereas here players share all parameters except their routing vector.
Given a set of query games with unknown identity, the backbone and shared adapters are kept fixed, and a new style vector is optimized to maximize the likelihood of the observed moves (i.e., best reproducing the query behavior).
Identification is then performed by comparing this inferred vector to the stored player-specific vectors using cosine similarity, and selecting the closest match.
The framework is evaluated across multiple domains, including chess, Rocket League, and image generation.
For chess, the backbone builds on a Maia-like convolutional residual network, the number of adapters per layer is 32, and training uses $\sim$244M games from 47.8K Lichess players (primarily amateur).
Complementarily, \cite{DBLP:conf/nips/McIlroy-YoungWS21} propose a non-generative stylometry approach based on a transformer-style encoder trained from scratch.
Move-level features are extracted from human-engineered 3D tensors, aggregated into game vectors by a ViT-inspired transformer encoder, and then averaged into player centroids.
At inference time, a query player is identified by comparing the embedding of their game samples to a pool of candidate player centroids.
The authors train their main model on 63.7M games from 16,181 amateur Lichess players (Elo 1,000--2,000), and report substantial performance degradation when restricting the training set to only a few hundred players or when considering higher-rated players, whose behavior is more homogeneous.

We do not target a large-scale identification benchmark.
Identifying a player from a pool of thousands of candidates lies outside the scope of this work.
Instead, our goal is to verify whether persona experts have acquired the distinctive playing signatures of their target \gms, and whether those signatures are preserved when the experts are combined within \mom.
In our paper, stylometry operates exclusively as a post-hoc diagnostic tool.
The \mom architecture---including expert training, weight merging, and gating---functions entirely independently of any stylometric signal.
Similarly, stylometry does not replace the strength-based evaluation against Stockfish.
It serves as a complementary lens for inspecting the structure of the learned representations and check for the absence of homogenization.
To implement this diagnostic tool, we go beyond measuring each expert's capacity to replicate the moves of its target \gm on held-out games, as done in \cite{DBLP:conf/kdd/McIlroy-YoungW022}.
We additionally analyze activation geometry, asking where persona-specific fine-tuning produces functional separation inside the experts and whether this separation remains visible after composition in the sparse \mom model.
This perspective is aligned with recent interpretability work in dense language modeling~\cite{DBLP:journals/corr/abs-2602-02464}.
In \cref{app:behavioral_stylometry}, we investigate alternative implementations of this diagnostic framework and introduce a novel model-based behavioral stylometry metric, inspired by~\cite{DBLP:conf/nips/McIlroy-YoungWS21}. We then show empirically that these approaches fail to consistently distinguish the stylistic signatures they are intended to validate, which motivates their omission from the main analysis.







\clearpage
\newpage

\section{Data details}
\label{app:data}

Datasets are constructed by merging PGN files from three sources of games: \textit{PGNMentor},\footnote[8]{\url{https://www.pgnmentor.com/}} a chess archive with $>$1M games mainly from notable players and over-the-board tournaments, curated for historical and instructional value; \textit{Chess.com}\footnote[9]{\url{https://www.chess.com/games}.} and \textit{Lichess},\footnote[10]{\url{https://huggingface.co/datasets/Lichess/tournament-chess-games}} the largest online chess platforms.
PGNMentor and Chess.com publicly release large collections of games for free download, and their data are widely used in academic research~\cite{DBLP:conf/acis/AdnanGXHK24,DBLP:conf/waw/BonatoW25,DBLP:journals/paapp/BurduliW23}.
Lichess releases its data under the Creative Commons CC0 license.

The data described in this section supports two distinct experimental threads.
\cref{subsec:data_lm} covers the construction of training and evaluation data for the core language modeling contribution, where the objective is to acquire play strength and internalize \gm-specific stylistic patterns via next-move prediction.
\cref{subsec:data_stylometry} describes the data used to construct and validate the stylometry diagnostic tools external to \mom, which are designed to predict player identity from observed move sequences.

\subsection{Language modeling}
\label{subsec:data_lm}
We follow established best practices for the filtering of chess game corpora suitable for language modeling training and evaluation~\cite{DBLP:journals/corr/abs-2401-16852,DBLP:conf/kdd/McIlroy-YoungW022}, and substantially extend them through a principled, multi-stage data preparation pipeline.
\begin{enumerate}
    \item \textit{Variant.} We retain only games from the standard chess variant, excluding Chess960 and games initialized from non-standard positions. Including variant rules or arbitrary initial positions would introduce heterogeneity in the generation of legal moves and board priors, thus degrading the model's ability to learn a coherent policy in a well-defined game manifold.
    \item \textit{Time control.} Games span a wide range of time controls, from classical formats to ultra-fast bullet games. When clock annotations are available, we exclude the fastest games, specifically those in which each player has 15 seconds for the entire game. Such games exhibit substantially elevated error rates due to extreme time constraints, where decision-making is dominated by reaction speed rather than strategic evaluation. In these regimes, players frequently employ heuristics, premoves, or intentionally suboptimal moves to avoid time forfeits, resulting in trajectories that deviate systematically from principled play. Including these samples would introduce high-variance, low-quality supervision signals and bias the learned distribution toward non-representative, time-induced behaviors, thereby degrading both policy learning and the fidelity of stylistic modeling. We acknowledge that per-move time usage constitutes an informative signal for capturing player behavior. However, such fine-grained temporal annotations are not available across the considered datasets. Incorporating move-level time information represents a promising direction for future work.
    \item \textit{Outcome.} We remove games with undefined or missing results (e.g., marked as \quotes{*}). These entries are typically truncated due to external factors (e.g., disconnections or aborts) rather than arising from the intrinsic game dynamics and therefore do not correspond to valid samples from the target sequence distribution. Including such sequences can bias the likelihood estimation and degrade the model’s ability to learn coherent termination behavior.
    \item \textit{PGN normalization.} All PGN strings are canonicalized by removing comments and quality glyphs (e.g., \quotes{?!}, \quotes{!!}). We further discard entries with malformed PGN formatting. This step enforces a clean and uniform token space.
    \item \textit{Minimum game length.} We remove games shorter than 5 moves. Such games are typically low-information samples arising from pre-arranged draws. From a sequence modeling perspective, they contribute negligible learning signal while disproportionately affecting length distributions and early-game priors.
    \item \textit{Source merging and deduplication.} We merge data from all sources and eliminate duplicate entries based on identical PGN strings, \gm identity, and color assignment. When duplicates are detected across sources, we prioritize entries from Lichess due to their richer metadata.
    \item \textit{Color balancing.} Within each \gm's collection, we enforce color balance by downsampling the overrepresented color (White or Black) to match the underrepresented one. This controls for known asymmetries in opening distributions and outcome statistics associated with color, preventing the model from conflating stylistic patterns with color-dependent priors.
    \item \textit{Mate completion augmentation.} \gm games frequently terminate via resignation before an explicit checkmate occurs, leading to underrepresentation of terminal mating sequences. For each game not ending in checkmate, we analyze the final position using Stockfish and determine whether a forced mate exists within a horizon of 10 moves. When such a sequence is identified, we extend the PGN by appending the shortest mating line. This augmentation enables the model to learn legally correct mating sequences while preserving stylistic optimality, as expert play typically minimizes mate length.
    \item \textit{Train--test split.} We partition each \gm's collection into training and test sets using an 80:20 ratio. Splits are constructed via stratified sampling with respect to \gm color and game outcome, ensuring that relative class frequencies are approximately preserved. Training sets are used both for the self-supervised training of individual persona expert models, and for the calibration of the routing and weight-merged layers during the construction of the unified sparse model. A subset of the training games, together with the held-out test sets, is further used for the construction and evaluation of behavioral stylometry diagnostics (\cref{subsec:data_stylometry}).
\end{enumerate}

\paragraph{Statistics}
Table~\ref{tab:grandmasters} provides an aggregated view of training and test splits.
It summarizes, for each \gm, the number of games, win-rate, Elo at play time, and descriptive statistics of game length in terms of moves.
Across \gm collections, the average Elo rating is 2,817; the relatively tight dispersion suggests that the datasets are well-concentrated around elite-level performance.
Figure~\ref{fig:unique_games} reports the distribution of unique game prefixes as a function of move index.
To avoid overestimating diversity due to duplicated games across \gm collections (e.g., when two selected \gms play each other), the analysis is conducted on the subset of games with unique PGN strings.
At move 0, all games share the same initial board configuration, yielding complete overlap.
As the game progresses, the proportion of unique prefixes increases rapidly, reflecting combinatorial branching in the game tree.
By move 30, only $3.5\%$ of sequences remain non-unique, indicating that the vast majority of games have diverged into distinct trajectories.
We further decompose these repetitions and observe that approximately $60\%$ originate from the same player (individual repertoire), while the remaining $40\%$ are shared among different players.
The latter reflects common opening theory and well-established lines, where engine-assisted preparation leads multiple \gms to follow identical sequences for several moves before diverging.
Table~\ref{tab:gm_dataset_1} and Table~\ref{tab:gm_dataset_2} report the top-5 opening classes for each \gm, divided by played color.
Openings are identified using the Lichess Encyclopedia of Chess Openings, which defines a fine-grained vocabulary of 3,627 classes.\footnote[12]{\url{https://huggingface.co/datasets/Lichess/chess-openings}}
Across players, the distributions are generally flat: for a given \gm and color, the most frequent opening accounts for only a small fraction of games (2\%--5\% range, with few exceptions).
While variations of the same system can appear among the top entries, the majority of top-5 openings for a given player span distinct families (e.g., Ruy Lopez, Sicilian Defense, English Opening).
This pattern suggests that elite players actively maintain multiple high-level options within their repertoire, rather than relying on dominant lines.

\begin{figure*}[!htb]
    \centering
    \begin{minipage}{.64\textwidth}
        \captionof{table}{\textbf{Grandmaster data statistics.} Aggregated view (train, test). Played games span from 1984 to 2025.}
        \centering
        \begin{adjustbox}{width=\linewidth}
        \begin{threeparttable}
            \begin{tabular}{llrrrrr}
            \toprule
            & & & & \multicolumn{3}{c}{\textbf{\# Moves / Game}} \\
            \cmidrule(lr){5-7}
            \textbf{Grandmaster} & \textbf{Elo (Avg$\pm$Std)} & \textbf{\# Games}\tnote{$\dagger$} & \textbf{\faTrophy\;(\%)}\tnote{$\ddagger$} & Min & Avg & Max \\
            \hline
            \ding{182} V. Anand & 2,752 $\pm$ 15 & 4,182 & 35 & 11 & 81 & 290 \\
            \ding{183} L. Aronian & 2,797 $\pm$ 100 & 5,192 & 37 & 10 & 92 & 321 \\
            \ding{184} M. Carlsen & 2,943 $\pm$ 162 & 6,756 & 50 & 9 & 93 & 321 \\
            \ding{185} F. Caruana & 2,800 $\pm$ 52 & 5,538 & 45 & 6 & 97 & 340 \\
            \ding{186} A. Firouzja & 2,792 $\pm$ 102 & 4,370 & 51 & 8 & 93 & 359 \\
            \ding{187} A. Giri & 2,745 $\pm$ 44 & 4,984 & 39 & 10 & 92 & 349 \\
            \ding{188} H. Nakamura & 2,910 $\pm$ 188 & 8,916 & 53 & 7 & 92 & 400 \\
            \ding{189} I. Nepomniachtchi & 2,791 $\pm$ 70 & 11,052 & 51 & 6 & 91 & 323 \\
            \ding{190} W. So & 2,820 $\pm$ 118 & 9,040 & 50 & 6 & 87 & 396 \\
            \ding{191} M. Vachier-Lagrave & 2,820 $\pm$ 135 & 1,208 & 37 & 10 & 92 & 253 \\
            \bottomrule
            \end{tabular}
        \begin{tablenotes}
        \item[$\dagger$] 27,565 from PGN Mentor, 26,943 from Chess.com, 6,730 from Lichess.\\
        \item[$\ddagger$] Proportion of games won.
        \end{tablenotes}
        \end{threeparttable}
        \end{adjustbox}
        \label{tab:grandmasters}
    \end{minipage}
    \hfill
    \begin{minipage}{.325\textwidth}
        \centering
        \hspace{-2mm}
        \begin{tikzpicture}
            \begin{axis}[
                width=1.15\linewidth, height=4.6cm,
                ymajorgrids=true,
                grid=both,
                grid style=dashed,
                axis background/.style={fill=plotbackground},
                ybar stacked,
                bar width=2pt,
                enlarge x limits=0.02,  
                xmin=-1, xmax=31,
                xtick={0,10,20,30},
                xlabel={Move},
                ymin=0,
                ylabel={Game},
                every tick label/.append style={font=\fontsize{8}{8}\selectfont},
                xlabel style={font=\fontsize{8}{8}\selectfont},
                ylabel style={font=\fontsize{8}{8}\selectfont},
            ]

            \addplot+[ybar, fill=uniquegamebar, draw=none] plot coordinates {
                (0,1) (1,20) (2,250) (3,907) (4,2073) (5,3948) (6,6613) (7,9653) (8,13330) (9,17288) (10,20913) (11,24875) (12,28741) (13,32305) (14,35548) (15,38494) (16,41118) (17,43432) (18,45439) (19,47238) (20,48740) (21,49962) (22,50928) (23,51742) (24,52405) (25,53017) (26,53466) (27,53854) (28,54126) (29,54405) (30,54641)
            };
            
            \addplot+[ybar, fill=nonuniquegamebar, draw=none] plot coordinates {
                (0,56575) (1,56556) (2,56326) (3,55669) (4,54503) (5,52628) (6,49963) (7,46923) (8,43246) (9,39288) (10,35663) (11,31701) (12,27835) (13,24271) (14,21028) (15,18082) (16,15458) (17,13144) (18,11137) (19,9338) (20,7836) (21,6614) (22,5648) (23,4834) (24,4171) (25,3559) (26,3110) (27,2722) (28,2450) (29,2171) (30,1935)
            };

            \end{axis}
        \end{tikzpicture}
        \captionof{figure}{\textbf{Distribution of unique games (\textcolor{uniquegamebar}{{\tiny\faCircle}}) by move.}}
        \label{fig:unique_games}
    \end{minipage}
    \vspace{-0.5cm}
\end{figure*}

\begin{table}[!htb]
    \centering
    \caption{\textbf{Statistical overview of the grandmaster persona datasets (part 1).}  The columns detail, from left to right: (i) the distribution of game results stratified by color; (ii) the frequency of the top-5 openings stratified by color; (iii) the provenance of PGN games.}
    \label{tab:gm_dataset_1}
    \resizebox{\textwidth}{!}{
    \begin{tabular}{p{2cm}p{3.2cm}p{10cm}p{2.8cm}}
    \toprule
    \textbf{Grandmaster} & \textbf{Results (\%)} & \textbf{Top-5 Openings (\%, log-scale)} & \textbf{Game Sources (\%)} \\
    \toprule
    \begin{minipage}{2cm}
        \centering
        \includegraphics[width=1.2cm]{figures/icons/anand.png}\\
        \ding{182} V. Anand
    \end{minipage} &
    \begin{minipage}{3.2cm}
        \begin{tikzpicture}
        \begin{axis}[
            width=0.8\linewidth, height=2cm,
            scale only axis,
            ymajorgrids=true,
            grid=both,
            grid style=dashed,
            axis background/.style={fill=plotbackground},
            ybar=0pt, 
            bar width=6pt,
            enlarge x limits=0.28,  
            xtick=data,
            symbolic x coords={
                Win,
                Draw,
                Lose
            },
            xlabel style={font=\small},
            ymax=100,
            ymin=0,
            ylabel style={font=\small},
            ytick={20,40,60,80,100},
            every tick label/.append style={font=\fontsize{8}{8}\selectfont},
        ]

        \addplot [fill=white, draw=black] coordinates {
            (Win, 43)
            (Draw, 45)
            (Lose, 12)
        };
        \addplot [fill=black, draw=black] coordinates {
            (Win, 27)
            (Draw, 55)
            (Lose, 18)
        };
        
        \draw (axis cs:{[normalized]\pgfkeysvalueof{/pgfplots/xmin}},0)
            -- (axis cs:{[normalized]\pgfkeysvalueof{/pgfplots/xmax}},0);
        
        \end{axis}
        \end{tikzpicture}
    \end{minipage} & 
    \begin{minipage}{5cm}
        \begin{tikzpicture}
        \begin{axis}[
            width=0.8\linewidth, height=1.2cm,
            scale only axis,
            ymajorgrids=true,
            grid=both,
            grid style=dashed,
            axis background/.style={fill=plotbackground},
            ybar=0pt, 
            bar width=6pt,
            enlarge x limits=0.1,  
            xtick=data,
            symbolic x coords={
                {Ruy Lopez:\\Berlin Def. AB},
                {Sicilian Def.:\\Najdorf},
                {Zukertort},
                {Ruy Lopez:\\Closed},
                {Caro-Kann\\Def.}
            },
            xlabel style={font=\small},
            xticklabel style={rotate=90, anchor=east, align=right},
            ymax=100,
            ymin=0,
            ylabel style={font=\small},
            ymode=log,
            every tick label/.append style={font=\fontsize{8}{8}\selectfont},
        ]
        
        \addplot [fill=white, draw=black] coordinates {
            (Ruy Lopez:\\Berlin Def. AB, 3.97)
            (Sicilian Def.:\\Najdorf, 2.44)
            (Zukertort, 2.30)
            (Ruy Lopez:\\Closed, 2.30)
            (Caro-Kann\\Def., 2.30)
        };
        
        \draw (axis cs:{[normalized]\pgfkeysvalueof{/pgfplots/xmin}},0)
            -- (axis cs:{[normalized]\pgfkeysvalueof{/pgfplots/xmax}},0);
        
        \end{axis}
        \end{tikzpicture}
    \end{minipage}
    \begin{minipage}{5cm}
        \begin{tikzpicture}
        \begin{axis}[
            width=0.8\linewidth, height=1.2cm,
            scale only axis,
            ymajorgrids=true,
            grid=both,
            grid style=dashed,
            axis background/.style={fill=plotbackground},
            ybar=0pt, 
            bar width=6pt,
            enlarge x limits=0.1,  
            xtick=data,
            symbolic x coords={
                {Zukertort},
                {Ruy Lopez:\\Berlin Def. AB},
                {Indian Def.:\\Knights},
                {English:\\King},
                {English:\\Symmetrical}
            },
            xlabel style={font=\small},
            xticklabel style={rotate=90, anchor=east, align=right},
            ymax=100,
            ymin=0,
            ylabel style={font=\small},
            ymode=log,
            every tick label/.append style={font=\fontsize{8}{8}\selectfont},
        ]
        
        \addplot [fill=black, draw=black] coordinates {
            (Zukertort, 3.97)
            (Ruy Lopez:\\Berlin Def. AB, 1.96)
            (Indian Def.:\\Knights, 1.82)
            (English:\\King, 1.63)
            (English:\\Symmetrical, 1.63)
        };
        
        \draw (axis cs:{[normalized]\pgfkeysvalueof{/pgfplots/xmin}},0)
            -- (axis cs:{[normalized]\pgfkeysvalueof{/pgfplots/xmax}},0);
        
        \end{axis}
        \end{tikzpicture}
    \end{minipage} &
    \begin{minipage}{2.5cm}
        \begin{tikzpicture}[font=\small]
            \def\WCtest#1#2{
                \pgfmathparse{
                    \WCpercentage>30?"#1":"#2"
                }
                \pgfmathresult
            }
            \wheelchart[
                data=\WCtest{}{\WCvarC\\(\WCpercentage)}, 
                data sep=0, 
                radius={0}{1}, 
                lines={\WCpercentage>30?0:0.2},
                lines sep=0, 
                lines pos=0.8, 
                lines style={dotted,thick},
                wheel data=\WCtest{\WCvarC\\(\WCpercentage)}{}, 
                wheel data pos=0.4, 
            ]{
            1/chesscom-color/\includegraphics[width=3mm]{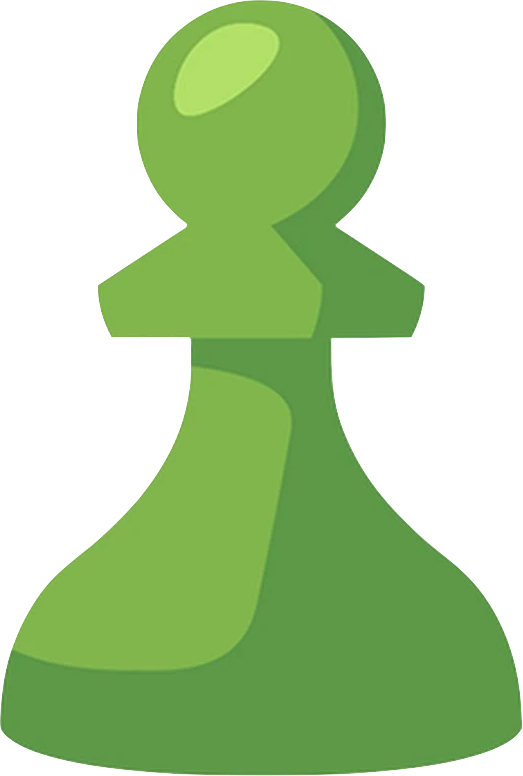},
            6/lichess-color/\includegraphics[width=4mm]{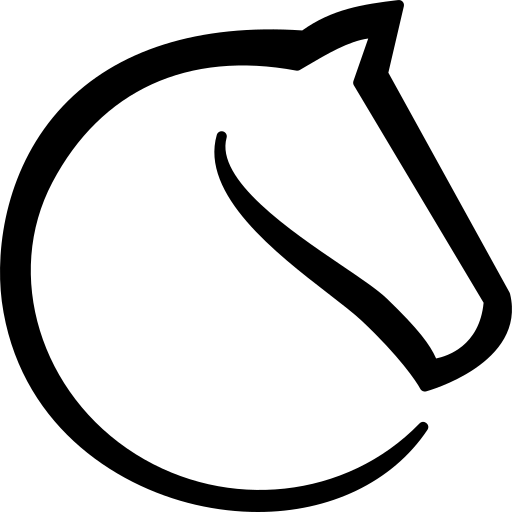},
            93/pgnmentor-color/\includegraphics[width=7mm]{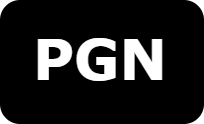}
            }
        \end{tikzpicture}
    \end{minipage} \\
    \midrule
    \begin{minipage}{2cm}
        \centering
        \includegraphics[width=1.2cm]{figures/icons/aronian.png}\\
        \ding{183} L. Aronian
    \end{minipage} &
    \begin{minipage}{3.2cm}
        \begin{tikzpicture}
        \begin{axis}[
            width=0.8\linewidth, height=2cm,
            scale only axis,
            ymajorgrids=true,
            grid=both,
            grid style=dashed,
            axis background/.style={fill=plotbackground},
            ybar=0pt, 
            bar width=6pt,
            enlarge x limits=0.28,  
            xtick=data,
            symbolic x coords={
                Win,
                Draw,
                Lose
            },
            xlabel style={font=\small},
            ymax=100,
            ymin=0,
            ylabel style={font=\small},
            ytick={20,40,60,80,100},
            every tick label/.append style={font=\fontsize{8}{8}\selectfont},
        ]

        \addplot [fill=white, draw=black] coordinates {
            (Win, 45)
            (Draw, 39)
            (Lose, 16)
        };
        \addplot [fill=black, draw=black] coordinates {
            (Win, 30)
            (Draw, 41)
            (Lose, 29)
        };
        
        \draw (axis cs:{[normalized]\pgfkeysvalueof{/pgfplots/xmin}},0)
            -- (axis cs:{[normalized]\pgfkeysvalueof{/pgfplots/xmax}},0);
        
        \end{axis}
        \end{tikzpicture}
    \end{minipage} & 
    \begin{minipage}{5cm}
        \begin{tikzpicture}
        \begin{axis}[
            width=0.8\linewidth, height=1.2cm,
            scale only axis,
            ymajorgrids=true,
            grid=both,
            grid style=dashed,
            axis background/.style={fill=plotbackground},
            ybar=0pt, 
            bar width=6pt,
            enlarge x limits=0.1,  
            xtick=data,
            symbolic x coords={
                {Zukertort},
                {Sicilian Def.:\\Del. Alapin},
                {English:\\Agincourt Def.},
                {Indian Def.:\\Knights},
                {Bishop:\\Berlin Def.}
            },
            xlabel style={font=\small},
            xticklabel style={rotate=90, anchor=east, align=right},
            ymax=100,
            ymin=0,
            ylabel style={font=\small},
            ymode=log,
            every tick label/.append style={font=\fontsize{8}{8}\selectfont},
        ]
        
        \addplot [fill=white, draw=black] coordinates {
            (Zukertort, 4.01)
            (Sicilian Def.:\\Del. Alapin, 2.85)
            (English:\\Agincourt Def., 2.27)
            (Indian Def.:\\Knights, 2.23)
            (Bishop:\\Berlin Def., 1.73)
        };
        
        \draw (axis cs:{[normalized]\pgfkeysvalueof{/pgfplots/xmin}},0)
            -- (axis cs:{[normalized]\pgfkeysvalueof{/pgfplots/xmax}},0);
        
        \end{axis}
        \end{tikzpicture}
    \end{minipage}
    \begin{minipage}{5cm}
        \begin{tikzpicture}
        \begin{axis}[
            width=0.8\linewidth, height=1.2cm,
            scale only axis,
            ymajorgrids=true,
            grid=both,
            grid style=dashed,
            axis background/.style={fill=plotbackground},
            ybar=0pt, 
            bar width=6pt,
            enlarge x limits=0.1,  
            xtick=data,
            symbolic x coords={
                {Zukertort},
                {Ruy Lopez:\\Berlin Def. AB},
                {English:\\Anglo-Ind. Def.},
                {Ruy Lopez:\\Berlin Def. RG},
                {King's Indian\\Attack}
            },
            xlabel style={font=\small},
            xticklabel style={rotate=90, anchor=east, align=right},
            ymax=100,
            ymin=0,
            ylabel style={font=\small},
            ymode=log,
            every tick label/.append style={font=\fontsize{8}{8}\selectfont},
        ]
        
        \addplot [fill=black, draw=black] coordinates {
            (Zukertort, 3.47)
            (Ruy Lopez:\\Berlin Def. AB, 3.08)
            (English:\\Anglo-Ind. Def., 2.58)
            (Ruy Lopez:\\Berlin Def. RG, 2.27)
            (King's Indian\\Attack, 2.12)
        };
        
        \draw (axis cs:{[normalized]\pgfkeysvalueof{/pgfplots/xmin}},0)
            -- (axis cs:{[normalized]\pgfkeysvalueof{/pgfplots/xmax}},0);
        
        \end{axis}
        \end{tikzpicture}
    \end{minipage} &
    \begin{minipage}{2.5cm}
        \begin{tikzpicture}[font=\small]
            \def\WCtest#1#2{
                \pgfmathparse{
                    \WCpercentage>30?"#1":"#2"
                }
                \pgfmathresult
            }
            \wheelchart[
                data=\WCtest{}{\WCvarC\\(\WCpercentage)}, 
                data sep=0, 
                radius={0}{1}, 
                lines={\WCpercentage>30?0:0.2},
                lines sep=0, 
                lines pos=0.8, 
                lines style={dotted,thick},
                wheel data=\WCtest{\WCvarC\\(\WCpercentage)}{}, 
                wheel data pos=0.4, 
            ]{
            15/chesscom-color/\includegraphics[width=3mm]{figures/icons/chesscom.png},
            16/lichess-color/\includegraphics[width=4mm]{figures/icons/lichess.png},
            69/pgnmentor-color/\includegraphics[width=7mm]{figures/icons/pgnmentor.png}
            }
        \end{tikzpicture}
    \end{minipage} \\
    \midrule
    \begin{minipage}{2cm}
        \centering
        \includegraphics[width=1.2cm]{figures/icons/carlsen.png}\\
        \ding{184} M. Carlsen
    \end{minipage} &
    \begin{minipage}{3.2cm}
        \begin{tikzpicture}
        \begin{axis}[
            width=0.8\linewidth, height=2cm,
            scale only axis,
            ymajorgrids=true,
            grid=both,
            grid style=dashed,
            axis background/.style={fill=plotbackground},
            ybar=0pt, 
            bar width=6pt,
            enlarge x limits=0.28,  
            xtick=data,
            symbolic x coords={
                Win,
                Draw,
                Lose
            },
            xlabel style={font=\small},
            ymax=100,
            ymin=0,
            ylabel style={font=\small},
            ytick={20,40,60,80,100},
            every tick label/.append style={font=\fontsize{8}{8}\selectfont},
        ]

        \addplot [fill=white, draw=black] coordinates {
            (Win, 59)
            (Draw, 28)
            (Lose, 13)
        };
        \addplot [fill=black, draw=black] coordinates {
            (Win, 41)
            (Draw, 19)
            (Lose, 40)
        };
        
        \draw (axis cs:{[normalized]\pgfkeysvalueof{/pgfplots/xmin}},0)
            -- (axis cs:{[normalized]\pgfkeysvalueof{/pgfplots/xmax}},0);
        
        \end{axis}
        \end{tikzpicture}
    \end{minipage} & 
    \begin{minipage}{5cm}
        \begin{tikzpicture}
        \begin{axis}[
            width=0.8\linewidth, height=1.2cm,
            scale only axis,
            ymajorgrids=true,
            grid=both,
            grid style=dashed,
            axis background/.style={fill=plotbackground},
            ybar=0pt, 
            bar width=6pt,
            enlarge x limits=0.1,  
            xtick=data,
            symbolic x coords={
                {Zukertort},
                {Ruy Lopez:\\Berlin Def. AB},
                {Indian Def.:\\Knights},
                {Indian Def.:\\Acc. London},
                {Indian Def.:\\Anti-Nimzo-Ind.}
            },
            xlabel style={font=\small},
            xticklabel style={rotate=90, anchor=east, align=right},
            ymax=100,
            ymin=0,
            ylabel style={font=\small},
            ymode=log,
            every tick label/.append style={font=\fontsize{8}{8}\selectfont},
        ]
        
        \addplot [fill=white, draw=black] coordinates {
            (Zukertort, 3.02)
            (Ruy Lopez:\\Berlin Def. AB, 2.28)
            (Indian Def.:\\Knights, 1.95)
            (Indian Def.:\\Acc. London, 1.78)
            (Indian Def.:\\Anti-Nimzo-Ind., 1.36)
        };
        
        \draw (axis cs:{[normalized]\pgfkeysvalueof{/pgfplots/xmin}},0)
            -- (axis cs:{[normalized]\pgfkeysvalueof{/pgfplots/xmax}},0);
        
        \end{axis}
        \end{tikzpicture}
    \end{minipage}
    \begin{minipage}{5cm}
        \begin{tikzpicture}
        \begin{axis}[
            width=0.8\linewidth, height=1.2cm,
            scale only axis,
            ymajorgrids=true,
            grid=both,
            grid style=dashed,
            axis background/.style={fill=plotbackground},
            ybar=0pt, 
            bar width=6pt,
            enlarge x limits=0.1,  
            xtick=data,
            symbolic x coords={
                {Zukertort},
                {Modern Def.},
                {Qween's Gambit\\Declined},
                {Sicilian Def.:\\NR Attack},
                {Pirc Def.}
            },
            xlabel style={font=\small},
            xticklabel style={rotate=90, anchor=east, align=right},
            ymax=100,
            ymin=0,
            ylabel style={font=\small},
            ymode=log,
            every tick label/.append style={font=\fontsize{8}{8}\selectfont},
        ]
        
        \addplot [fill=black, draw=black] coordinates {
            (Zukertort, 2.58)
            (Modern Def., 1.84)
            (Qween's Gambit\\Declined, 1.51)
            (Sicilian Def.:\\NR Attack, 1.51)
            (Pirc Def., 1.30)
        };
        
        \draw (axis cs:{[normalized]\pgfkeysvalueof{/pgfplots/xmin}},0)
            -- (axis cs:{[normalized]\pgfkeysvalueof{/pgfplots/xmax}},0);
        
        \end{axis}
        \end{tikzpicture}
    \end{minipage} &
    \begin{minipage}{2.5cm}
        \begin{tikzpicture}[font=\small]
            \def\WCtest#1#2{
                \pgfmathparse{
                    \WCpercentage>30?"#1":"#2"
                }
                \pgfmathresult
            }
            \wheelchart[
                data=\WCtest{}{\WCvarC\\(\WCpercentage)}, 
                data sep=0, 
                radius={0}{1}, 
                lines={\WCpercentage>30?0:0.2},
                lines sep=0, 
                lines pos=0.8, 
                lines style={dotted,thick},
                wheel data=\WCtest{\WCvarC\\(\WCpercentage)}{}, 
                wheel data pos=0.4, 
            ]{
            41/chesscom-color/\includegraphics[width=3mm]{figures/icons/chesscom.png},
            12/lichess-color/\includegraphics[width=4mm]{figures/icons/lichess.png},
            47/pgnmentor-color/\includegraphics[width=7mm]{figures/icons/pgnmentor.png}
            }
        \end{tikzpicture}
    \end{minipage} \\
    \midrule
    \begin{minipage}{2cm}
        \centering
        \includegraphics[width=1.2cm]{figures/icons/caruana.png}\\
        \ding{185} F. Caruana
    \end{minipage} &
    \begin{minipage}{3.2cm}
        \begin{tikzpicture}
        \begin{axis}[
            width=0.8\linewidth, height=2cm,
            scale only axis,
            ymajorgrids=true,
            grid=both,
            grid style=dashed,
            axis background/.style={fill=plotbackground},
            ybar=0pt, 
            bar width=6pt,
            enlarge x limits=0.28,  
            xtick=data,
            symbolic x coords={
                Win,
                Draw,
                Lose
            },
            xlabel style={font=\small},
            ymax=100,
            ymin=0,
            ylabel style={font=\small},
            ytick={20,40,60,80,100},
            every tick label/.append style={font=\fontsize{8}{8}\selectfont},
        ]

        \addplot [fill=white, draw=black] coordinates {
            (Win, 50)
            (Draw, 30)
            (Lose, 20)
        };
        \addplot [fill=black, draw=black] coordinates {
            (Win, 40)
            (Draw, 20)
            (Lose, 40)
        };
        
        \draw (axis cs:{[normalized]\pgfkeysvalueof{/pgfplots/xmin}},0)
            -- (axis cs:{[normalized]\pgfkeysvalueof{/pgfplots/xmax}},0);
        
        \end{axis}
        \end{tikzpicture}
    \end{minipage} & 
    \begin{minipage}{5cm}
        \begin{tikzpicture}
        \begin{axis}[
            width=0.8\linewidth, height=1.2cm,
            scale only axis,
            ymajorgrids=true,
            grid=both,
            grid style=dashed,
            axis background/.style={fill=plotbackground},
            ybar=0pt, 
            bar width=6pt,
            enlarge x limits=0.1,  
            xtick=data,
            symbolic x coords={
                {Ruy Lopez:\\Berlin Def. AB},
                {English:\\Agincourt Def.},
                {Sicilian Def.:\\NR Attack},
                {English:\\Anglo-Ind. Def.},
                {Sicilian Def.:\\Najdorf},
            },
            xlabel style={font=\small},
            xticklabel style={rotate=90, anchor=east, align=right},
            ymax=100,
            ymin=0,
            ylabel style={font=\small},
            ymode=log,
            every tick label/.append style={font=\fontsize{8}{8}\selectfont},
        ]
        
        \addplot [fill=white, draw=black] coordinates {
            (Ruy Lopez:\\Berlin Def. AB, 5.20)
            (English:\\Agincourt Def., 2.35)
            (Sicilian Def.:\\NR Attack, 2.24)
            (English:\\Anglo-Ind. Def., 2.24)
            (Sicilian Def.:\\Najdorf, 1.77)
        };
        
        \draw (axis cs:{[normalized]\pgfkeysvalueof{/pgfplots/xmin}},0)
            -- (axis cs:{[normalized]\pgfkeysvalueof{/pgfplots/xmax}},0);
        
        \end{axis}
        \end{tikzpicture}
    \end{minipage}
    \begin{minipage}{5cm}
        \begin{tikzpicture}
        \begin{axis}[
            width=0.8\linewidth, height=1.2cm,
            scale only axis,
            ymajorgrids=true,
            grid=both,
            grid style=dashed,
            axis background/.style={fill=plotbackground},
            ybar=0pt, 
            bar width=6pt,
            enlarge x limits=0.1,  
            xtick=data,
            symbolic x coords={
                {Zukertort},
                {Indian Def.:\\Knights},
                {Sicilian Def.:\\NR Attack},
                {Ruy Lopez:\\Berlin Def. AB},
                {Queen's Pawn\\Game}
            },
            xlabel style={font=\small},
            xticklabel style={rotate=90, anchor=east, align=right},
            ymax=100,
            ymin=0,
            ylabel style={font=\small},
            ymode=log,
            every tick label/.append style={font=\fontsize{8}{8}\selectfont},
        ]
        
        \addplot [fill=black, draw=black] coordinates {
            (Zukertort, 2.78)
            (Indian Def.:\\Knights, 2.78)
            (Sicilian Def.:\\NR Attack, 1.77)
            (Ruy Lopez:\\Berlin Def. AB, 1.63)
            (Queen's Pawn\\Game, 1.37)
        };
        
        \draw (axis cs:{[normalized]\pgfkeysvalueof{/pgfplots/xmin}},0)
            -- (axis cs:{[normalized]\pgfkeysvalueof{/pgfplots/xmax}},0);
        
        \end{axis}
        \end{tikzpicture}
    \end{minipage} &
    \begin{minipage}{2.5cm}
        \begin{tikzpicture}[font=\small]
            \def\WCtest#1#2{
                \pgfmathparse{
                    \WCpercentage>30?"#1":"#2"
                }
                \pgfmathresult
            }
            \wheelchart[
                data=\WCtest{}{\WCvarC\\(\WCpercentage)}, 
                data sep=0, 
                radius={0}{1}, 
                lines={\WCpercentage>30?0:0.2},
                lines sep=0, 
                lines pos=0.8, 
                lines style={dotted,thick},
                wheel data=\WCtest{\WCvarC\\(\WCpercentage)}{}, 
                wheel data pos=0.4, 
            ]{
            36/chesscom-color/\includegraphics[width=3mm]{figures/icons/chesscom.png},
            16/lichess-color/\includegraphics[width=4mm]{figures/icons/lichess.png},
            48/pgnmentor-color/\includegraphics[width=7mm]{figures/icons/pgnmentor.png}
            }
        \end{tikzpicture}
    \end{minipage} \\
    \midrule
    \begin{minipage}{2cm}
        \centering
        \includegraphics[width=1.2cm]{figures/icons/firouzja.png}\\
        \ding{186} A. Firouzja
    \end{minipage} &
    \begin{minipage}{3.2cm}
        \begin{tikzpicture}
        \begin{axis}[
            width=0.8\linewidth, height=2cm,
            scale only axis,
            ymajorgrids=true,
            grid=both,
            grid style=dashed,
            axis background/.style={fill=plotbackground},
            ybar=0pt, 
            bar width=6pt,
            enlarge x limits=0.28,  
            xtick=data,
            symbolic x coords={
                Win,
                Draw,
                Lose
            },
            xlabel style={font=\small},
            ymax=100,
            ymin=0,
            ylabel style={font=\small},
            ytick={20,40,60,80,100},
            every tick label/.append style={font=\fontsize{8}{8}\selectfont},
        ]

        \addplot [fill=white, draw=black] coordinates {
            (Win, 59)
            (Draw, 21)
            (Lose, 20)
        };
        \addplot [fill=black, draw=black] coordinates {
            (Win, 43)
            (Draw, 10)
            (Lose, 47)
        };
        
        \draw (axis cs:{[normalized]\pgfkeysvalueof{/pgfplots/xmin}},0)
            -- (axis cs:{[normalized]\pgfkeysvalueof{/pgfplots/xmax}},0);
        
        \end{axis}
        \end{tikzpicture}
    \end{minipage} & 
    \begin{minipage}{5cm}
        \begin{tikzpicture}
        \begin{axis}[
            width=0.8\linewidth, height=1.2cm,
            scale only axis,
            ymajorgrids=true,
            grid=both,
            grid style=dashed,
            axis background/.style={fill=plotbackground},
            ybar=0pt, 
            bar width=6pt,
            enlarge x limits=0.1,  
            xtick=data,
            symbolic x coords={
                {Trompowsky\\Attack},
                {Ruy Lopez:\\Berlin Def. AB},
                {Queen's Pawn\\Game: Levitsky},
                {Sicilian Def.:\\Najdorf},
                {King's Indian\\Attack},
            },
            xlabel style={font=\small},
            xticklabel style={rotate=90, anchor=east, align=right},
            ymax=100,
            ymin=0,
            ylabel style={font=\small},
            ymode=log,
            every tick label/.append style={font=\fontsize{8}{8}\selectfont},
        ]
        
        \addplot [fill=white, draw=black] coordinates {
            (Trompowsky\\Attack, 3.94)
            (Ruy Lopez:\\Berlin Def. AB, 3.07)
            (Queen's Pawn\\Game: Levitsky, 2.15)
            (Sicilian Def.:\\Najdorf, 1.69)
            (King's Indian\\Attack, 1.56)
        };
        
        \draw (axis cs:{[normalized]\pgfkeysvalueof{/pgfplots/xmin}},0)
            -- (axis cs:{[normalized]\pgfkeysvalueof{/pgfplots/xmax}},0);
        
        \end{axis}
        \end{tikzpicture}
    \end{minipage}
    \begin{minipage}{5cm}
        \begin{tikzpicture}
        \begin{axis}[
            width=0.8\linewidth, height=1.2cm,
            scale only axis,
            ymajorgrids=true,
            grid=both,
            grid style=dashed,
            axis background/.style={fill=plotbackground},
            ybar=0pt, 
            bar width=6pt,
            enlarge x limits=0.1,  
            xtick=data,
            symbolic x coords={
                {Zukertort},
                {King's Indian\\Attack},
                {Queen's Pawn\\Game: Modern},
                {Zukertort:\\Pirc Invit.},
                {Zukertort:\\Kingside F.}
            },
            xlabel style={font=\small},
            xticklabel style={rotate=90, anchor=east, align=right},
            ymax=100,
            ymin=0,
            ylabel style={font=\small},
            ymode=log,
            every tick label/.append style={font=\fontsize{8}{8}\selectfont},
        ]
        
        \addplot [fill=black, draw=black] coordinates {
            (Zukertort, 4.67)
            (King's Indian\\Attack, 2.61)
            (Queen's Pawn\\Game: Modern, 2.43)
            (Zukertort:\\Pirc Invit., 1.92)
            (Zukertort:\\Kingside F., 1.92)
        };
        
        \draw (axis cs:{[normalized]\pgfkeysvalueof{/pgfplots/xmin}},0)
            -- (axis cs:{[normalized]\pgfkeysvalueof{/pgfplots/xmax}},0);
        
        \end{axis}
        \end{tikzpicture}
    \end{minipage} &
    \begin{minipage}{2.5cm}
        \begin{tikzpicture}[font=\small]
            \def\WCtest#1#2{
                \pgfmathparse{
                    \WCpercentage>30?"#1":"#2"
                }
                \pgfmathresult
            }
            \wheelchart[
                data=\WCtest{}{\WCvarC\\(\WCpercentage)}, 
                data sep=0, 
                radius={0}{1}, 
                lines={\WCpercentage>30?0:0.2},
                lines sep=0, 
                lines pos=0.8, 
                lines style={dotted,thick},
                wheel data=\WCtest{\WCvarC\\(\WCpercentage)}{}, 
                wheel data pos=0.4, 
            ]{
            49/chesscom-color/\includegraphics[width=3mm]{figures/icons/chesscom.png},
            8/lichess-color/\includegraphics[width=4mm]{figures/icons/lichess.png},
            43/pgnmentor-color/\includegraphics[width=7mm]{figures/icons/pgnmentor.png}
            }
        \end{tikzpicture}
    \end{minipage} \\
    \bottomrule
    \end{tabular}
    }
\end{table}

\begin{table}[t]
    \centering
    \caption{\textbf{Statistical overview of the grandmaster persona datasets (part 2).}}
    \label{tab:gm_dataset_2}
    \resizebox{\textwidth}{!}{
    \begin{tabular}{p{2cm}p{3.2cm}p{10cm}p{2.8cm}}
    \toprule
    \textbf{Grandmaster} & \textbf{Results (\%)} & \textbf{Top-5 Openings (\%, log-scale)} & \textbf{Game Sources (\%)} \\
    \toprule
    \begin{minipage}{2cm}
        \centering
        \includegraphics[width=1.2cm]{figures/icons/giri.png}\\
        \ding{187} A. Giri
    \end{minipage} &
    \begin{minipage}{3.2cm}
        \begin{tikzpicture}
        \begin{axis}[
            width=0.8\linewidth, height=2cm,
            scale only axis,
            ymajorgrids=true,
            grid=both,
            grid style=dashed,
            axis background/.style={fill=plotbackground},
            ybar=0pt, 
            bar width=6pt,
            enlarge x limits=0.28,  
            xtick=data,
            symbolic x coords={
                Win,
                Draw,
                Lose
            },
            xlabel style={font=\small},
            ymax=100,
            ymin=0,
            ylabel style={font=\small},
            ytick={20,40,60,80,100},
            every tick label/.append style={font=\fontsize{8}{8}\selectfont},
        ]

        \addplot [fill=white, draw=black] coordinates {
            (Win, 44)
            (Draw, 40)
            (Lose, 16)
        };
        \addplot [fill=black, draw=black] coordinates {
            (Win, 35)
            (Draw, 40)
            (Lose, 25)
        };
        
        \draw (axis cs:{[normalized]\pgfkeysvalueof{/pgfplots/xmin}},0)
            -- (axis cs:{[normalized]\pgfkeysvalueof{/pgfplots/xmax}},0);
        
        \end{axis}
        \end{tikzpicture}
    \end{minipage} & 
    \begin{minipage}{5cm}
        \begin{tikzpicture}
        \begin{axis}[
            width=0.8\linewidth, height=1.2cm,
            scale only axis,
            ymajorgrids=true,
            grid=both,
            grid style=dashed,
            axis background/.style={fill=plotbackground},
            ybar=0pt, 
            bar width=6pt,
            enlarge x limits=0.1,  
            xtick=data,
            symbolic x coords={
                {Zukertort},
                {Indian Def.:\\Anti-Nimzo-Ind.},
                {Italian Game:\\Classical},
                {Ruy Lopez:\\Berlin Def. AB},
                {Zukertort:\\Sicilian Invit.}
            },
            xlabel style={font=\small},
            xticklabel style={rotate=90, anchor=east, align=right},
            ymax=100,
            ymin=0,
            ylabel style={font=\small},
            ymode=log,
            every tick label/.append style={font=\fontsize{8}{8}\selectfont},
        ]
        
        \addplot [fill=white, draw=black] coordinates {
            (Zukertort, 8.35)
            (Indian Def.:\\Anti-Nimzo-Ind., 2.65)
            (Italian Game:\\Classical, 2.05)
            (Ruy Lopez:\\Berlin Def. AB, 1.97)
            (Zukertort:\\Sicilian Invit., 1.85)
        };
        
        \draw (axis cs:{[normalized]\pgfkeysvalueof{/pgfplots/xmin}},0)
            -- (axis cs:{[normalized]\pgfkeysvalueof{/pgfplots/xmax}},0);
        
        \end{axis}
        \end{tikzpicture}
    \end{minipage}
    \begin{minipage}{5cm}
        \begin{tikzpicture}
        \begin{axis}[
            width=0.8\linewidth, height=1.5cm,
            scale only axis,
            ymajorgrids=true,
            grid=both,
            grid style=dashed,
            axis background/.style={fill=plotbackground},
            ybar=0pt, 
            bar width=6pt,
            enlarge x limits=0.1,  
            xtick=data,
            symbolic x coords={
                {Zukertort},
                {Sicilian Def.:\\Najdorf},
                {Sicilian Def.:\\Moscow},
                {Indian Def.:\\Knights},
                {Nimzo-Larsen\\Attack: Classical}
            },
            xlabel style={font=\small},
            xticklabel style={rotate=90, anchor=east, align=right},
            ymax=100,
            ymin=0,
            ylabel style={font=\small},
            ymode=log,
            every tick label/.append style={font=\fontsize{8}{8}\selectfont},
        ]
        
        \addplot [fill=black, draw=black] coordinates {
            (Zukertort, 5.50)
            (Sicilian Def.:\\Najdorf, 2.93)
            (Sicilian Def.:\\Moscow, 2.89)
            (Indian Def.:\\Knights, 2.61)
            (Nimzo-Larsen\\Attack: Classical, 1.97)
        };
        
        \draw (axis cs:{[normalized]\pgfkeysvalueof{/pgfplots/xmin}},0)
            -- (axis cs:{[normalized]\pgfkeysvalueof{/pgfplots/xmax}},0);
        
        \end{axis}
        \end{tikzpicture}
    \end{minipage} &
    \begin{minipage}{2.5cm}
        \begin{tikzpicture}[font=\small]
            \def\WCtest#1#2{
                \pgfmathparse{
                    \WCpercentage>30?"#1":"#2"
                }
                \pgfmathresult
            }
            \wheelchart[
                data=\WCtest{}{\WCvarC\\(\WCpercentage)}, 
                data sep=0, 
                radius={0}{1}, 
                lines={\WCpercentage>30?0:0.2},
                lines sep=0, 
                lines pos=0.8, 
                lines style={dotted,thick},
                wheel data=\WCtest{\WCvarC\\(\WCpercentage)}{}, 
                wheel data pos=0.4, 
            ]{
            28/chesscom-color/\includegraphics[width=3mm]{figures/icons/chesscom.png},
            15/lichess-color/\includegraphics[width=4mm]{figures/icons/lichess.png},
            57/pgnmentor-color/\includegraphics[width=7mm]{figures/icons/pgnmentor.png}
            }
        \end{tikzpicture}
    \end{minipage} \\
    \midrule
    \begin{minipage}{2cm}
        \centering
        \includegraphics[width=1.2cm]{figures/icons/nakamura.png}\\
        \ding{188} H. Nakamura
    \end{minipage} &
    \begin{minipage}{3.2cm}
        \begin{tikzpicture}
        \begin{axis}[
            width=0.8\linewidth, height=2cm,
            scale only axis,
            ymajorgrids=true,
            grid=both,
            grid style=dashed,
            axis background/.style={fill=plotbackground},
            ybar=0pt, 
            bar width=6pt,
            enlarge x limits=0.28,  
            xtick=data,
            symbolic x coords={
                Win,
                Draw,
                Lose
            },
            xlabel style={font=\small},
            ymax=100,
            ymin=0,
            ylabel style={font=\small},
            ytick={20,40,60,80,100},
            every tick label/.append style={font=\fontsize{8}{8}\selectfont},
        ]

        \addplot [fill=white, draw=black] coordinates {
            (Win, 61)
            (Draw, 26)
            (Lose, 13)
        };
        \addplot [fill=black, draw=black] coordinates {
            (Win, 44)
            (Draw, 10)
            (Lose, 46)
        };
        
        \draw (axis cs:{[normalized]\pgfkeysvalueof{/pgfplots/xmin}},0)
            -- (axis cs:{[normalized]\pgfkeysvalueof{/pgfplots/xmax}},0);
        
        \end{axis}
        \end{tikzpicture}
    \end{minipage} & 
    \begin{minipage}{5cm}
        \begin{tikzpicture}
        \begin{axis}[
            width=0.8\linewidth, height=1.2cm,
            scale only axis,
            ymajorgrids=true,
            grid=both,
            grid style=dashed,
            axis background/.style={fill=plotbackground},
            ybar=0pt, 
            bar width=6pt,
            enlarge x limits=0.1,  
            xtick=data,
            symbolic x coords={
                {Zukertort},
                {Zukertort:\\Nimzo-Larsen},
                {Nimzo-Larsen\\Attack: Classical},
                {Zuckertort:\\Sicilian Invit.},
                {Ruz Lopez:\\Berlin Def. AB}
            },
            xlabel style={font=\small},
            xticklabel style={rotate=90, anchor=east, align=right},
            ymax=100,
            ymin=0,
            ylabel style={font=\small},
            ymode=log,
            every tick label/.append style={font=\fontsize{8}{8}\selectfont},
        ]
        
        \addplot [fill=white, draw=black] coordinates {
            (Zukertort, 9.74)
            (Zukertort:\\Nimzo-Larsen, 6.75)
            (Nimzo-Larsen\\Attack: Classical, 6.17)
            (Zuckertort:\\Sicilian Invit., 2.47)
            (Ruz Lopez:\\Berlin Def. AB, 2.04)
        };
        
        \draw (axis cs:{[normalized]\pgfkeysvalueof{/pgfplots/xmin}},0)
            -- (axis cs:{[normalized]\pgfkeysvalueof{/pgfplots/xmax}},0);
        
        \end{axis}
        \end{tikzpicture}
    \end{minipage}
    \begin{minipage}{5cm}
        \begin{tikzpicture}
        \begin{axis}[
            width=0.8\linewidth, height=1.5cm,
            scale only axis,
            ymajorgrids=true,
            grid=both,
            grid style=dashed,
            axis background/.style={fill=plotbackground},
            ybar=0pt, 
            bar width=6pt,
            enlarge x limits=0.1,  
            xtick=data,
            symbolic x coords={
                {Queen's Pawn\\Game: Modern},
                {Modern Def.},
                {Zukertort:\\Kingside F.},
                {Nimzo-Larsen\\Attack: Classical},
                {Nimzo-Larsen\\Attack: Modern}
            },
            xlabel style={font=\small},
            xticklabel style={rotate=90, anchor=east, align=right},
            ymax=100,
            ymin=0,
            ylabel style={font=\small},
            ymode=log,
            every tick label/.append style={font=\fontsize{8}{8}\selectfont},
        ]
        
        \addplot [fill=black, draw=black] coordinates {
            (Queen's Pawn\\Game: Modern, 6.82)
            (Modern Def., 5.23)
            (Zukertort:\\Kingside F., 3.48)
            (Nimzo-Larsen\\Attack: Classical, 3.10)
            (Nimzo-Larsen\\Attack: Modern, 2.89)
        };
        
        \draw (axis cs:{[normalized]\pgfkeysvalueof{/pgfplots/xmin}},0)
            -- (axis cs:{[normalized]\pgfkeysvalueof{/pgfplots/xmax}},0);
        
        \end{axis}
        \end{tikzpicture}
    \end{minipage} &
    \begin{minipage}{2.5cm}
        \begin{tikzpicture}[font=\small]
            \def\WCtest#1#2{
                \pgfmathparse{
                    \WCpercentage>30?"#1":"#2"
                }
                \pgfmathresult
            }
            \wheelchart[
                data=\WCtest{}{\WCvarC\\(\WCpercentage)}, 
                data sep=0, 
                radius={0}{1}, 
                lines={\WCpercentage>30?0:0.2},
                lines sep=0, 
                lines pos=0.8, 
                lines style={dotted,thick},
                wheel data=\WCtest{\WCvarC\\(\WCpercentage)}{}, 
                wheel data pos=0.4, 
            ]{
            49/chesscom-color/\includegraphics[width=3mm]{figures/icons/chesscom.png},
            6/lichess-color/\includegraphics[width=4mm]{figures/icons/lichess.png},
            45/pgnmentor-color/\includegraphics[width=7mm]{figures/icons/pgnmentor.png}
            }
        \end{tikzpicture}
    \end{minipage} \\
    \midrule
    \begin{minipage}{2cm}
        \centering
        \includegraphics[width=1.2cm]{figures/icons/nepomniachtchi.png}\\
        \ding{189} I. Nepomniachtchi
    \end{minipage} &
    \begin{minipage}{3.2cm}
        \begin{tikzpicture}
        \begin{axis}[
            width=0.8\linewidth, height=2cm,
            scale only axis,
            ymajorgrids=true,
            grid=both,
            grid style=dashed,
            axis background/.style={fill=plotbackground},
            ybar=0pt, 
            bar width=6pt,
            enlarge x limits=0.28,  
            xtick=data,
            symbolic x coords={
                Win,
                Draw,
                Lose
            },
            xlabel style={font=\small},
            ymax=100,
            ymin=0,
            ylabel style={font=\small},
            ytick={20,40,60,80,100},
            every tick label/.append style={font=\fontsize{8}{8}\selectfont},
        ]

        \addplot [fill=white, draw=black] coordinates {
            (Win, 56)
            (Draw, 21)
            (Lose, 23)
        };
        \addplot [fill=black, draw=black] coordinates {
            (Win, 47)
            (Draw, 23)
            (Lose, 30)
        };
        
        \draw (axis cs:{[normalized]\pgfkeysvalueof{/pgfplots/xmin}},0)
            -- (axis cs:{[normalized]\pgfkeysvalueof{/pgfplots/xmax}},0);
        
        \end{axis}
        \end{tikzpicture}
    \end{minipage} & 
    \begin{minipage}{5cm}
        \begin{tikzpicture}
        \begin{axis}[
            width=0.8\linewidth, height=1.2cm,
            scale only axis,
            ymajorgrids=true,
            grid=both,
            grid style=dashed,
            axis background/.style={fill=plotbackground},
            ybar=0pt, 
            bar width=6pt,
            enlarge x limits=0.1,  
            xtick=data,
            symbolic x coords={
                {Sicilian Def.:\\Najdorf},
                {Caro-Kann\\Def.},
                {Center Game:\\Berger},
                {Ruy Lopez:\\Closed M},
                {Ruy Lopez:\\Berlin Def. AB}
            },
            xlabel style={font=\small},
            xticklabel style={rotate=90, anchor=east, align=right},
            ymax=100,
            ymin=0,
            ylabel style={font=\small},
            ymode=log,
            every tick label/.append style={font=\fontsize{8}{8}\selectfont},
        ]
        
        \addplot [fill=white, draw=black] coordinates {
            (Sicilian Def.:\\Najdorf, 3.89)
            (Caro-Kann\\Def., 1.99)
            (Center Game:\\Berger, 1.68)
            (Ruy Lopez:\\Closed M, 1.61)
            (Ruy Lopez:\\Berlin Def. AB, 1.38)
        };
        
        \draw (axis cs:{[normalized]\pgfkeysvalueof{/pgfplots/xmin}},0)
            -- (axis cs:{[normalized]\pgfkeysvalueof{/pgfplots/xmax}},0);
        
        \end{axis}
        \end{tikzpicture}
    \end{minipage}
    \begin{minipage}{5cm}
        \begin{tikzpicture}
        \begin{axis}[
            width=0.8\linewidth, height=1.5cm,
            scale only axis,
            ymajorgrids=true,
            grid=both,
            grid style=dashed,
            axis background/.style={fill=plotbackground},
            ybar=0pt, 
            bar width=6pt,
            enlarge x limits=0.1,  
            xtick=data,
            symbolic x coords={
                {Zukertort},
                {Zukertort:\\Sicilian Invit.},
                {Sicilian Def.:\\Najdorf},
                {Zukertort:\\Kingside F.},
                {Sicilian Def.:\\Moscow},
            },
            xlabel style={font=\small},
            xticklabel style={rotate=90, anchor=east, align=right},
            ymax=100,
            ymin=0,
            ylabel style={font=\small},
            ymode=log,
            every tick label/.append style={font=\fontsize{8}{8}\selectfont},
        ]
        
        \addplot [fill=black, draw=black] coordinates {
            (Zukertort, 3.35)
            (Zukertort:\\Sicilian Invit., 2.37)
            (Sicilian Def.:\\Najdorf, 2.15)
            (Zukertort:\\Kingside F., 1.86)
            (Sicilian Def.:\\Moscow, 1.54)
        };
        
        \draw (axis cs:{[normalized]\pgfkeysvalueof{/pgfplots/xmin}},0)
            -- (axis cs:{[normalized]\pgfkeysvalueof{/pgfplots/xmax}},0);
        
        \end{axis}
        \end{tikzpicture}
    \end{minipage} &
    \begin{minipage}{2.5cm}
        \begin{tikzpicture}[font=\small]
            \def\WCtest#1#2{
                \pgfmathparse{
                    \WCpercentage>30?"#1":"#2"
                }
                \pgfmathresult
            }
            \wheelchart[
                data=\WCtest{}{\WCvarC\\(\WCpercentage)}, 
                data sep=0, 
                radius={0}{1}, 
                lines={\WCpercentage>30?0:0.2},
                lines sep=0, 
                lines pos=0.8, 
                lines style={dotted,thick},
                wheel data=\WCtest{\WCvarC\\(\WCpercentage)}{}, 
                wheel data pos=0.4, 
            ]{
            68/chesscom-color/\includegraphics[width=3mm]{figures/icons/chesscom.png},
            7/lichess-color/\includegraphics[width=4mm]{figures/icons/lichess.png},
            25/pgnmentor-color/\includegraphics[width=7mm]{figures/icons/pgnmentor.png}
            }
        \end{tikzpicture}
    \end{minipage} \\
    \midrule
    \begin{minipage}{2cm}
        \centering
        \includegraphics[width=1.2cm]{figures/icons/so.png}\\
        \ding{190} W. So
    \end{minipage} &
    \begin{minipage}{3.2cm}
        \begin{tikzpicture}
        \begin{axis}[
            width=0.8\linewidth, height=2cm,
            scale only axis,
            ymajorgrids=true,
            grid=both,
            grid style=dashed,
            axis background/.style={fill=plotbackground},
            ybar=0pt, 
            bar width=6pt,
            enlarge x limits=0.28,  
            xtick=data,
            symbolic x coords={
                Win,
                Draw,
                Lose
            },
            xlabel style={font=\small},
            ymax=100,
            ymin=0,
            ylabel style={font=\small},
            ytick={20,40,60,80,100},
            every tick label/.append style={font=\fontsize{8}{8}\selectfont},
        ]

        \addplot [fill=white, draw=black] coordinates {
            (Win, 54)
            (Draw, 26)
            (Lose, 20)
        };
        \addplot [fill=black, draw=black] coordinates {
            (Win, 46)
            (Draw, 30)
            (Lose, 24)
        };
        
        \draw (axis cs:{[normalized]\pgfkeysvalueof{/pgfplots/xmin}},0)
            -- (axis cs:{[normalized]\pgfkeysvalueof{/pgfplots/xmax}},0);
        
        \end{axis}
        \end{tikzpicture}
    \end{minipage} & 
    \begin{minipage}{5cm}
        \begin{tikzpicture}
        \begin{axis}[
            width=0.8\linewidth, height=1.2cm,
            scale only axis,
            ymajorgrids=true,
            grid=both,
            grid style=dashed,
            axis background/.style={fill=plotbackground},
            ybar=0pt, 
            bar width=6pt,
            enlarge x limits=0.1,  
            xtick=data,
            symbolic x coords={
                {Zukertort},
                {Ruy Lopez:\\Berlin Def. AB},
                {Nimzo-Larsen\\Attack: Classical},
                {Italian Game:\\Giuoco P.},
                {Van't Kruijs},
            },
            xlabel style={font=\small},
            xticklabel style={rotate=90, anchor=east, align=right},
            ymax=100,
            ymin=0,
            ylabel style={font=\small},
            ymode=log,
            every tick label/.append style={font=\fontsize{8}{8}\selectfont},
        ]
        
        \addplot [fill=white, draw=black] coordinates {
            (Zukertort, 8.67)
            (Ruy Lopez:\\Berlin Def. AB, 2.52)
            (Nimzo-Larsen\\Attack: Classical, 1.90)
            (Italian Game:\\Giuoco P., 1.66)
            (Van't Kruijs, 1.66)
        };
        
        \draw (axis cs:{[normalized]\pgfkeysvalueof{/pgfplots/xmin}},0)
            -- (axis cs:{[normalized]\pgfkeysvalueof{/pgfplots/xmax}},0);
        
        \end{axis}
        \end{tikzpicture}
    \end{minipage}
    \begin{minipage}{5cm}
        \begin{tikzpicture}
        \begin{axis}[
            width=0.8\linewidth, height=1.5cm,
            scale only axis,
            ymajorgrids=true,
            grid=both,
            grid style=dashed,
            axis background/.style={fill=plotbackground},
            ybar=0pt, 
            bar width=6pt,
            enlarge x limits=0.1,  
            xtick=data,
            symbolic x coords={
                {Ruy Lopez:\\Berlin Def. AB},
                {Indian Def.:\\Knights},
                {Zukertort},
                {Zukertort:\\Sicilian Invit.},
                {Ruy Lopez:\\Berlin Def. RG},
            },
            xlabel style={font=\small},
            xticklabel style={rotate=90, anchor=east, align=right},
            ymax=100,
            ymin=0,
            ylabel style={font=\small},
            ymode=log,
            every tick label/.append style={font=\fontsize{8}{8}\selectfont},
        ]
        
        \addplot [fill=black, draw=black] coordinates {
            (Ruy Lopez:\\Berlin Def. AB, 3.41)
            (Indian Def.:\\Knights, 3.01)
            (Zukertort, 2.83)
            (Zukertort:\\Sicilian Invit., 2.54)
            (Ruy Lopez:\\Berlin Def. RG, 2.35)
        };
        
        \draw (axis cs:{[normalized]\pgfkeysvalueof{/pgfplots/xmin}},0)
            -- (axis cs:{[normalized]\pgfkeysvalueof{/pgfplots/xmax}},0);
        
        \end{axis}
        \end{tikzpicture}
    \end{minipage} &
    \begin{minipage}{2.5cm}
        \begin{tikzpicture}[font=\small]
            \def\WCtest#1#2{
                \pgfmathparse{
                    \WCpercentage>30?"#1":"#2"
                }
                \pgfmathresult
            }
            \wheelchart[
                data=\WCtest{}{\WCvarC\\(\WCpercentage)}, 
                data sep=0, 
                radius={0}{1}, 
                lines={\WCpercentage>30?0:0.2},
                lines sep=0, 
                lines pos=0.8, 
                lines style={dotted,thick},
                wheel data=\WCtest{\WCvarC\\(\WCpercentage)}{}, 
                wheel data pos=0.4, 
            ]{
            62/chesscom-color/\includegraphics[width=3mm]{figures/icons/chesscom.png},
            10/lichess-color/\includegraphics[width=4mm]{figures/icons/lichess.png},
            28/pgnmentor-color/\includegraphics[width=7mm]{figures/icons/pgnmentor.png}
            }
        \end{tikzpicture}
    \end{minipage} \\
    \midrule
    \begin{minipage}{2cm}
        \centering
        \includegraphics[width=1.2cm]{figures/icons/vachier_lagrave.png}\\
        \ding{191} M. Vachier-Lagrave
    \end{minipage} &
    \begin{minipage}{3.2cm}
        \begin{tikzpicture}
        \begin{axis}[
            width=0.8\linewidth, height=2cm,
            scale only axis,
            ymajorgrids=true,
            grid=both,
            grid style=dashed,
            axis background/.style={fill=plotbackground},
            ybar=0pt, 
            bar width=6pt,
            enlarge x limits=0.28,  
            xtick=data,
            symbolic x coords={
                Win,
                Draw,
                Lose
            },
            xlabel style={font=\small},
            ymax=100,
            ymin=0,
            ylabel style={font=\small},
            ytick={20,40,60,80,100},
            every tick label/.append style={font=\fontsize{8}{8}\selectfont},
        ]

        \addplot [fill=white, draw=black] coordinates {
            (Win, 44)
            (Draw, 39)
            (Lose, 17)
        };
        \addplot [fill=black, draw=black] coordinates {
            (Win, 34)
            (Draw, 26)
            (Lose, 40)
        };
        
        \draw (axis cs:{[normalized]\pgfkeysvalueof{/pgfplots/xmin}},0)
            -- (axis cs:{[normalized]\pgfkeysvalueof{/pgfplots/xmax}},0);
        
        \end{axis}
        \end{tikzpicture}
    \end{minipage} & 
    \begin{minipage}{5cm}
        \begin{tikzpicture}
        \begin{axis}[
            width=0.8\linewidth, height=1.2cm,
            scale only axis,
            ymajorgrids=true,
            grid=both,
            grid style=dashed,
            axis background/.style={fill=plotbackground},
            ybar=0pt, 
            bar width=6pt,
            enlarge x limits=0.1,  
            xtick=data,
            symbolic x coords={
                {Ruy Lopez:\\Berlin Def. RG},
                {Ruy Lopez:\\Berlin Def. H},
                {Ruy Lopez:\\Closed M},
                {Modern Def.},
                {Ruy Lopez:\\Closed AM}
            },
            xlabel style={font=\small},
            xticklabel style={rotate=90, anchor=east, align=right},
            ymax=100,
            ymin=0,
            ylabel style={font=\small},
            ymode=log,
            every tick label/.append style={font=\fontsize{8}{8}\selectfont},
        ]
        
        \addplot [fill=white, draw=black] coordinates {
            (Ruy Lopez:\\Berlin Def. RG, 4.96)
            (Ruy Lopez:\\Berlin Def. H, 4.37)
            (Ruy Lopez:\\Closed M, 3.77)
            (Modern Def., 2.38)
            (Ruy Lopez:\\Closed AM, 1.79)
        };
        
        \draw (axis cs:{[normalized]\pgfkeysvalueof{/pgfplots/xmin}},0)
            -- (axis cs:{[normalized]\pgfkeysvalueof{/pgfplots/xmax}},0);
        
        \end{axis}
        \end{tikzpicture}
    \end{minipage}
    \begin{minipage}{5cm}
        \begin{tikzpicture}
        \begin{axis}[
            width=0.8\linewidth, height=1.5cm,
            scale only axis,
            ymajorgrids=true,
            grid=both,
            grid style=dashed,
            axis background/.style={fill=plotbackground},
            ybar=0pt, 
            bar width=6pt,
            enlarge x limits=0.1,  
            xtick=data,
            symbolic x coords={
                {Zukertort},
                {Zukertort:\\Sicilian Invit.},
                {Sicilian Def.:\\Najdorf},
                {Trompowsky\\Attack},
                {King's Indian\\Attack},
            },
            xlabel style={font=\small},
            xticklabel style={rotate=90, anchor=east, align=right},
            ymax=100,
            ymin=0,
            ylabel style={font=\small},
            ymode=log,
            every tick label/.append style={font=\fontsize{8}{8}\selectfont},
        ]
        
        \addplot [fill=black, draw=black] coordinates {
            (Zukertort, 3.77)
            (Zukertort:\\Sicilian Invit., 2.78)
            (Sicilian Def.:\\Najdorf, 2.38)
            (Trompowsky\\Attack, 2.18)
            (King's Indian\\Attack, 1.79)
        };
        
        \draw (axis cs:{[normalized]\pgfkeysvalueof{/pgfplots/xmin}},0)
            -- (axis cs:{[normalized]\pgfkeysvalueof{/pgfplots/xmax}},0);
        
        \end{axis}
        \end{tikzpicture}
    \end{minipage} &
    \begin{minipage}{2.5cm}
        \begin{tikzpicture}[font=\small]
            \def\WCtest#1#2{
                \pgfmathparse{
                    \WCpercentage>30?"#1":"#2"
                }
                \pgfmathresult
            }
            \wheelchart[
                data=\WCtest{}{\WCvarC\\(\WCpercentage)}, 
                data sep=0, 
                radius={0}{1}, 
                lines={\WCpercentage>30?0:0.2},
                lines sep=0, 
                lines pos=0.8, 
                lines style={dotted,thick},
                wheel data=\WCtest{\WCvarC\\(\WCpercentage)}{}, 
                wheel data pos=0.4, 
            ]{
            35/chesscom-color/\includegraphics[width=3mm]{figures/icons/chesscom.png},
            54/lichess-color/\includegraphics[width=4mm]{figures/icons/lichess.png},
            11/pgnmentor-color/\includegraphics[width=7mm]{figures/icons/pgnmentor.png}
            }
        \end{tikzpicture}
    \end{minipage} \\
    \bottomrule
    \end{tabular}
    }
\end{table}

\subsection{Behavioral stylometry}
\label{subsec:data_stylometry}

\paragraph{Training}
Stylometry techniques generally require supervised training on labeled examples of the form (\textit{played moves within a game}, \textit{player identity}).
We construct a dedicated training dataset in which each \gm is represented by an equal, uniform number of $1{,}000$ games---a threshold met by every player in our cohort.
These games are drawn exclusively from the training splits defined in \cref{subsec:data_lm}.
Capping each player at $1{,}000$ games is a deliberate design choice: training a player classifier on the full, unequalized collections would expose it to highly skewed class distributions, as \gms differ substantially in their total number of available games.
Such imbalance can bias the classifier toward over-represented players and obscure the evaluation of genuine stylometric discriminability.
Individual sets are formed via stratified sampling to guarantee a balanced 50:50 White--Black distribution, preventing color-based confounds.
This departs from prior work which uses unequal amounts of training data across players and imposes no control over color distribution.
We exclude games in which the opponent has committed a blunder affecting the subsequent course of play.
Following a significant error at the elite level, the remainder of the game becomes largely forced: the stronger side is expected to find the objectively best continuation at each turn, leaving little room for individual stylistic expression.
Such games are therefore uninformative as stylometric training examples, even though they remain valuable for language modeling purposes, where exposure to a diverse range of positions and continuations is desirable.

\paragraph{Evaluation}
Test sets are drawn directly from the held-out splits defined in \cref{subsec:data_lm}, and are not subject to any additional filtering beyond what was applied during language modeling preparation.
These evaluation games serve as gold-standard, player-attributed data against which the reliability of each stylometry metric is first established.
Only metrics that demonstrate sufficient discriminative power on this controlled benchmark can be subsequently applied to measure the style expressed by games generated by individual persona experts or by \mom when playing against Stockfish.

\clearpage
\newpage

\section{Behavioral stylometry baselines for persona validation}
\label{app:behavioral_stylometry}

This section discusses and evaluates the most prominent stylometry validation tools for persona experts within \mom, investigating the degree to which individual \gm styles are recoverable from model activations or predicted moves.
These approaches considered here are found to fall short of reliably separating stylistic signatures in our regime---a negative result that characterizes the difficulty of the problem and motivates the region-level activation-space analysis in the main paper.
\cref{subsec:stylometry_prior_work_vs_ours} situates our setting with respect to prior encoder-based stylometry work, with particular attention to~\cite{DBLP:conf/nips/McIlroy-YoungWS21} and~\cite{DBLP:journals/corr/abs-2502-14998}, and makes explicit why applying such methods to a small cohort of Super-\gms constitutes a substantially harder and previously unaddressed problem.
\cref{subsec:vision_encoder_baseline} introduces and evaluates a new vision-based encoder inspired by the architecture of~\cite{DBLP:conf/nips/McIlroy-YoungWS21}.

\subsection{Prior work and why our setting is harder}
\label{subsec:stylometry_prior_work_vs_ours}

We review the encoder-based stylometry framework of \cite{DBLP:conf/nips/McIlroy-YoungWS21}, which provides the main reference for identifying players from move sequences via learned representations.
A close inspection of this work is necessary for two reasons.
First, it provides the methodological blueprint for the vision-based encoder we later propose in \cref{subsec:vision_encoder_baseline}.
Second, and more critically, its ablation studies directly characterize the conditions under which current encoder-based stylometry breaks down---which are precisely the conditions that define our regime.

\paragraph{Task formalization}
The authors consider the following identification setting: given a set of query games played by a particular unknown player, the goal is to identify them from amongst a pool of candidate players, each associated with a set of reference games.
Two disjoint pools govern the evaluation:
\begin{itemize}
    \item \textit{Evaluation pool} $E$: the set of target players to be identified. Each player in $E$ is associated with a small set of query games, representing what is observed about the unknown player at inference time.
    \item \textit{Candidate pool} $C$: the set of possible identities. Each player in $C$ is associated with a small set of reference games, representing prior knowledge of that candidate.
\end{itemize}
The players to be identified are guaranteed to belong to $C$, so that $E \subseteq C$.
In general, $C$ can be substantially larger than $E$.

\paragraph{Model architecture}
The model proceeds in two stages, mapping individual moves to a compact player-level representation.
\begin{enumerate}
    \item \textit{Move encoding.} Each move is represented as a pair of $34$-channel $8{\times}8$ board tensors, encoding the position immediately before and after the player's move. The 34 channels are human-engineered: the first 24 encode specific piece types (6 per side across two positions), while the remaining 10 capture position metadata---including repetition count, castling rights, the active player's side, 50-move-rule count, and border information. Dual-board inputs are passed through a series of residual CNN blocks, producing a 320-dimensional feature vector for each move, which is then projected to a 1024-dimensional representation via a two-layer MLP.
    \item \textit{Game encoding.} The sequence of $1024$-dimensional move vectors for a game is processed by a ViT-based encoder with sinusoidal positional encoding. The resulting output vectors are averaged and projected to a $512$-dimensional game vector via an additional MLP layer.
\end{enumerate}
Game vectors are trained by adapting the GE2E loss~\cite{DBLP:conf/icassp/WanWPL18}, maximizing intra-player compactness and inter-player discrepancy.
The centroid vector of each player is computed by averaging all of their game vectors within the batch.

\paragraph{Data}
Training and evaluation draw from two sources:
\begin{itemize}
    \item \textit{Amateur players} (Elo 1,000--2,000): ${\sim}68$M games from $41{,}184$ Lichess players, bucketed by historical game volume as 1K--5K, 5K--10K, 10K--20K, 20K--30K, 30K--40K, and 40K+.
    \item \textit{High-ranked players}: ${\sim}9.6$M games from $2{,}264$ players drawn from the top-1,500 leaderboards of both Lichess and Chess.com.
\end{itemize}

For each source, players are divided into \emph{seen} and \emph{unseen} splits: seen players contribute games to model training, whereas unseen players are held out entirely.
Each player's games are further randomly partitioned into training games (80\%), reference games (10\%), and query games (10\%); training games are available only for seen players.
For the experiments reported in the main paper \cite{DBLP:conf/nips/McIlroy-YoungWS21}, training is conducted exclusively on the amateur source: $63.7$M games from $16{,}181$ seen players across all buckets; high-ranked players are reserved for evaluation (unseen split).

\paragraph{Evaluation protocol}
The predicted identity is the top-1 ranked candidate, and performance is measured by Precision@1 (P@1).
After training, the authors: (i) infer the player vector of each candidate in $C$ from their reference games; (ii) infer the player vector of the unknown target from their query games; (iii) compute the cosine distance between the unknown player vector and all candidate vectors; and (iv) return the nearest candidate as the predicted identity.

\paragraph{Main results}
All reported results use $100$ games per player for both the reference and query sets.
\begin{itemize}
    \item \textit{Amateur setting.} The candidate pool is $C = 2{,}844$ amateur players (2,266 seen $+$ 578 unseen) with 10K--40K+ games played, thereby excluding the majority of amateur players in the 1K--5K range (13,915 seen $+$ 24,425 unseen).
    With $E =$ 2,266 \emph{seen} players, the authors achieve P@1 $= 0.853$ (games starting at move $k{=}15$) and P@1 $= 0.982$ ($k{=}0$).
    \item \textit{High-ranked setting.} $C$ comprises the complete set of high-ranked players, with $E$ restricted to their \emph{unseen} split.
    With $k{=}15$, P@1 drops to $0.308$---a setting in which, as the authors note, the target players are not only unseen by the model, but belong to a substantially higher skill tier than any player encountered during training.
\end{itemize}

\paragraph{Failure modes}
McIlroy-Young et al.~\cite{DBLP:conf/nips/McIlroy-YoungWS21} further investigate, in their Appendix, three key causes of performance degradation beyond the high-ranked evaluation setting.
\begin{itemize}
    \item \errortag{Evaluating on low-volume amateur players.} Even with query and reference sets fixed at $100$ games each, broadening both $C$ and $E$ to include all seen and unseen players with a history depth of $\geq 1$K games---thereby introducing 1K--5K and 5K--10K buckets into the pool---causes P@1 to drop sharply from $0.860$ to $0.540$. Crucially, this decline is not caused by restricting evaluation exclusively to the low-data regime, but merely by adding lower-volume players to the high-data pool (10K--40K+), enlarging $C$ and $E$ with players of less distinct and more mutually overlapping styles. Players with large game histories tend to converge toward a stable repertoire and exhibit consistent decision-making habits, whereas low-volume players display more variable and less individuated behavior.
    \item \errortag{Training on a small player cohort.} To assess sensitivity to the number of training players, the authors train a model on only the $400$ players used in \cite{DBLP:conf/kdd/McIlroy-YoungW022}---an order-of-magnitude reduction in cohort size. The resulting model fails to retain discriminative power: on seen players in the main 10K--40K+ task, P@1 collapses to $0.204$ at $k{=}15$ and $0.464$ at $k{=}0$; on the high-ranked evaluation, P@1 falls further to $0.080$ at $k{=}15$ and $0.243$ at $k{=}0$. In the authors' words: \textit{\quotes{It does not generalize well and performs badly on the high-ranked players task. These results suggest that more than 400 players are needed to learn the space of chess-playing style.}}
    \item \errortag{Training exclusively on high-ranked players.} The authors also train a dedicated model on high-ranked players only---a smaller cohort than the main amateur training set---motivated by the hypothesis that direct in-distribution training might improve identification of elite players: \textit{\quotes{Having a model that performs well on high-skill players is of interest to the chess community. While our final model presented in the main text performs well above the baseline, we were interested in evaluating whether directly training on high-skill players would perform even better.}} With $E =$ \emph{seen} high-ranked players and $C =$ the complete high-ranked pool, P@1 reaches only $0.273$ at $k{=}15$ and $0.569$ at $k{=}0$, substantially underperforming the main amateur-trained model on this task. The authors conclude: \textit{\quotes{Similarly to the 400 model, it under-performs our final model by a large margin. Notably, the model explicitly trained on high-ranked players performs worse than our main model on the high-ranked players task, suggesting that high-ranked players are a harder set to learn from. This is consistent with our observation that they are harder to distinguish.}} This dramatic degradation at higher skill levels reflects a fundamental characteristic of elite chess: convergence toward objective optimality. While amateur players exhibit wide variation in their responses to common positions, \gms share an extensive body of theoretical knowledge and deeply ingrained pattern recognition. The objectively correct response to many positions is known and agreed upon at the highest level, leaving fewer opportunities for individual stylistic expression.
\end{itemize}

\paragraph{Remark on Omi et al.~\cite{DBLP:journals/corr/abs-2502-14998}}
Omi et al.\ scaled the above framework, extending the $>$90\% P@1 reported by McIlroy-Young et al.\ on seen players with full query games to a candidate pool of ${\sim}47{,}800$ players.
However, this work still relies on a large cohort of Lichess amateur players (Elo 1,000--2,000), dramatically increasing training data to ${\sim}244$M games.
Unlike~\cite{DBLP:conf/nips/McIlroy-YoungWS21}, no ablation is performed to assess performance degradation under reduced training data, higher target Elo, or smaller player cohorts.\\

\begin{challenge}
Our setting diverges from both~\cite{DBLP:conf/nips/McIlroy-YoungWS21} and~\cite{DBLP:journals/corr/abs-2502-14998} across every critical dimension.

We target \textbf{10 Super-\gms}---$40\times$ smaller than even the extreme 400-player case considered by McIlroy-Young et al.---at an \textbf{average Elo of 2,816}---over 800 points above the maximum level of the amateur population where stylometry has proven successful.

Our setting is further compounded by distributional differences. In the high-ranked games considered by McIlroy-Young et al., opening choice proves to be highly discriminative, yielding a performance delta of ${\sim}30$ percentage points. By contrast, as analyzed in detail in \cref{app:data}, \textbf{openings in our dataset are weakly discriminative}. Consequently, unlike the dataset of McIlroy-Young et al.---where openings serve as a strong stylometric \quotes{tell}---our regime requires detecting considerably subtler cues in the middlegame and endgame. This confirms that the two studies operate on statistically distinct populations: ours is characterized by a high degree of distributional overlap, driven by the objective optimality that governs elite-level play.

We study both training-free and training-dependent stylometry techniques.
For the latter, rather than exploiting the full historical record of each player, we consider a more accessible and appealing scenario in which \textbf{each \gm is represented by a small, equal number of 1,000 training games}---roughly two orders of magnitude fewer games than existing literature.
We refer the reader to \cref{subsec:data_stylometry} for full details on the training and evaluation data for all stylometry experiments.
\end{challenge}

\subsection{Vision encoder}
\label{subsec:vision_encoder_baseline}

\begin{figure*}[!t]
    \centering
    \includegraphics[width=\linewidth]{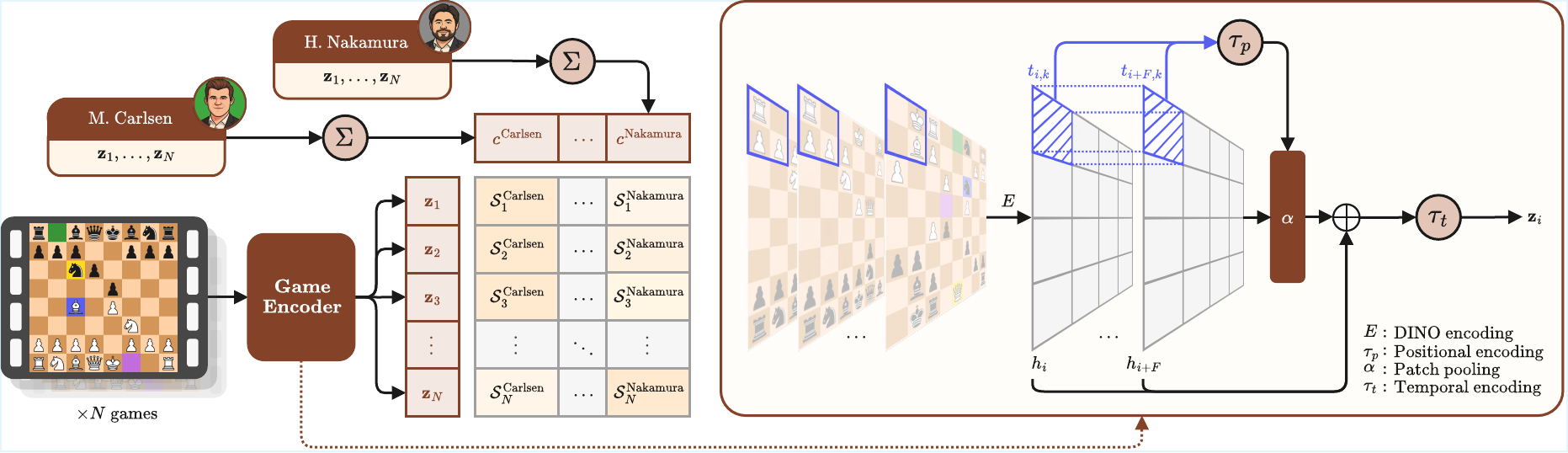}
    \caption{\textbf{Overview of the visual chess player identification system.} \textit{Left:} During training, game embeddings are processed through contrastive learning against \gm-specific centroids to enforce intra-player similarity and inter-player distinctiveness. \textit{Right:} The visual encoding pipeline processes consecutive chess board frames to extract and temporally aggregate spatial patch tokens (in blue), with positional and temporal encodings generating the final game embedding.}
    \label{fig:behavioral_stylometry}
\end{figure*}

We develop a training-dependent stylometry framework that leverages the knowledge of a pretrained vision transformer and embeds games into a representation space jointly capturing the spatial structure of the board and the temporal dynamics of play (Figure~\ref{fig:behavioral_stylometry}).

\paragraph{Model architecture}

Let a game $g$ by the $p$-th player be represented as a sequence of video frames $\smash{\mathcal{V}_g^p = \{I_1^p, I_2^p, \ldots, I_T^p\}}$, where $\smash{I_j^p}$ denotes the board configuration following the $j$-th move by $p$.
We extract fixed-length subsequences $\smash{\mathcal{F}_g^p}$ of size $F$ to standardize input sequences and expose stylistic variation across different stages of play.
Each frame $I_j^p$ is processed by a pretrained vision transformer $\smash{E_\psi}$ to produce $L$ patch-token embeddings $\smash{\{t_{j,k}^p\}_{k=1}^L}$.
To summarize information across space and time, we form two complementary views of these embeddings.
First, we aggregate temporally within the frame window $[i, i+F-1]$, producing patch representations that encapsulate local region evolution 
$\smash{r_k^p = \frac{1}{F} \sum_{j=i}^{i+F-1} t_{j,k}^p}$.
Second, we aggregate spatially within each frame $j$, producing frame representations $\smash{h_{j}^p = \frac{1}{L} \sum_{k=1}^{L} t_{j,k}^p}$.
We construct time-aware frame representations $\mathbf{e}_{j}^p$ by combining $h_{j}^p$ with an attention-weighted transformation $\alpha$ of the temporally-smoothed patch features, augmented with positional embeddings $\tau_p(j)$.
We process the resulting sequence through an LSTM network $\tau_t$.
The final sampled game embedding, $\mathbf{z}_g^p \in \mathbb{R}^{d}$, is defined as:
\begin{equation}
\begin{split}
\mathbf{e}_{j}^p &= h_{j}^p + \alpha \left ( \{r_{k}^p\}_{k=1}^L + \tau_p(j) \right ) \\ \mathbf{z}_g^p &= \tau_t(\{\mathbf{e}_{i}^p, \mathbf{e}_{i+1}^p, \ldots, \mathbf{e}_{i+F-1}^p\}).
\end{split}
\end{equation}
\noindent
We extend the GE2E loss~\cite{DBLP:conf/icassp/WanWPL18} by introducing additional regularization mechanisms to better organize $\mathcal{V}_g^p$ video frames into clusters of \gms' games embeddings.
This paradigm uses contrastive learning to group the learned representations of games from the same player into nearby regions of the embedding space.
Specifically, let us consider data batches composed of $N$ players and $M = \lvert G_p \rvert$ games per player.
For each combination, we compute the similarity score $\mathcal{S}^{p,q}_g$ between the embedding $\mathbf{z}^p_g$ of a specific game $g$ (by player $p$) and the centroid of player $q$, calculated as the mean of their respective game embeddings within the batch.
Crucially, to prevent the query game from inflating its similarity score, we omit $g$ from the centroid aggregation during self-comparisons:
\begin{equation}
\begin{split}
&\mathcal{S}_{g}^{p, q} = W \cdot \cos \left (\mathbf{z}_g^p,  c^{q}_{g} \right ) + b,\\
&\text{where} \quad
c^{q}_{g} = \begin{cases}
\frac{1}{M-1} \sum_{\tilde{g} \in G_q \setminus \{g\}} \mathbf{z}^q_{\tilde{g}} & \text{if } p = q  \\
c^q \coloneqq \frac{1}{M} \sum_{\tilde{g} \in G_q} \mathbf{z}^q_{\tilde{g}} & \text{if } p \neq q
\end{cases}
\end{split}
\end{equation}
with $W,b \in \mathbb{R}$ serving as learnable scaling parameters.
The training objective extends InfoNCE~\cite{Oord2018RepresentationLW}:
\begin{equation}
    \begin{split}
    \mathcal{L}_{\text{style}} &= -\frac{1}{NM} \sum_{p=1}^{N} \sum_{g \in G_p} \log \frac{\exp(\mathcal{S}_{g}^{p, p})}{\sum_{q=1}^{N} \exp(\mathcal{S}_{g}^{p, q})} + \frac{\lambda_m}{N(N-1)} \sum_{\substack{p,q=1\\p \neq q}}^{N} \max(0, \cos(c^{p}, c^{q}) + \mu) \\ &\quad +\frac{\lambda_c}{NM} \sum_{p=1}^{N} \sum_{g \in G_p} \left(1 - \cos(\mathbf{z}_g^p, c^{p}_{g})\right)
    \end{split}
\end{equation}
where $\lambda_m$ and $\lambda_c$ are regularization weights, and $\mu$ is the margin parameter.
The \emph{first term} is a standard contrastive term that maximizes the similarity of each game embedding to its own player centroid relative to all other centroids in the batch.
The \emph{second term} enforces margin-based inter-player separation, with larger values of $\mu$ imposing more aggressive constraints.
It penalizes any pair of player centroids $(c^p, c^q)$ whose cosine similarity exceeds $-\mu$.
With $\mu > 0$, the target threshold is a negative cosine similarity: this requires the angle between any two distinct player centroids to be strictly obtuse, which is a stronger condition than mere orthogonality ($\cos = 0$).
In geometric terms, the loss pushes centroids to point in broadly opposite directions in the embedding space, rather than simply being uncorrelated.
The \emph{third term} promotes intra-player compactness by minimizing the cosine distance between each game embedding and its player centroid.

\paragraph{Data preparation}

Each PGN string is transformed into a sequence of video frames, one per move played by the target \gm.
We render board positions using the \texttt{python-chess} library (v1.11.2).\footnote[11]{\url{https://python-chess.readthedocs.io/en/v1.11.2/}}
Each frame highlights two moves simultaneously: the preceding reply by the opponent, and the \gm's response to it.
This form of input image augmentation falls under the paradigm of \emph{visual prompting}~\cite{DBLP:conf/aaai/FrisoniMBM26}, which has demonstrated consistent benefits across a range of vision tasks and domains by guiding model attention toward task-relevant regions.
The board's perspective is standardized across all frames: it is always oriented so that the target \gm's pieces appear at the bottom, which involves rotating the board when they are playing as Black.

\paragraph{Implementation details}

We adopt \texttt{DINOv3}~\cite{DBLP:journals/corr/abs-2508-10104} (21.6M parameters) as the vision backbone $E_\psi$, a state-of-the-art self-supervised vision transformer for image encoding.
Prior to style-directed training, we subject $E_\psi$ to a preliminary fine-tuning stage on a supervised classification objective: predicting the board tile on which the next piece will be moved.
This stage runs for $15\text{k}$ steps and serves solely as a domain-adaptive initialization, steering the patch-token embeddings toward chess-relevant spatial features before contrastive training begins; the classification head is discarded afterwards.
We then fine-tune $E_\psi$ end-to-end with $\mathcal{L}_{\text{style}}$ for $25\text{k}$ steps, using in-batch negatives composed of $N{=}10$ players and $M{=}5$ games per player, with a frame window of $F{=}5$.

\paragraph{Results}

Figure~\ref{tab:vision_stylometry_results} reports the results of the vision-based stylometry model across efficacy and consistency dimensions.
Regarding \emph{efficacy}, in held-out real \gm games, the model achieves a mean P@5 of $0.80$ across players, peaking at $0.97$ for Vachier-Lagrave and $0.91$ for So.
Tightening the evaluation to P@3 reveals a significant drop, with the mean falling to $0.68$---a reminder that confident top-1 identification in this regime remains elusive.
Despite being sub-optimal, the results achieved are non-trivial given the distributional overlap and data constraints described in \cref{subsec:stylometry_prior_work_vs_ours}, which impose calibrated expectations.
When applied to synthetic games generated by individual persona experts against Stockfish, performance degrades only modestly, with mean P@5 dropping to $0.72$
This parallel between real and synthetic attribution suggests that persona experts have internalized identifiable, \gm-specific behavioral signatures that persist in generated play and transcend mere move emulation.
Regarding \emph{consistency}, for each \gm we repeatedly sample random subsets of increasing size from the pool of expert-generated games against Stockfish (from 30\% to 90\% of the total) and predict the corresponding style centroid for each subset.
The y-axis reports the relative change in cosine distance between centroids computed at each sampling proportion and the centroid computed at the baseline proportion of 30\%.
All \gms exhibit small relative drift across splits, with a maximum absolute deviation of ${\sim}4.2\%$ (Firouzja) and most players remaining well within ${\pm}2\%$ throughout.
This indicates that the style centroids induced by expert-generated games are internally stable: even small subsets already capture a consistent stylometric signature, and enlarging the sample produces only marginal shifts in the centroid position.
In other words, the representations are coherent rather than artifacts of sampling noise.

\begin{figure*}[!htb]
    \centering
    %
    \newcommand{\lmark}[1]{%
        \tikz[baseline=-0.5ex]{%
            \draw[color=#1, line width=1pt] (0,0) -- (0.22,0)
            node[pos=0.5, circle, fill=white, draw=#1, inner sep=0.6pt] {};}}
    %
    \begin{minipage}[c]{0.30\textwidth}
        \centering
        {\fontsize{8}{8}\selectfont\textbf{(a)~Efficacy in correctly identifying players on held-out real \gm games and expert-generated games against Stockfish}\par}
        \vspace{0.4cm}
        \begin{tikzpicture}[node font=\scriptsize]
        \begin{axis}[
            name=barplot,
            width=\linewidth, height=7.5cm,
            xmajorgrids=true,
            ymajorgrids=false,
            xminorgrids=true,
            minor x tick num=1,
            grid style={gray!30!white, dashed},
            minor grid style={gray!30!white, dashed},
            axis background/.style={fill=plotbackground},
            every y tick/.style={draw=none},
            xbar=0pt,
            bar width=2pt,
            enlarge y limits=0.08,
            xmin=0.37, xmax=1,
            xlabel={Precision},
            xtick={0.4, 0.6, 0.8, 1.0},
            ytick={1,...,10},
            yticklabels={\ding{182},\ding{183},\ding{184},\ding{185},
                         \ding{186},\ding{187},\ding{188},\ding{189},
                         \ding{190},\ding{191}},
            yticklabel style={font=\fontsize{9}{9}\selectfont},
            y dir=reverse,
            point meta=explicit,
            axis on top=true,
            axis line style={draw=black},
        ]
        \addplot [draw=none, fill=crayonRed] coordinates {
            (0.42,1)(0.63,2)(0.70,3)(0.71,4)(0.61,5)
            (0.61,6)(0.58,7)(0.83,8)(0.84,9)(0.89,10)};
        \addplot [draw=crayonRed!80, fill=crayonRed!30,
                  pattern={Lines[distance=1.2pt,angle=45]},
                  pattern color=crayonRed!80] coordinates {
            (0.395,1)(0.58,2)(0.6,3)(0.67,4)(0.58,5)
            (0.55,6)(0.5,7)(0.73,8)(0.72,9)(0.79,10)};
        \addplot [draw=none, fill=crayonBlue] coordinates {
            (0.71,1)(0.69,2)(0.72,3)(0.74,4)(0.69,5)
            (0.63,6)(0.60,7)(0.85,8)(0.89,9)(0.95,10)};
        \addplot [draw=crayonBlue!80, fill=crayonBlue!30,
                  pattern={Lines[distance=1.2pt,angle=45]},
                  pattern color=crayonBlue!80] coordinates {
            (0.67,1)(0.6,2)(0.68,3)(0.69,4)(0.64,5)
            (0.58,6)(0.57,7)(0.78,8)(0.82,9)(0.85,10)};
        \addplot [draw=crayonYellow, fill=crayonYellow] coordinates {
            (0.74,1)(0.71,2)(0.73,3)(0.76,4)(0.76,5)
            (0.81,6)(0.71,7)(0.87,8)(0.91,9)(0.97,10)};
        \addplot [draw=crayonYellow!80, fill=crayonYellow!30,
                  pattern={Lines[distance=1.2pt,angle=45]},
                  pattern color=crayonYellow!80] coordinates {
            (0.68,1)(0.63,2)(0.69,3)(0.7,4)(0.67,5)
            (0.72,6)(0.62,7)(0.81,8)(0.86,9)(0.86,10)};
        \end{axis}
        \end{tikzpicture}\\[0.4cm]
        %
        {\fontsize{7}{8.5}\selectfont
        \begin{tabular}{@{}l@{\hspace{5pt}}l@{\hspace{5pt}}l@{}}
            \legendBox{crayonRed}   P@3 &
            \legendBox{crayonBlue}  P@4 &
            \legendBox{crayonYellow}P@5 \\[3pt]
            \multicolumn{3}{@{}l@{}}{%
                \tikz[baseline=0.1ex]{%
                    \fill[gray!65] (0,0) rectangle (0.22,0.15);}~Real games
                \hspace{6pt}%
                \tikz[baseline=0.1ex]{%
                    \draw[gray!65, line width=0.5pt] (0,0) rectangle (0.22,0.15);
                    \fill[gray!20, pattern=north east lines,
                          pattern color=gray!65] (0,0) rectangle (0.22,0.15);}~Synthetic games%
            }\\
        \end{tabular}}
    \end{minipage}%
    %
    \hfill
    \begin{minipage}[c]{0.01\textwidth}
        \centering
        \rule{0.5pt}{9.5cm}
    \end{minipage}
    \hfill
    %
    \begin{minipage}[c]{0.63\textwidth}
        \centering
        {\fontsize{8}{8}\selectfont\textbf{(b)~Centroid stability across random subsamples\\of expert-generated games against Stockfish}\par}
        \vspace{0.4cm}
        \begin{tikzpicture}
        \begin{axis}[
            width=1.02\linewidth, height=4cm,
            ymajorgrids=true,
            xmajorgrids=true,
            grid=both,
            grid style=dashed,
            axis background/.style={fill=plotbackground},
            xmin=29, xmax=91,
            xtick={30,40,50,60,70,80,90},
            xlabel={Proportion of Generated Games (\%)},
            xlabel style={font=\footnotesize},
            ymin=-4, ymax=7,
            ytick={-4,-2,0,2,4,6},
            yticklabels={$-4$,$-2$,$0$,$+2$,$+4$,$+6$},
            ylabel={\shortstack{$\Delta$ Cosine Distance\\across Subsamples (\%)}},
            ylabel style={font=\footnotesize},
            every tick label/.append style={font=\fontsize{7}{7}\selectfont},
            extra y ticks={0},
            extra y tick style={grid=major, grid style={black, thick, dashed}},
        ]
        \addplot [name path=U1, draw=none, forget plot] coordinates {
            (30,1.77)(40,2.47)(50,3.36)(60,4.06)(70,3.71)(80,3.53)(90,3.89)};
        \addplot [name path=L1, draw=none, forget plot] coordinates {
            (30,-1.77)(40,-0.35)(50,0.88)(60,1.94)(70,1.59)(80,1.77)(90,2.12)};
        \addplot [color=blue!80!white, fill opacity=0.15, draw=none, forget plot]
            fill between [of=U1 and L1];
        \addplot [color=blue!80!white, line width=1.2pt,
                  mark=*, mark size=1.2pt, mark options={fill=white}] coordinates {
            (30,0.00)(40,1.06)(50,2.12)(60,3.00)(70,2.65)(80,2.65)(90,3.00)};
        \addplot [name path=U2, draw=none, forget plot] coordinates {
            (30,1.43)(40,1.08)(50,2.87)(60,3.41)(70,3.94)(80,3.58)(90,3.23)};
        \addplot [name path=L2, draw=none, forget plot] coordinates {
            (30,-1.43)(40,-1.43)(50,-0.00)(60,0.54)(70,1.08)(80,1.43)(90,1.08)};
        \addplot [color=red!80!white, fill opacity=0.15, draw=none, forget plot]
            fill between [of=U2 and L2];
        \addplot [color=red!80!white, line width=1.2pt,
                  mark=*, mark size=1.2pt, mark options={fill=white}] coordinates {
            (30,0.00)(40,-0.18)(50,1.43)(60,1.97)(70,2.51)(80,2.51)(90,2.15)};
        \addplot [name path=U3, draw=none, forget plot] coordinates {
            (30,1.00)(40,-0.17)(50,-1.34)(60,-1.50)(70,-0.83)(80,-1.17)(90,-0.50)};
        \addplot [name path=L3, draw=none, forget plot] coordinates {
            (30,-1.00)(40,-2.17)(50,-3.01)(60,-3.17)(70,-2.50)(80,-2.84)(90,-2.17)};
        \addplot [color=green!60!black, fill opacity=0.15, draw=none, forget plot]
            fill between [of=U3 and L3];
        \addplot [color=green!60!black, line width=1.2pt,
                  mark=*, mark size=1.2pt, mark options={fill=white}] coordinates {
            (30,0.00)(40,-1.17)(50,-2.17)(60,-2.34)(70,-1.67)(80,-2.00)(90,-1.34)};
        \addplot [name path=U4, draw=none, forget plot] coordinates {
            (30,1.33)(40,1.66)(50,1.00)(60,0.17)(70,-0.33)(80,-0.33)(90,-0.83)};
        \addplot [name path=L4, draw=none, forget plot] coordinates {
            (30,-1.33)(40,-0.66)(50,-1.33)(60,-2.16)(70,-2.65)(80,-2.65)(90,-3.15)};
        \addplot [color=violet!80!white, fill opacity=0.15, draw=none, forget plot]
            fill between [of=U4 and L4];
        \addplot [color=violet!80!white, line width=1.2pt,
                  mark=*, mark size=1.2pt, mark options={fill=white}] coordinates {
            (30,0.00)(40,0.50)(50,-0.17)(60,-1.00)(70,-1.49)(80,-1.49)(90,-1.99)};
        \addplot [name path=U5, draw=none, forget plot] coordinates {
            (30,1.65)(40,4.03)(50,5.31)(60,4.40)(70,4.95)(80,5.68)(90,5.86)};
        \addplot [name path=L5, draw=none, forget plot] coordinates {
            (30,-1.65)(40,0.73)(50,2.01)(60,1.10)(70,1.65)(80,2.38)(90,2.56)};
        \addplot [color=brown!80!white, fill opacity=0.15, draw=none, forget plot]
            fill between [of=U5 and L5];
        \addplot [color=brown!80!white, line width=1.2pt,
                  mark=*, mark size=1.2pt, mark options={fill=white}] coordinates {
            (30,0.00)(40,2.38)(50,3.66)(60,2.75)(70,3.30)(80,4.03)(90,4.21)};
        \addplot [name path=U6, draw=none, forget plot] coordinates {
            (30,1.03)(40,1.72)(50,2.06)(60,0.69)(70,0.00)(80,0.69)(90,0.34)};
        \addplot [name path=L6, draw=none, forget plot] coordinates {
            (30,-1.03)(40,-0.34)(50,-0.00)(60,-1.37)(70,-2.06)(80,-1.37)(90,-1.72)};
        \addplot [color=magenta!80!white, fill opacity=0.15, draw=none, forget plot]
            fill between [of=U6 and L6];
        \addplot [color=magenta!80!white, line width=1.2pt,
                  mark=*, mark size=1.2pt, mark options={fill=white}] coordinates {
            (30,0.00)(40,0.69)(50,1.03)(60,-0.34)(70,-1.03)(80,-0.34)(90,-0.69)};
        \addplot [name path=U7, draw=none, forget plot] coordinates {
            (30,1.27)(40,3.27)(50,3.27)(60,2.36)(70,1.82)(80,1.64)(90,1.27)};
        \addplot [name path=L7, draw=none, forget plot] coordinates {
            (30,-1.27)(40,0.73)(50,0.73)(60,-0.18)(70,-0.73)(80,-0.55)(90,-0.91)};
        \addplot [color=gray!80!white, fill opacity=0.15, draw=none, forget plot]
            fill between [of=U7 and L7];
        \addplot [color=gray!80!white, line width=1.2pt,
                  mark=*, mark size=1.2pt, mark options={fill=white}] coordinates {
            (30,0.00)(40,2.00)(50,2.00)(60,1.09)(70,0.55)(80,0.55)(90,0.18)};
        \addplot [name path=U8, draw=none, forget plot] coordinates {
            (30,1.05)(40,-0.35)(50,-1.05)(60,0.00)(70,-0.18)(80,-0.35)(90,-0.18)};
        \addplot [name path=L8, draw=none, forget plot] coordinates {
            (30,-1.05)(40,-2.46)(50,-3.16)(60,-2.11)(70,-2.28)(80,-2.11)(90,-2.28)};
        \addplot [color=lime!60!black, fill opacity=0.15, draw=none, forget plot]
            fill between [of=U8 and L8];
        \addplot [color=lime!60!black, line width=1.2pt,
                  mark=*, mark size=1.2pt, mark options={fill=white}] coordinates {
            (30,0.00)(40,-1.40)(50,-2.11)(60,-1.05)(70,-1.23)(80,-1.23)(90,-1.23)};
        \addplot [name path=U9, draw=none, forget plot] coordinates {
            (30,1.21)(40,1.03)(50,1.90)(60,1.90)(70,2.07)(80,2.93)(90,3.10)};
        \addplot [name path=L9, draw=none, forget plot] coordinates {
            (30,-1.21)(40,-1.38)(50,-0.52)(60,-0.52)(70,-0.34)(80,0.52)(90,0.69)};
        \addplot [color=cyan!80!black, fill opacity=0.15, draw=none, forget plot]
            fill between [of=U9 and L9];
        \addplot [color=cyan!80!black, line width=1.2pt,
                  mark=*, mark size=1.2pt, mark options={fill=white}] coordinates {
            (30,0.00)(40,-0.17)(50,0.69)(60,0.69)(70,0.86)(80,1.72)(90,1.90)};
        \addplot [name path=U10, draw=none, forget plot] coordinates {
            (30,1.05)(40,0.87)(50,0.87)(60,0.35)(70,0.52)(80,0.87)(90,1.40)};
        \addplot [name path=L10, draw=none, forget plot] coordinates {
            (30,-1.05)(40,-1.22)(50,-1.22)(60,-1.75)(70,-1.22)(80,-1.22)(90,-0.70)};
        \addplot [color=orange!80!white, fill opacity=0.15, draw=none, forget plot]
            fill between [of=U10 and L10];
        \addplot [color=orange!80!white, line width=1.2pt,
                  mark=*, mark size=1.2pt, mark options={fill=white}] coordinates {
            (30,0.00)(40,-0.17)(50,-0.17)(60,-0.70)(70,-0.35)(80,-0.17)(90,0.35)};
        \end{axis}
        \end{tikzpicture}\\[0.4cm]
        %
        {\fontsize{7}{7}\selectfont
        \arrayrulecolor{tablebordercolor}
        \begin{tabular}{@{}l@{\hspace{8pt}}l@{\hspace{8pt}}l@{}}
            \lmark{blue!80!white}    {\fontsize{9}{9}\selectfont\ding{182}}~Anand &
            \lmark{red!80!white}     {\fontsize{9}{9}\selectfont\ding{183}}~Aronian &
            \lmark{green!60!black}   {\fontsize{9}{9}\selectfont\ding{184}}~Carlsen \\[1pt]
            \lmark{violet!80!white}  {\fontsize{9}{9}\selectfont\ding{185}}~Caruana &
            \lmark{brown!80!white}   {\fontsize{9}{9}\selectfont\ding{186}}~Firouzja &
            \lmark{magenta!80!white} {\fontsize{9}{9}\selectfont\ding{187}}~Giri \\[1pt]
            \lmark{gray!80!white}    {\fontsize{9}{9}\selectfont\ding{188}}~Nakamura &
            \lmark{lime!60!black}    {\fontsize{9}{9}\selectfont\ding{189}}~Nepomniachtchi &
            \lmark{cyan!80!black}    {\fontsize{9}{9}\selectfont\ding{190}}~So \\[1pt]
            \lmark{orange!80!white}  {\fontsize{9}{9}\selectfont\ding{191}}~Vachier-Lagrave & & \\
        \end{tabular}}
    \end{minipage}%
    \caption{\textbf{Results of the vision-based behavioral stylometry model.} \textit{(a) Efficacy}: Precision at $k$ (P@$k$, $k \in \{3,4,5\}$) in identifying the target grandmaster from held-out real \gm games (solid bars) and expert-generated games played against Stockfish (hatched bars). \textit{(b) Consistency}: Relative change in cosine distance between expert-specific style centroids computed from random subsamples of increasing size.}
    \label{tab:vision_stylometry_results}
\end{figure*}

\paragraph{Motivations for departures from prior literature}

\emph{Why a new encoder-based solution.}
The code and model weights of \cite{DBLP:conf/nips/McIlroy-YoungWS21} are not publicly available, and direct contact with the authors was unsuccessful.
Rather than risking unfair baseline reporting through an unverifiable re-implementation, and given that the ablation studies reviewed in \cref{subsec:stylometry_prior_work_vs_ours} already demonstrate that this family of approaches degrades severely under the exact conditions that define our regime, we instead design a new solution that directly targets these known failure modes.
\emph{Visual input representation.}
Previous work constructs input representations via domain-specific feature engineering.
While effective, this requires substantial domain expertise and may inadvertently encode assumptions about which features carry stylistic signal.
We instead operate on raw visual input, eliminating manual feature engineering and allowing the model to discover stylistically relevant patterns directly from pixel-level observations.
The visual modality also aligns with how human experts perceive and recall chess positions, potentially capturing gestalt properties of board configurations that symbolic encodings may miss.
This decision is additionally justified by the fact that the visual formulation enables controlled, player-specific encoding, makes spatial reasoning explicit, and supports augmentation strategies that naturally take advantage of the 2D board layout.
\emph{Architecture and hyperparameter selection.}
The final architecture for this secondary analysis was determined through over 100 controlled experiments spanning architectural and hyperparameter variations.
Removing the LSTM in favor of mean pooling, replacing learned positional encodings with fixed sinusoidal alternatives, or simplifying spatial aggregation (e.g., CLS-only or patch-mean without dual-axis attention) each consistently reduced retrieval accuracy and increased contrastive loss; all retained components were selected on the basis of consistent empirical gains.
\emph{Retrieval-based evaluation instead of top-1 accuracy.}
We adopt a retrieval-based protocol, measuring whether the target \gm appears among the $k$ nearest centroids for each expert's generated games (P@$k$, $k \in \{3,4,5\}$).
This formulation acknowledges that elite players exhibit substantially overlapping stylistic signatures.
Top-$k$ retrieval captures whether meaningful style correspondence exists without demanding that each \gm be perfectly separable from all others.

\clearpage
\newpage

\section{Reinforcement learning for improved legality}
\label{app:rl}

Chess language models trained with SSL must infer move legality solely from statistical patterns over game transcripts.
This leaves models susceptible to generating illegal moves due to overfitting or distributional shift at generation time.
The presented section examines whether GRPO can complement SSL by instilling higher \emph{intrinsic} legality---that is, legal play as a native model behavior.
Intrinsic legality is not merely a theoretical concern: its value must be assessed alongside playing strength rather than assumed to be compensable at inference time.
Kaggle's Game Arena on chess, for instance, provides no legal move list and allows at most three retries per position; failure to produce a legal move within four total attempts results in disqualification and forfeiture of the game.

\paragraph{Why not optimize for strength}

An alternative to legality-based RL would be to optimize directly for strength---either via move-quality rewards derived from Stockfish centipawn evaluation, or via long-horizon game outcome signals.
Both approaches are fundamentally incompatible with our research objectives.
They push all experts toward convergence on objectively optimal play, necessarily erasing the \gm-specific behavioral patterns acquired during SSL.
Outcome optimization additionally requires rolling out complete game trajectories via self-play, with hundreds of sequential forward passes per game.
In contrast, our legality-based reward maintains a clean separation between rule adherence and strategic choice.
The model receives strong feedback for generating legal moves but the choice remains governed by the patterns learned during SSL from each \gm's actual games.
This design preserves stylistic diversity while improving robustness, in line with our broader goal of building interpretable, human-aligned chess AI rather than maximally strong models.

\subsection{Method}

\paragraph{GRPO formulation}

We refine each expert $\varepsilon_{\phi_p}$ using the GRPO algorithm.
Our RL stage operates on a single-step basis: the reward evaluates only the syntactic correctness and legality of the next predicted move, with no lookahead.
For each position, GRPO generates multiple candidate moves from the current policy, computes rewards, and then optimizes the policy to increase the relative probability of high-reward moves over low-reward alternatives within the same context.
Given a board state $s$, we sample a set of $M$ candidate moves $\{m_i\}_{i=1}^M \sim \varepsilon_{\phi_p}(s)$ via temperature-controlled decoding.
Each candidate is evaluated along two axes: (i)~\textit{syntactic correctness} $\rho_\text{synt}$, whether the predicted PGN substring is well-formed; and (ii)~\textit{legality} $\rho_\text{leg}$, whether the move conforms to chess rules.
Consistent with the notation of~\cite{DBLP:journals/corr/abs-2402-03300}, the policy optimization objective is:
\begin{equation}
\begin{split}
\mathcal{J}_\text{GRPO} &= \frac{1}{M} \sum_{i=1}^{M} \min \left(\mathcal{R}_{i} \cdot \hat{A}_{i};\,
    \text{clip}_{1\pm\epsilon}(\mathcal{R}_{i}) \cdot \hat{A}_{i} \right)
    - \beta\,\mathbb{D}_{KL}\!\left(\varepsilon_{\phi_p} \Vert \varepsilon_{\phi_p^\text{old}}\right), \\
\text{with}\;\; &\mathcal{R}_{i} = \frac{\varepsilon_{\phi_p}(m_{i}\mid s)}{\varepsilon_{\phi_p^\text{old}}(m_{i}\mid s)}, \qquad
\hat{A}_{i} = \frac{\rho_\text{synt}(m_{i}) + \rho_\text{leg}(s, m_{i}) - \mu_\mathbf{r}}{\sigma_\mathbf{r}},
\end{split}
\end{equation}
where the advantage $\hat{A}_{i}$ is normalized across the batch of candidate moves.
All tokens forming a single candidate move inherit the same cumulative reward.

\paragraph{Reward structure}

The reward function decomposes additively over two components that jointly assess the quality of a candidate move $m_i$ generated from board state $s$:
\begin{equation}
r(s, m_i) = \rho_{\text{synt}}(m_i) + \rho_{\text{leg}}(s, m_i).
\end{equation}
The \emph{syntactic correctness} component $\rho_{\text{synt}}(m_i)$ is defined as follows:
\begin{equation}
\rho_{\text{synt}}(m_i) = \begin{cases}
-1.0 & \text{if } m_i \text{ is malformed PGN,} \\
\phantom{-}0.0 & \text{if } m_i \text{ is well-formed PGN.}
\end{cases}
\end{equation}
It penalizes outputs that violate PGN conventions---invalid piece notation, missing coordinates, or malformed special-move indicators---which SSL models occasionally produce, particularly under temperature sampling or in out-of-distribution positions.
The \emph{legality} component $\rho_{\text{leg}}(s, m_i)$ provides finer-grained feedback for syntactically correct candidates:
\begin{equation}
\rho_{\text{leg}}(s, m_i) = \begin{cases}
1.0 & \text{if } m_i \in \mathcal{L}(s), \\
0.5 - d_{\text{edit}}(m_i,\, m_i^*) & \text{otherwise,}
\end{cases}
\end{equation}
where $\mathcal{L}(s)$ denotes the set of legal moves in state $s$, and $m_i^* = \arg\min_{m' \in \mathcal{L}(s)} d_{\text{edit}}(m_i, m')$ is the closest legal move under normalized edit distance $d_{\text{edit}} \in [0, 1]$.
This formulation grants partial credit to illegal moves that closely resemble a legal alternative: for instance, \texttt{Nf6} when \texttt{Ne6} is correct receives a higher reward than a completely nonsensical string, providing a smoother gradient signal than a binary legality indicator.
The combined reward $r(s, m_i)$ spans the range $[-1, 1]$: a value of $-1$ indicates malformed syntax; values in $(-1, 0)$ indicate syntactically valid but illegal moves, with partial credit scaled by proximity to the nearest legal alternative; and a value of $1$ indicates a fully legal move.

\paragraph{Implementation details}

We fine-tune each expert $\varepsilon_{\phi_p}$ independently for $6{,}000$ steps using groups of $M{=}8$ candidate moves per position, a batch size of $B{=}8$ positions, and a learning rate of $6{\times}10^{-7}$.

\subsection{Training dynamics and convergence analysis}

\begin{figure}[t]
\centering
\includegraphics[width=\linewidth]{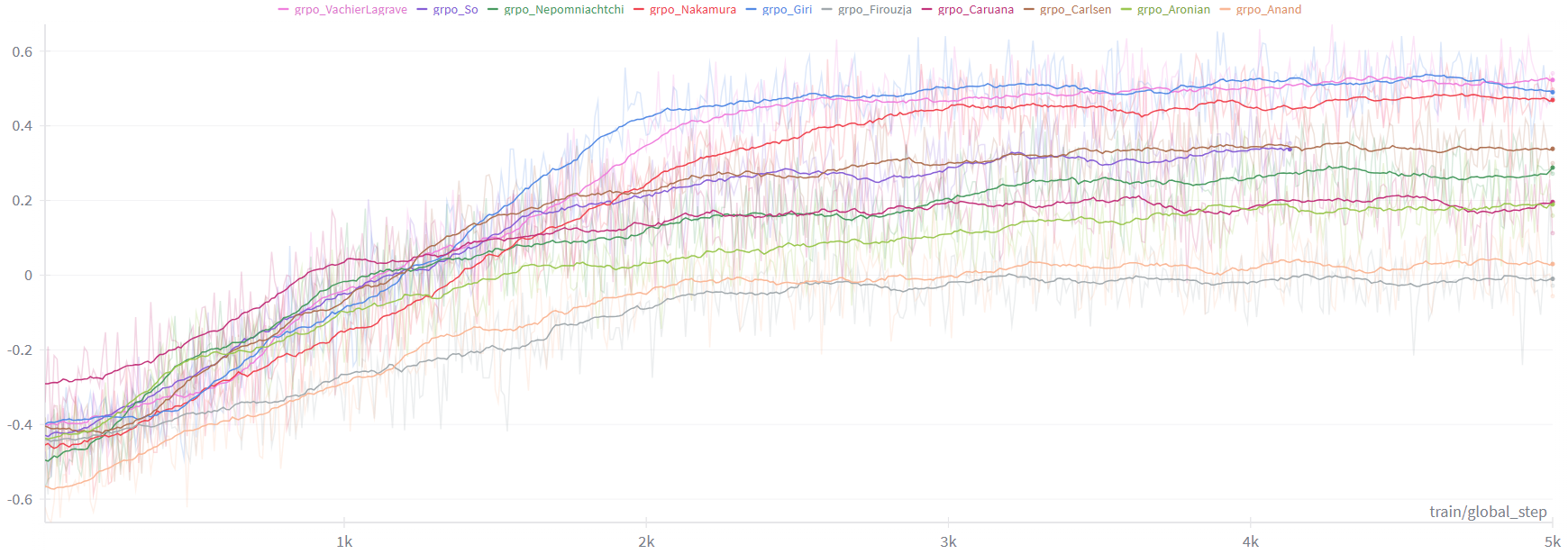}
\caption{\textbf{GRPO training legality reward curves for all ten \gm experts.} Each curve reports the mean reward $r(s, m_i)$ over training steps, averaged across $B \times M$ evaluations per step.}
\label{fig:grpo_curves}
\end{figure}

Figure~\ref{fig:grpo_curves} presents the complete reward trajectories for all \gm experts.
All experts exhibit monotonically positive reward growth throughout training, transitioning from negative starting values---indicating frequent illegal moves under the SSL initialization---to positive convergence points, confirming that GRPO consistently improves intrinsic legality across all personas.
Starting rewards range from $-0.603$ (Anand) to $-0.279$ (Caruana), while final rewards span from $0.006$ (Firouzja) to $0.498$ (Nakamura).
The magnitude of improvement varies across experts, revealing an interaction pattern between SSL initialization quality and RL effectiveness.
Experts with stronger SSL baselines (e.g., Caruana, $-0.279$) require less correction from the reward signal and exhibit moderate RL gains ($0.190$), suggesting diminishing returns when the initialization is already robust.
Conversely, experts with weaker baselines (e.g., Anand, $-0.603$) achieve lower final rewards ($0.022$), indicating that representational deficits inherited from SSL can limit what RL can recover within a fixed training budget.
Nakamura presents an interesting middle ground, combining a moderately weak SSL foundation ($-0.466$) with the highest final reward ($0.498$).
This heterogeneity in convergence profiles is not an artifact to be eliminated.
Rather, it reflects genuinely complementary expert behaviors that motivate the \mom architecture.
Experts with high final rewards---such as Nakamura ($0.498$) and Giri---provide reliable
legal move generation in complex positions.
Experts with moderate rewards---such as Caruana ($0.190$) and Carlsen---retain a stronger balance between legality and stylistic fidelity to their target \gm.
The learned router can dynamically allocate positions to the expert best suited for the current game context.

\subsection{Results}

\paragraph{Does SSL + RL result in greater legality than SSL alone?}

Figure~\ref{fig:ssl_rl} reports legality distributions for SSL and SSL+RL across all \gm experts, pooled over ten runs against Stockfish level~0.
GRPO consistently shifts the interquartile range upward for nine out of ten experts, confirming that legality-based RL instills rule adherence.
\begin{figure}[!t]
    \centering
    \begin{tikzpicture}
        \fontsize{7}{7}\selectfont
        \node[draw=none, inner sep=2pt, align=center] {
        \begin{tabular}{ll}
            \tikz\fill[ssl-color, fill opacity=0.7, draw=black] (0,0) rectangle (0.22,0.22); SSL &
            \tikz\fill[sslrl-color, fill opacity=0.7, draw=black] (0,0) rectangle (0.22,0.22); SSL+RL \\
        \end{tabular}
        };
    \end{tikzpicture}\\[1mm]
    \begin{tikzpicture}
    \begin{axis}[
        width=\linewidth,
        height=6cm,
        grid=both,
        grid style={gray!30, dashed},
        axis background/.style={fill=plotbackground},
        xmin=65,
        boxplot/draw direction=x,
        xlabel=Legality (\%),
        xticklabel style={font=\fontsize{7}{7}\selectfont},
        xlabel style={font=\fontsize{9}{9}\selectfont},
        y dir=reverse,
        ytick={3, 6, 9, 12, 15, 18, 21, 24, 27, 30},
        yticklabels={\ding{182},\ding{183},\ding{184},\ding{185},\ding{186},
                     \ding{187},\ding{188},\ding{189},\ding{190},\ding{191}},
        yticklabel style={font=\fontsize{9}{9}\selectfont},
        ylabel style={font=\fontsize{9}{9}\selectfont},
        ymin=1.0, ymax=32.5,
        ylabel=Experts,
    ]
    \pgfmathsetmacro{\s}{0.45}
    \addplot+[boxplot prepared={lower whisker=70,lower quartile=73.3,median=78,upper quartile=81.8,upper whisker=85,box extend=0.65}, draw=black,line width=0.8pt,fill=ssl-color,fill opacity=0.7,solid,boxplot/draw position={3-\s}] coordinates {};
    \addplot+[boxplot prepared={lower whisker=75,lower quartile=78.3,median=81.5,upper quartile=82.8,upper whisker=84,box extend=0.65}, draw=black,line width=0.8pt,fill=sslrl-color,fill opacity=0.7,solid,boxplot/draw position={3+\s}] coordinates {};
    \addplot+[boxplot prepared={lower whisker=72,lower quartile=76,median=79,upper quartile=81.5,upper whisker=86,box extend=0.65}, draw=black,line width=0.8pt,fill=ssl-color,fill opacity=0.7,solid,boxplot/draw position={6-\s}] coordinates {};
    \addplot+[boxplot prepared={lower whisker=75,lower quartile=78.5,median=81.5,upper quartile=83.8,upper whisker=86,box extend=0.65}, draw=black,line width=0.8pt,fill=sslrl-color,fill opacity=0.7,solid,boxplot/draw position={6+\s}] coordinates {};
    \addplot+[boxplot prepared={lower whisker=71,lower quartile=80,median=83.5,upper quartile=84,upper whisker=88,box extend=0.65}, draw=black,line width=0.8pt,fill=ssl-color,fill opacity=0.7,solid,boxplot/draw position={9-\s}] coordinates {};
    \addplot+[boxplot prepared={lower whisker=77,lower quartile=82.5,median=87,upper quartile=89.8,upper whisker=93,box extend=0.65}, draw=black,line width=0.8pt,fill=sslrl-color,fill opacity=0.7,solid,boxplot/draw position={9+\s}] coordinates {};
    \addplot+[boxplot prepared={lower whisker=76,lower quartile=81.3,median=84,upper quartile=85.5,upper whisker=88,box extend=0.65}, draw=black,line width=0.8pt,fill=ssl-color,fill opacity=0.7,solid,boxplot/draw position={12-\s}] coordinates {};
    \addplot+[boxplot prepared={lower whisker=78,lower quartile=83.3,median=85,upper quartile=87.5,upper whisker=88,box extend=0.65}, draw=black,line width=0.8pt,fill=sslrl-color,fill opacity=0.7,solid,boxplot/draw position={12+\s}] coordinates {};
    \addplot+[boxplot prepared={lower whisker=74,lower quartile=78.25,median=79.5,upper quartile=83.5,upper whisker=88,box extend=0.65}, draw=black,line width=0.8pt,fill=ssl-color,fill opacity=0.7,solid,boxplot/draw position={15-\s}] coordinates {};
    \addplot+[boxplot prepared={lower whisker=78,lower quartile=81,median=82,upper quartile=86,upper whisker=87,box extend=0.65}, draw=black,line width=0.8pt,fill=sslrl-color,fill opacity=0.7,solid,boxplot/draw position={15+\s}] coordinates {};
    \addplot+[boxplot prepared={lower whisker=78,lower quartile=82.5,median=84,upper quartile=85.8,upper whisker=87,box extend=0.65}, draw=black,line width=0.8pt,fill=ssl-color,fill opacity=0.7,solid,boxplot/draw position={18-\s}] coordinates {};
    \addplot+[boxplot prepared={lower whisker=81,lower quartile=83,median=84.5,upper quartile=87,upper whisker=92,box extend=0.65}, draw=black,line width=0.8pt,fill=sslrl-color,fill opacity=0.7,solid,boxplot/draw position={18+\s}] coordinates {};
    \addplot+[boxplot prepared={lower whisker=79,lower quartile=82.3,median=85,upper quartile=88,upper whisker=89,box extend=0.65}, draw=black,line width=0.8pt,fill=ssl-color,fill opacity=0.7,solid,boxplot/draw position={21-\s}] coordinates {};
    \addplot+[boxplot prepared={lower whisker=82,lower quartile=83,median=85.5,upper quartile=88.8,upper whisker=90,box extend=0.65}, draw=black,line width=0.8pt,fill=sslrl-color,fill opacity=0.7,solid,boxplot/draw position={21+\s}] coordinates {};
    \addplot+[boxplot prepared={lower whisker=75,lower quartile=84,median=86,upper quartile=89.5,upper whisker=91,box extend=0.65}, draw=black,line width=0.8pt,fill=ssl-color,fill opacity=0.7,solid,boxplot/draw position={24-\s}] coordinates {};
    \addplot+[boxplot prepared={lower whisker=75.8,lower quartile=82,median=86,upper quartile=89.5,upper whisker=92,box extend=0.65}, draw=black,line width=0.8pt,fill=sslrl-color,fill opacity=0.7,solid,boxplot/draw position={24+\s}] coordinates {};
    \addplot+[boxplot prepared={lower whisker=80,lower quartile=83.3,median=85,upper quartile=88.5,upper whisker=93,box extend=0.65}, draw=black,line width=0.8pt,fill=ssl-color,fill opacity=0.7,solid,boxplot/draw position={27-\s}] coordinates {};
    \addplot+[boxplot prepared={lower whisker=78,lower quartile=86.25,median=87.5,upper quartile=88.8,upper whisker=92,box extend=0.65}, draw=black,line width=0.8pt,fill=sslrl-color,fill opacity=0.7,solid,boxplot/draw position={27+\s}] coordinates {};
    \addplot+[boxplot prepared={lower whisker=75,lower quartile=78,median=79,upper quartile=80,upper whisker=86,box extend=0.65}, draw=black,line width=0.8pt,fill=ssl-color,fill opacity=0.7,solid,boxplot/draw position={30-\s}] coordinates {};
    \addplot+[boxplot prepared={lower whisker=74,lower quartile=76,median=78,upper quartile=81.8,upper whisker=90,box extend=0.65}, draw=black,line width=0.8pt,fill=sslrl-color,fill opacity=0.7,solid,boxplot/draw position={30+\s}] coordinates {};
    \end{axis}
    \end{tikzpicture}
    \caption{\textbf{Effect of RL on expert legality.} Stockfish lv.0, pooled over 10 runs.}
    \label{fig:ssl_rl}
\end{figure}

\paragraph{How does RL affect playing strength?}

Table~\ref{tab:rl_results} reports the effect of RL on Draw Rate, Win Rate, and FIDEScore.
The picture that emerges is consistent across experts: RL universally increases the Draw Rate (by $+3$--$+8$ percentage points), while Win Rate decreases slightly in most configurations.
The increase in Draw Rate reflects a behavioral shift toward caution and a reduced tendency to make risky moves.

\begin{table*}[!ht]
    \caption{\textbf{Effect of RL on game results.} Stockfish lv.0, pooled over 10 runs. The top-5 experts balancing legality and FIDEScore are bolded.}
    \centering
    \begin{adjustbox}{width=\linewidth}
    \begin{threeparttable}
    \begin{tabular}{llcccccccccc}
    \hline
    & \textbf{Metric}\tnote{$\dagger$} & {\Large\ding{182}} & {\Large\ding{183}} & {\Large\ding{184}} & {\Large\ding{185}} & {\Large\ding{186}} & {\Large\ding{187}} & {\Large\ding{188}} & {\Large\ding{189}} & {\Large\ding{190}} & {\Large\ding{191}} \\
    \hline
    \multirow{3}{*}{\rotatebox{90}{SSL}}
    & Draw Rate
        & 14.7$_{\pm3.0}$ & 14.6$_{\pm3.7}$ & 15.2$_{\pm3.7}$ & 14.4$_{\pm3.1}$ & 15.8$_{\pm5.3}$
        & 16.2$_{\pm3.1}$ & 16.1$_{\pm2.6}$ & 13.5$_{\pm4.3}$ & 18.5$_{\pm4.9}$ & 16.3$_{\pm1.8}$ \\
    & Win Rate
        & 52.0$_{\pm4.5}$ & 52.6$_{\pm4.2}$ & 55.0$_{\pm5.4}$ & 55.3$_{\pm4.1}$ & 51.2$_{\pm6.5}$
        & 55.6$_{\pm5.2}$ & 55.4$_{\pm3.1}$ & 58.5$_{\pm5.5}$ & 56.4$_{\pm5.0}$ & 49.6$_{\pm4.1}$ \\
    & FIDEScore
        & 59.4$_{\pm3.8}$ & 59.9$_{\pm4.6}$ & 62.6$_{\pm5.3}$ & 62.5$_{\pm4.6}$ & 59.1$_{\pm5.5}$
        & 63.7$_{\pm4.4}$ & 63.5$_{\pm4.0}$ & 65.3$_{\pm4.1}$ & 65.6$_{\pm4.4}$ & 57.8$_{\pm4.1}$ \\
    \hdashline
    \multirow{3}{*}{\rotatebox{90}{\begin{tabular}{@{}c@{}}SSL\\+RL\end{tabular}}}
    & Draw Rate
        & \cellcolor{pos-delta-color!18}15.8$_{\pm4.4}$
        & \cellcolor{pos-delta-color!18}20.5$_{\pm2.5}$
        & \cellcolor{pos-delta-color!18}23.3$_{\pm4.7}$
        & \cellcolor{pos-delta-color!18}20.4$_{\pm3.8}$
        & \cellcolor{pos-delta-color!18}18.7$_{\pm3.0}$
        & \cellcolor{pos-delta-color!18}19.4$_{\pm4.1}$
        & \cellcolor{pos-delta-color!18}20.9$_{\pm3.9}$
        & \cellcolor{pos-delta-color!18}19.4$_{\pm5.8}$
        & \cellcolor{pos-delta-color!18}22.6$_{\pm5.7}$
        & \cellcolor{pos-delta-color!18}17.1$_{\pm3.7}$ \\
    & Win Rate
        & \cellcolor{neg-delta-color!18}51.4$_{\pm4.9}$
        & \cellcolor{neg-delta-color!18}47.3$_{\pm3.1}$
        & \cellcolor{neg-delta-color!18}51.1$_{\pm4.3}$
        & \cellcolor{neg-delta-color!18}48.5$_{\pm4.7}$
        & \cellcolor{pos-delta-color!18}51.6$_{\pm5.0}$
        & \cellcolor{neg-delta-color!18}54.1$_{\pm4.4}$
        & \cellcolor{neg-delta-color!18}53.8$_{\pm4.4}$
        & \cellcolor{neg-delta-color!18}53.2$_{\pm5.3}$
        & \cellcolor{neg-delta-color!18}52.7$_{\pm4.5}$
        & \cellcolor{pos-delta-color!18}49.8$_{\pm3.9}$ \\
    & FIDEScore
        & \cellcolor{neg-delta-color!18}59.3$_{\pm4.0}$
        & \cellcolor{neg-delta-color!18}57.6$_{\pm2.6}$
        & \cellcolor{pos-delta-color!18}\textbf{62.8$\boldsymbol{_{\pm3.8}}$}
        & \cellcolor{neg-delta-color!18}58.7$_{\pm3.5}$
        & \cellcolor{pos-delta-color!18}61.0$_{\pm4.2}$
        & \cellcolor{pos-delta-color!18}\textbf{63.8$\boldsymbol{_{\pm3.3}}$}
        & \cellcolor{pos-delta-color!18}\textbf{64.3$\boldsymbol{_{\pm3.1}}$}
        & \cellcolor{neg-delta-color!18}\textbf{62.9$\boldsymbol{_{\pm3.3}}$}
        & \cellcolor{neg-delta-color!18}\textbf{64.0$\boldsymbol{_{\pm3.4}}$}
        & \cellcolor{pos-delta-color!18}58.4$_{\pm4.9}$ \\
    \hline
    \end{tabular}
    \begin{tablenotes}
    \item[$\dagger$] Game metrics are Avg$\pm$Std (\%).
    Seed model: 24.0$\pm$2.4 (Draw Rate), 42.1$\pm$4.0 (Win Rate), 54.1$\pm$4.1 (FIDEScore).
    \end{tablenotes}
    \end{threeparttable}
    \end{adjustbox}
    \label{tab:rl_results}
\end{table*}

\paragraph{How does RL compare with constrained decoding?}

Table~\ref{tab:cd_fidescore} contextualizes RL against constrained decoding (CD), which enforces legality at inference time by restricting generation to $\mathcal{L}(s)$ at each step and represents the practical upper bound on legality-induced performance gains.
CD yields substantial playing strength improvements over unconstrained SSL across all experts ($+7.3$--$+11.2\%$ in FIDEScore), driven by a systematic increase in Win Rate alongside a reduction in Draw Rate.
The combination SSL+RL+CD attains the best FIDEScore in four out of five top experts and improves the \mom aggregate from $78.1$ (SSL+CD) to $78.7$, confirming that RL and CD are complementary rather than redundant: RL conditions the model toward legal play, while CD removes residual violations.

\begin{table}[!ht]
\centering
\caption{\textbf{Effect of constrained decoding (CD) on \mom FIDEScore.} FIDEScore for the top-5 experts, mean, and \mom. Stockfish lv.0. Best result per row in bold.}
\label{tab:cd_fidescore}
\renewcommand{\arraystretch}{1.25}
\setlength{\tabcolsep}{8pt}
\begin{adjustbox}{width=.54\linewidth}
\begin{tabular}{lcccc}
\hline
\rowcolor{colHd}
\textbf{Exp.} & \textbf{SSL} & \textbf{SSL+CD} & \textbf{SSL+RL} & \textbf{SSL+RL+CD} \\
\hline
\ding{184} & 62.6 & 76.4          & 62.8 & \textbf{77.1} \\
\ding{187} & 63.7 & 79.0          & 63.8 & \textbf{79.7} \\
\ding{188} & 63.5 & 77.0          & 64.3 & \textbf{77.8} \\
\ding{189} & 65.3 & \textbf{76.8} & 62.9 & 75.9          \\
\ding{190} & 65.6 & 76.4          & 64.0 & \textbf{76.8} \\
\hline
\rowcolor{gray!8}
Avg.  & 64.1 & 77.1 & 63.6 & \textbf{77.5} \\
\rowcolor{gray!8}
\mom  & 69.7 & 78.1 & 69.1 & \textbf{78.7} \\
\hline
\end{tabular}
\end{adjustbox}
\end{table}

\clearpage
\newpage

\section{On the estimation of playing strength}
\label{app:play_strength_estimation}

\subsection{Rating systems: Elo and Glicko-2}

The Elo rating system, introduced by Arpad Elo in 1967 and adopted by FIDE, models each player's strength as a single scalar and updates it after each game based on the expected outcome and the actual result, scaled by a $K$-factor that controls update magnitude.
BayesElo departs from this sequential update rule by treating all game results as a batch: it finds the set of ratings that best explains all results simultaneously.
Elo is simple to compute and interpret, but it makes no distinction between a rating derived from three games and one derived from three thousand: uncertainty in the estimate is not tracked.

Glicko-2 \cite{glickman2012example}, developed by Mark Glickman and adopted by Lichess, extends Elo with two additional quantities per player: a \emph{rating deviation} (RD) capturing uncertainty, and a \emph{volatility} parameter capturing how consistently a player performs.
This yields confidence intervals around every rating estimate and makes Glicko-2 statistically preferable to Elo when the number of games is small---a regime highly relevant for chess AI evaluation, where controlled experiments are expensive.

In practice, both systems tend to converge to similar absolute values when enough games are played, but Glicko-2 reaches accurate estimates faster and reports uncertainty bounds that Elo lacks entirely.

\subsection{Stockfish skill levels and Elo correspondence}

Stockfish exposes 21 skill levels (0--20) that do not limit search depth or time, but instead inject stochasticity into move selection by deliberately choosing among sub-optimal candidates with a probability that increases at lower levels.
This mechanism produces qualitatively different play than depth-limited engines: a Stockfish lv0 engine can still \quotes{see} deeply, but will frequently choose the wrong move from its candidate list.
Crucially, skill levels are \emph{not} linearly spaced on the Elo scale.
Figure~\ref{fig:stockfish_elo} reports the calibrated Elo estimates from the official Stockfish repository,\footnote{\url{https://bit.ly/sf-elo}} derived from self-play round-robin tournaments.
These values are hardware-, version-, time-control-, and hash-dependent, and should be treated as approximate anchors rather than fixed ground truth.
Levels 0 and 1, the most commonly used evaluation points in the chess AI literature, correspond to approximately $1{,}320$ and $1{,}468$ Elo respectively, placing them solidly in the intermediate amateur range.
Level 5 (${\approx}2{,}204$) already exceeds the performance of most club players, and level 20 reaches ${\geq}3{,}200$ Elo.

\begin{figure}[!htb]
\centering
\definecolor{chessbrown}{RGB}{118,75,22}
\definecolor{chessgreen}{RGB}{78,112,62}
\definecolor{chessfill}{RGB}{212,228,198}
\definecolor{fideline}{RGB}{98,52,12}
{\small
\textbf{CM} = Candidate Master\enspace\textbf{FM} = FIDE Master\enspace
\textbf{IM} = International Master\enspace\textbf{GM} = Grandmaster}\\[5pt]
\begin{tikzpicture}
\begin{axis}[
    width=0.85\linewidth,
    height=7.2cm,
    axis background/.style={fill=white},
    xmin=0, xmax=20,
    enlarge x limits=false,
    xtick={0,1,...,20},
    xticklabel style={font=\fontsize{7}{7}\selectfont},
    xlabel={Stockfish Skill Level},
    xlabel style={font=\fontsize{9}{9}\selectfont, yshift=-2pt},
    ymin=1200, ymax=3320,
    ytick={1200,1400,1600,1800,2000,2200,2400,2600,2800,3000,3200},
    yticklabel style={font=\fontsize{7}{7}\selectfont},
    ylabel={Approximate Elo},
    ylabel style={font=\fontsize{9}{9}\selectfont},
    ymajorgrids=false,
    xmajorgrids=false,
    clip=false,
    axis on top=false,
    tick align=outside,
    axis line style={black!35, thin},
]
%
\addplot[chessfill!60, fill=chessfill!50, line width=0pt, forget plot]
    coordinates{
        (0,1320)(1,1468)(2,1608)(3,1742)(4,1923)(5,2204)
        (6,2363)(7,2500)(8,2596)(9,2703)(10,2788)(11,2856)
        (12,2923)(13,2973)(14,3025)(15,3070)(16,3111)(17,3141)
        (18,3170)(19,3191)(20,3200)(20,1200)(0,1200)
    } \closedcycle;
%
\foreach \yval in {1200,1400,1600,1800,2000,2200,2400,2600,2800,3000,3200}{
    \addplot[gray!30, thin, solid, forget plot]
        coordinates{(0,\yval)(20,\yval)};
}
\foreach \xval in {0,2,4,6,8,10,12,14,16,18,20}{
    \addplot[gray!30, thin, solid, forget plot]
        coordinates{(\xval,1200)(\xval,3320)};
}
%
\addplot[fideline, solid, line width=1.1pt, forget plot]
    coordinates{(0,2200)(20,2200)};
\addplot[fideline, solid, line width=1.1pt, forget plot]
    coordinates{(0,2300)(20,2300)};
\addplot[fideline, solid, line width=1.1pt, forget plot]
    coordinates{(0,2400)(20,2400)};
\addplot[fideline, solid, line width=1.1pt, forget plot]
    coordinates{(0,2500)(20,2500)};
%
\addplot[fideline!60, dashed, line width=0.8pt, forget plot]
    coordinates{(0,1380)(20,1380)};
\addplot[fideline!60, dashed, line width=0.8pt, forget plot]
    coordinates{(0,1790)(20,1790)};
\addplot[fideline!60, dashed, line width=0.8pt, forget plot]
    coordinates{(0,2700)(20,2700)};
%
\node[font=\fontsize{8.5}{8.5}\selectfont\itshape, text=black!40,
      anchor=west] at (axis cs:20.3,1380) {Intermediate};
\node[font=\fontsize{8.5}{8.5}\selectfont\itshape, text=black!40,
      anchor=west] at (axis cs:20.3,1790) {Club};
\node[font=\fontsize{8.5}{8.5}\selectfont\bfseries, text=fideline,
      anchor=west] at (axis cs:20.3,2200)
    {\textbf{CM}};
\node[font=\fontsize{8.5}{8.5}\selectfont\bfseries, text=fideline,
      anchor=west] at (axis cs:20.3,2300)
    {\textbf{FM}};
\node[font=\fontsize{8.5}{8.5}\selectfont\bfseries, text=fideline,
      anchor=west] at (axis cs:20.3,2400)
    {\textbf{IM}};
\node[font=\fontsize{8.5}{8.5}\selectfont\bfseries, text=fideline,
      anchor=west] at (axis cs:20.3,2500)
    {\textbf{GM}};
\node[font=\fontsize{8.5}{8.5}\selectfont, text=black!40,
      anchor=west] at (axis cs:20.3,2700)
    {\textit{Super-GM}};
%
\addplot[
    color=chessbrown,
    line width=2.0pt,
    mark=none,
] coordinates {
    (0,1320)(1,1468)(2,1608)(3,1742)(4,1923)(5,2204)
    (6,2363)(7,2500)(8,2596)(9,2703)(10,2788)(11,2856)
    (12,2923)(13,2973)(14,3025)(15,3070)(16,3111)(17,3141)
    (18,3170)(19,3191)(20,3200)
};
%
\addplot[
    only marks, mark=*, mark size=2.3pt,
    mark options={fill=chessgreen, draw=chessgreen!70!black, line width=0.5pt},
] coordinates {
    (0,1320)(1,1468)(2,1608)(3,1742)(4,1923)(5,2204)
    (6,2363)(7,2500)(8,2596)(9,2703)(10,2788)(11,2856)
    (12,2923)(13,2973)(14,3025)(15,3070)(16,3111)(17,3141)
    (18,3170)(19,3191)(20,3200)
};
\end{axis}
\end{tikzpicture}
\vspace{3pt}
{\footnotesize\textit{Super-GM} is an informal community term, not an official FIDE title. \textit{Intermediate} and \textit{Club} are likewise informal tiers.}
\caption{\textbf{Approximate Elo ratings for Stockfish skill levels 0--20.} Horizontal reference lines mark the four official FIDE title thresholds.}
\label{fig:stockfish_elo}
\end{figure}

\subsection{Evaluation practices in prior work}

Table~\ref{tab:elo_survey} summarizes the strength estimation practices of the chess language model papers surveyed in \cref{app:related_work}.
Only 5 out of 13 papers perform any form of rating estimation; the remaining 8 focus exclusively on puzzle solving, state tracking, move generation quality, or auxiliary generative tasks and report no game-playing evaluation.
Among the 5 that do, no two share the same combination of rating system, opponent type, and game count---a fragmentation that renders cross-paper comparisons unreliable without a shared controlled protocol.

\paragraph{Opponents}

Two methodologically distinct evaluation families emerge from the surveyed literature, differing fundamentally in the nature of the opponents the model is pitted against.
The \emph{first family} evaluates the model against real human players on a live platform, deriving a strength estimate from accumulated rated game results.
From a methodological standpoint, this is the gold standard: it anchors the model's strength to the same scale used for human players.
\cite{DBLP:conf/nips/RuossDMGLCRLVG24} are the only in our survey to adopt this protocol, accumulating ${\sim}150$ rated blitz games on Lichess.
However, human evaluation is also expensive, slow, and prone to selection bias---the population of players willing to challenge an unknown bot on Lichess may not represent the broader player distribution.
Furthermore, accumulating a statistically meaningful number of games requires prolonged deployment, making this approach inaccessible to most academic groups.
The \emph{second family}---by far the more common---replaces human opponents with automated ones, either Stockfish at specified skill levels, bots, or a pool of model variants in a round-robin tournament.
\cite{DBLP:journals/corr/abs-2403-15498} and \cite{DBLP:conf/naacl/ZhangHLCL25} evaluate directly against Stockfish at fixed levels (lv.0--5 and lv.0--2 respectively), anchoring win/draw/loss outcomes to the approximate, official level-to-Elo mapping.
\cite{DBLP:conf/nips/ZhangZSKETKM24} also derive ratings by playing against Stockfish (lv.1,3,5); prior to model evaluation, Maia bots (trained at rating bins 1100--1200, 1500--1600, and 1900--2000) are used to calibrate the ratings of the three Stockfish levels on Lichess's platform.
\cite{DBLP:conf/nips/RuossDMGLCRLVG24} make the automatic tournament evaluation with a large pool of baselines, comprising own model variants, AlphaZero in three configurations, Leela Chess Zero in three analogous configurations, and Stockfish at two time controls.
\cite{DBLP:journals/corr/abs-2409-12272} play $1{,}000$ games against the GC-270M agent of \cite{DBLP:conf/nips/RuossDMGLCRLVG24}.
Stockfish (with levels up to 5) is the dominant opponent choice among scrutinized papers.

\paragraph{Rating systems}

Elo, Glicko-2, and BayesElo each account for roughly one third of the rating-estimation papers, reflecting the absence of a community consensus on a preferred protocol.

\paragraph{Reproducibility and comparability}

The evaluations surveyed are, in practice, mutually incomparable.
No two papers use the same tournament setting: they differ in rating systems, reference populations, and game counts.
The resulting rating scores cannot be fairly compared across papers, nor with ratings reported on online platforms such as Lichess or Chess.com.
Furthermore, key experimental quantities are frequently omitted or underspecified---such as opponent hyperparameters, hardware platform, precise Elo anchoring formula, and tournament settings, making independent reproduction effectively impossible even when the nominal protocol appears identical.

\begin{table*}[!ht]
\caption{\textbf{Ranking practices across chess language models reviewed in \cref{app:related_work}.}}
\label{tab:elo_survey}
\centering
\setlength{\tabcolsep}{4pt}
\renewcommand{\arraystretch}{1.3}
\begin{adjustbox}{width=\linewidth}
\begin{threeparttable}
\begin{tabular}{llccp{3.2cm}cc}
\hline
\textbf{Paper} & \textbf{Venue} &
\textbf{Rating est.} &
\textbf{Rating system} &
\textbf{Opponent type} &
\textbf{\# Distinct opponents}\tnote{$\star$} &
\textbf{Games / opponent} \\
\hline
\cite{DBLP:journals/corr/abs-2008-04057} Noever et al.
    & ArXiv 2020
    & \xmark & --- & --- & --- & --- \\
\rowcolor{gray!6}
\cite{DBLP:conf/aaai/ToshniwalWLG22} Toshniwal et al.
    & AAAI 2022
    & \xmark & --- & --- & --- & --- \\
\cite{DBLP:conf/nips/FengLWTYSM0W23} Feng et al.
    & NeurIPS 2023
    & \xmark & --- & --- & --- & --- \\
\rowcolor{gray!6}
\cite{carlini2023playing} Carlini et al.
    & ArXiv 2023
    & \xmark & --- & --- & --- & --- \\
\cite{DBLP:journals/corr/abs-2403-15498} Karvonen
    & COLM 2024
    & \cmark
    & Elo
    & Stockfish {[lv.0--5]}
    & 6
    & 1,000 \\
\rowcolor{gray!6}
\cite{DBLP:conf/nips/ZhangZSKETKM24} Zhang et al.
    & NeurIPS 2024
    & \cmark
    & Glicko-2
    & Stockfish {[lv.1,3,5]}
    & 3
    & 100 \\
\cite{DBLP:conf/nips/RuossDMGLCRLVG24} Ruoss et al.
    & NeurIPS 2024
    & \cmark
    & Glicko-2
    & Humans
    & N/A
    & N/A ($\sim$150 total) \\
\rowcolor{gray!6}
\cite{DBLP:conf/nips/RuossDMGLCRLVG24} Ruoss et al.
    & NeurIPS 2024
    & \cmark
    & BayesElo
    & Variants of Stockfish {[N/A]}, AlphaZero, LC0
    & 8
    & 400 \\
\cite{DBLP:journals/corr/abs-2409-12272} Monroe \& LC0 Team
    & ArXiv 2024
    & \cmark
    & BayesElo
    & GC-270M \cite{DBLP:conf/nips/RuossDMGLCRLVG24}
    & 1
    & 1,000 \\
\rowcolor{gray!6}
\cite{DBLP:conf/naacl/WangJWZLHW25} Wang et al.
    & NAACL 2025
    & \xmark & --- & --- & --- & --- \\
\cite{DBLP:conf/naacl/ZhangHLCL25} Zhang et al.
    & NAACL 2025
    & \cmark
    & Elo
    & Stockfish {[lv.0--2]}
    & 3
    & 100 \\
\rowcolor{gray!6}
\cite{DBLP:journals/corr/abs-2507-12215} Su et al.
    & ArXiv 2025
    & \xmark & --- & --- & --- & --- \\
\cite{DBLP:journals/corr/abs-2507-00726} Hwang et al.
    & ArXiv 2025
    & \xmark & --- & --- & --- & --- \\
\hline
\end{tabular}
\begin{tablenotes}
\item[$\star$] Counts exclude any variants of the authors' own models that participate as agents in the same tournament pool.
\end{tablenotes}
\end{threeparttable}
\end{adjustbox}
\end{table*}

\clearpage
\newpage

\section{Merging techniques}
\label{app:merging}

A potential consequence of weight interpolation in \mom stitching is the catastrophic forgetting of acquired chess capabilities. To determine the optimal merging configuration for downstream performance, we systematically evaluate diverse parameter consolidation techniques spanning weight-based, gradient-informed, and subspace-oriented approaches over the fully merged model. Concretely, we merge the five \mom masters---\ding{184}~M. Carlsen, \ding{187}~A. Giri, \ding{188}~H. Nakamura, \ding{189}~I. Nepomniachtchi, and \ding{190}~W. So---and evaluate the resulting merged model across 10 experimental runs of 300 games against Stockfish level 0.

Weight-based methods form the foundation of our analysis, with naive averaging~\cite{DBLP:conf/icml/WortsmanIGRLMNF22} serving as the baseline approach, which assigns uniform importance to all parameters. Task arithmetic~\cite{DBLP:conf/iclr/IlharcoRWSHF23} provides a more principled alternative by leveraging task-specific weight differences relative to the base model, thereby preserving specialized capabilities during integration. To capture higher-order parameter relationships, we evaluate KnOTS~\cite{DBLP:conf/iclr/StoicaRECH25}, which employs Singular Value Decomposition to identify and merge critical parameter subspaces that simpler averaging methods might compromise.

Beyond weight-centric methodologies, we investigate gradient-based approaches utilizing Fisher information matrices~\cite{DBLP:conf/nips/MatenaR22,DBLP:conf/naacl/LeeLWWCW25}. These methods approximate the Hessian of the loss function to weight parameters according to their empirical importance in the optimization landscape, with parameters exhibiting higher Fisher information values receiving proportionally greater influence during consolidation. This information-theoretic paradigm fundamentally differs from uniform weighting by prioritizing parameters that contribute most significantly to model performance.

As shown in Figure~\ref{fig:mergingTec}, Fisher-based merging achieves the highest win rate (56.2\%) followed closely by naive averaging (56.8\%) and KnOTS (54.6\%), all of which surpass the average player baseline (54.16\%). Task arithmetic performs considerably worse (42.0\%), falling well below the baseline. Wilcoxon signed-rank tests confirm that these performance differences are statistically significant across our experimental runs.

However, raw win rate alone does not capture the full picture. A critical analysis of the underlying parameter distributions reveals a fundamental imbalance in several merging strategies when applied to our five masters. As reported in Table~\ref{tab:merging_weights}, naive averaging is the only method that preserves a perfectly uniform contribution of 20\% per expert. In contrast, KnOTS exhibits severe imbalance, with \ding{190}~W. So dominating at 50.00\% and the remaining masters each contributing only 6.25--12.50\%. Task arithmetic collapses almost entirely onto the base model (98.00\%), effectively discarding all expert knowledge. Fisher weighting, despite its strong win rate, introduces a moderate but measurable bias toward \ding{188}~H. Nakamura (23.15\%) at the expense of the other masters.

This imbalance directly undermines the democratic principle at the core of \mom architectures, wherein each expert should contribute equally to the merged model. 
Notably, the Fisher method---aptly named for our chess domain---achieves competitive performance, yet its inherent bias toward \ding{188}~H. Nakamura contradicts the democratic principle at the core of \mom architectures. For this reason, we adopt naive averaging as our merging method of choice for the primary analysis, as it is the only technique that guarantees equal expert contribution while still achieving competitive performance above the average player baseline.

\begin{figure}[!h]
    \centering
    \begin{subfigure}[t]{.26\linewidth}
    \begin{tikzpicture}
        \begin{axis}[
            ybar,
            bar width=6pt,
            width=\linewidth,
            height=5cm,
            enlarge x limits=0.22,
            ymajorgrids=true,
            grid style=dashed,
            axis background/.style={fill=plotbackground},
            ylabel={Win Rate},
            ymin=0,
            ymax=70,
            xlabel={Merging Method},
            symbolic x coords={fisher, naive, knots, task},
            xtick=data,
            xticklabels={{Fisher Weighted}, {Naive Average}, {KnOTS}, {Task Arithmetic}},
            xticklabel style={rotate=90, align=right, text width=2.4cm, font=\fontsize{8}{8}\selectfont},
            nodes near coords,
            nodes near coords align={vertical},
            every node near coord/.append style={font=\bfseries\fontsize{6}{6}\selectfont, text=black, yshift=-2pt},
            every tick label/.append style={font=\fontsize{8}{8}\selectfont},
            xlabel style={font=\fontsize{8}{8}\selectfont},
            ylabel style={font=\fontsize{8}{8}\selectfont},
        ]
    
        \addplot+ [fill=bar-color, draw=none] coordinates {
            (fisher, 56.2) (naive, 56.8) (knots, 54.6)
            (task, 42.0)
        };

        \draw [delta-color, thick, dashed] (rel axis cs:0,0.7737) -- (rel axis cs:1,0.7737)
            node [anchor=south, xshift=29pt, yshift=-2pt, font=\fontsize{6}{6}\selectfont, text=delta-color] {Avg player (54.16)};
        
        \end{axis}
    \end{tikzpicture}
    \caption{Win rate comparison.}
    \end{subfigure}
    \hfill
    \begin{subfigure}[t]{.26\linewidth}
    \begin{tikzpicture}
        \begin{axis}[
            ybar,
            bar width=6pt,
            width=\linewidth,
            height=5cm,
            enlarge x limits=0.22,
            ymajorgrids=true,
            grid style=dashed,
            axis background/.style={fill=plotbackground},
            ylabel={Improvement over Avg Player},
            ymin=-15,
            ymax=10,
            xlabel={Merging Method},
            symbolic x coords={fisher, naive, knots, task},
            xtick=data,
            xticklabels={{
                Fisher Weighted}, 
                {Naive Average}, 
                {KnOTS}, 
                {Task Arithmetic}
            },
            xticklabel style={rotate=90, align=right, text width=2.4cm, font=\fontsize{8}{8}\selectfont},
            nodes near coords,
            nodes near coords align={vertical},
            every node near coord/.append style={
                font=\bfseries\fontsize{6}{6}\selectfont,
                text=black,
                yshift=2pt
            },
            every tick label/.append style={font=\fontsize{8}{8}\selectfont},
            xlabel style={font=\fontsize{8}{8}\selectfont},
            ylabel style={font=\fontsize{8}{8}\selectfont},
        ]
    
        \addplot+ [
            ybar,
            fill=delta-color,
            draw=none
        ] coordinates {
            (fisher, 2.04) 
            (naive, 2.64) 
            (knots, 0.44)
            (task, -12.16)
        };
        
        \end{axis}
    \end{tikzpicture}
    \caption{Win rate improvement of the merged models over the avg player performance.}
    \end{subfigure}
    \hfill
    \begin{subfigure}[t]{.42\linewidth}
    \begin{tikzpicture}
        \begin{axis}[
            ybar,
            bar width=6pt,
            width=\linewidth,
            height=5cm,
            enlarge x limits=0.12,
            ymajorgrids=true,
            grid style=dashed,
            axis background/.style={fill=plotbackground},
            ymode=log,
            log basis y={10},
            ylabel={P-value},
            ymin=0.0005,
            ymax=0.2,
            xlabel={Merging Method},
            symbolic x coords={1, 2, 3, 4, 5, 6},
            xtick=data,
            xticklabels={{Naive Average vs.\\Fisher Weighted}, {Naive Average vs.\\Task Arithmetic}, {Naive Average vs.\\KnOTS}, {Task Arithmetic vs.\\Fisher Weighted}, {KnOTS vs.\\Fisher Weighted}, {KnOTS vs.\\Task Arithmetic}},
            xticklabel style={rotate=90, align=right, text width=2.4cm, font=\fontsize{8}{8}\selectfont},
            nodes near coords={**},
            nodes near coords align={vertical},
            every node near coord/.append style={font=\bfseries\fontsize{6}{6}\selectfont, text=black, yshift=2pt},
            every tick label/.append style={font=\fontsize{8}{8}\selectfont},
            xlabel style={font=\fontsize{8}{8}\selectfont},
            ylabel style={font=\fontsize{8}{8}\selectfont},
        ]
    
        \addplot+ [fill=bar-color, draw=none] coordinates {
            (1, 0.002) (2, 0.002) (3, 0.002)
            (4, 0.002) (5, 0.002) (6, 0.002)
        };
        
        \end{axis}
    \end{tikzpicture}
    \caption{Statistical significance of pairwise comparisons (Wilcoxon signed-rank test).}
    \end{subfigure}
    \caption{\textbf{Win Rate comparison between merging algorithms.}}
    \label{fig:mergingTec}
\end{figure}

\begin{table}[!ht]
\centering
\caption{\textbf{Parameter contribution per expert across merging methods.} Percentage of each master's contribution in the merged model for each merging strategy.}
\label{tab:merging_weights}
\renewcommand{\arraystretch}{1.25}
\setlength{\tabcolsep}{8pt}
\begin{adjustbox}{width=.85\linewidth}
\begin{tabular}{lcccccc}
\hline
\rowcolor{colHd}
\textbf{Merge Method} & \textbf{\ding{184} Carlsen} & \textbf{\ding{187} Giri} & \textbf{\ding{188} Nakamura} & \textbf{\ding{189} Nepomniachtchi} & \textbf{\ding{190} So} & \textbf{Karv.} \\
\hline
Fisher          & 18.57\% & 20.88\% & 23.15\% & 18.83\% & 18.57\% & 0.00\% \\
KnOTS           &  6.25\% &  6.25\% & 12.50\% & 25.00\% & 50.00\% & 0.00\% \\
Naive Average   & 20.00\% & 20.00\% & 20.00\% & 20.00\% & 20.00\% & 0.00\% \\
Task Arithmetic &  0.40\% &  0.40\% &  0.40\% &  0.40\% &  0.40\% & 98.00\% \\
\hline
\end{tabular}
\end{adjustbox}
\end{table}

\clearpage
\newpage

\section{Implementation details and hardware setup}
\label{app:implementation_hw}

\paragraph{Tokenizer}

To ensure a lightweight and parameter-efficient model architecture, we selected a minimal 32-character vocabulary (Table~\ref{tab:vocabulary}), the most compact set necessary for representing PGN sequences.
This decision is directly informed by \cite{DBLP:journals/corr/abs-2403-15498}, who demonstrated that employing the GPT-3.5's default BPE tokenizer with 50,257 entries would inflate the model's parameter count by 25M.
Furthermore, Karvonen's analysis revealed that the larger tokenizer provides no commensurate improvement in encoding efficiency for this domain, as it already encodes PGN strings with slightly over 1 character per token (excluding spaces).
Accordingly, all our experiments were conducted using seed models that rely exclusively on this tokenization scheme.
In line with \cite{DBLP:journals/corr/abs-2403-15498} and \cite{DBLP:conf/nips/ZhangZSKETKM24}, we ensured that--during training--every batch began with the sequence \quotes{;1.} to serve as a delimiter for a new game.
Note that the token \texttt{e3} shown at the output of
Figure~\ref{fig:mom} is a simplification for readability; in
practice, each character is tokenized individually.

\paragraph{\mom parameters}

A single \gm model has 50,905,088 parameters.
\mom (5 experts) has 185,245,696 parameters.

\paragraph{Random seed}

Reproducibility is ensured by fixing the random seed to 960--which we call the \quotes{\textit{Fischer seed}}.

\begin{table}[!htb]
    \centering
    \caption{\textbf{\mom vocabulary.} A 32-token character set containing all necessary letters, digits, and symbols to describe move sequences within PGN games.}
    \label{tab:vocabulary}
    \resizebox{\textwidth}{!}{
    \begin{tabular}{llp{5cm}p{9cm}}
    \toprule
    \textbf{Category} & \textbf{\# Tokens} & \textbf{Tokens} & \textbf{Comment} \\
    \midrule
    Files & 8 & \tokentag{a} \tokentag{b} \tokentag{c} \tokentag{d} \tokentag{e} \tokentag{f} \tokentag{g} \tokentag{h} & Board columns \\[2mm]
    Digits & 10 & \tokentag{0} \tokentag{1} \tokentag{2} \tokentag{3} \tokentag{4} \tokentag{5} \tokentag{6} \tokentag{7} \tokentag{8} \tokentag{9} & Board rows (ranks) and move numbers \\[2mm]
    Pieces & 5 & \tokentag{K} \tokentag{Q} \tokentag{R} \tokentag{B} \tokentag{N} & King, queen, rook, bishop, knight -- pawns have no letter in SAN\\[2mm]
    Castling & 1 & \tokentag{O} & King-side (O-O) and queen-side (O-O-O) castling \\[2mm]
    Move & 5 & \tokentag{x} \tokentag{+} \tokentag{\#} \tokentag{=} \tokentag{-} & Capture, check, checkmate, promotion, and castling dash \\[2mm]
    Separators & 3 & \tokentag{.} \tokentag{ } \tokentag{;} & Period after move numbers, space separator between moves, special delimiter denoting game start/end \\
    \bottomrule
    \end{tabular}
    }
\end{table}

\paragraph{Game validation}
Automatic legality checks on PGN strings for dataset construction and legality metric implementation are performed using the \texttt{python-chess} (v1.11.2) library.

\paragraph{Hyperparameters}
All models were trained using hyperparameters optimized through Gaussian process-based Bayesian optimization for the most critical parameters, with remaining settings determined via standard search methods. The optimization ranges and final selected configurations are presented in Tables~\ref{tab:hyperparameters_sslrl} and Table~\ref{tab:hyperparameters}.

\begin{table}[!htb]
\centering
\caption{\textbf{SSL and RL hyperparameter sweep.} Ranges come from the optimization config; starred values correspond to the selected training setting.}
\label{tab:hyperparameters_sslrl}
\fontsize{9}{11}\selectfont
\begin{threeparttable}
\begin{tabular}{@{}l@{}}
\toprule
\textbf{SSL/RL Hyperparameter Setting} \\
\midrule
\textbf{SSL Training configuration:} \\
$\text{lr}=[1e-7\ldots2e-6^\ast\ldots1e-4], \; \text{weight decay}=[1e-5\ldots1e{-4}^\ast\ldots1e-1]$ \\
$\text{dropout}=0.0, \text{batch size}=8$ \\
$\text{warmup steps}=600, \text{training steps}=6,000$ \\
\textbf{RL Training configuration:} \\
$\text{lr}=[1e-7\ldots6e-7^\ast\ldots1e-4], \; \beta=[0.01\ldots0.06^\ast\ldots0.1]$ \\
$\text{group size}=8,\; \text{batch size}=64$ \\
$\text{warmup steps}=10,00$,\; \text{training steps}=10,000 \\
\bottomrule
\end{tabular}
\end{threeparttable}
\end{table}

\begin{table}[!htb]
\centering
\caption{\textbf{Vision encoder stylometry hyperparameter sweep.} Ranges come from the optimization config; starred values correspond to the selected training setting.}
\label{tab:hyperparameters}
\fontsize{9}{11}\selectfont
\begin{threeparttable}
\begin{tabular}{@{}l@{}}
\toprule
\textbf{Stylometry Hyperparameter Setting} \\
\midrule
\textcolor{brown}{$\mathbf{\S}$} \textbf{\textcolor{brown}{Pretraining hyper-parameters}}\\
\textbf{Training configuration:} \\
$\text{lr}\,E_\psi=[1e-7 \ldots 1e{-6}^\ast \ldots 1e-3],\; \text{lr classifier}=[1e-5,\ldots1e{-4}^\ast\ldots1e-2]$\\
$\text{weight decay}=[1e-5\ldots 1e{-4}^\ast\ldots 1e-1],\; \text{dropout}=[0.15\ldots{0.3}^\ast\ldots 0.4]$\\
$\text{batch size}=92,\;\; \text{pos weight}=30$ \\
$\text{epochs}=40,\; \text{warmup steps}=1500$,\; \text{training steps}=15000\\ 
$\text{classifier hidden dim}=256$\\
\hdashline
\textcolor{brown}{$\mathbf{\S}$} \textbf{\textcolor{brown}{Finetuning hyper-parameters}}\\
\textbf{GE2E Loss parameters:} \\
$\lambda_{\text{m}}=[0.05 \ldots 0.8^\ast \ldots 1.0]$ \\
$\lambda_{\text{c}}=[0.001 \ldots 0.7^\ast \ldots 1.0]$ \\
$\mu=[0.1\ldots0.5^\ast\ldots1]$\\
\addlinespace[0.5ex]
\textbf{GE2E parameters:} \\
$W=[1.0 \ldots 8.5^\ast \ldots 15.0]$ \\
$b=[-12.0 \ldots -10^\ast \ldots 2.0]$ \\
\addlinespace[0.5ex]
\textbf{Training configuration:} \\
$\text{lr}=[1e-7 \ldots 5e{-6}^\ast\ldots1e-3], \; \text{weight decay}=[1e-6\ldots2e{-3}^\ast\ldots1e-1]$ \\
$\text{dropout}=[0.05\ldots0.15^\ast\ldots0.4]$ \\
$\text{batch size}=4,\; N=5,\; M=10,\; F=5$ \\
$\text{epochs}=20,\; \text{warmup steps}=2500$,\; \text{training steps}=25000 \\
$ \text{LSTM hidden dim}=512$ \\
\bottomrule
\end{tabular}
\end{threeparttable}
\end{table}

\paragraph{Compute resources}
All experiments were performed on a workstation running Ubuntu 20.04.3 LTS, equipped with an Intel® Core™ i9-10900X CPU @ 3.70GHz and 128GB of RAM. The optimization of behavioral stylometry models was conducted on two NVIDIA GeForce RTX3090 GPUs (24GB VRAM), while all remaining computations were executed on a NVIDIA GeForce RTX5090 (32GB VRAM).
On an RTX 5090 32GB, SSL takes about 25 minutes for 5k steps, consistent with previous work on the same 50 million parameters model. During the GPRO phase, training lasts approximately 45 minutes for 5k steps.

\clearpage
\newpage

\section{Impact Statement}
\label{app:impact-statement}

This work introduces \mom, a sparse chess language model that composes independently trained \gm experts through lightweight routing.
Its primary contribution is methodological: it shows that persona-specialized branches can be combined into a stronger and more controllable autoregressive model, while preserving behavioral diversity that may be suppressed by dense training on aggregated, player-undistinguished data.
Although chess is a bounded and highly structured domain, it provides a useful testbed for studying broader questions in generative modeling, including specialization, composition, interpretability, and the tension between performance and diversity.

The broader relevance of \mom lies in its evidence that sparse composition can exploit coherent behavioral variation across experts rather than collapsing it into a single averaged model.
This has potential implications for MoE systems beyond chess, where expert specialization is often discussed primarily in terms of scale or computational efficiency.
Our results suggest that specialization may also serve as a mechanism for preserving complementary behaviors that improve downstream decision-making.
In this respect, the paper contributes to a growing line of work on modular and compositional AI systems, where smaller specialized components are combined into more capable general systems.

We believe \mom holds practical potential for chess education: its modular architecture naturally supports configurable opponents that emulate distinct \gm styles, offering learners a structured way to study and train against diverse strategic profiles within a single system.

\mom should not be understood as a replacement for solver methods, search-based engines, or top-level automatic chess systems.
The model is designed to study chess as a language modeling problem, where moves are generated autoregressively from PGN sequences without explicit tree search.
Its impact is therefore best framed not as advancing the frontier of engine strength, but as showing how sparse language models can combine specialized behaviors in a controlled sequential decision-making environment.
We hope this perspective encourages further work on compositional models that are not only stronger, but also more inspectable, controllable, and behaviorally diverse.


\clearpage
\newpage

\section{Limitations}
\label{app:limitations}

The empirical scope of this work is intentionally centered on chess as a language-modeling problem.
\mom predicts the next move autoregressively from PGN sequences and is not designed to compete directly with solver methods, search-based engines, or automatic chess systems with top-level playing strength.
Systems based on explicit tree search, value estimation, engine-guided policy optimization, tablebases, or other specialized planning mechanisms operate under substantially different assumptions from the transformer-based next-move prediction setting considered here.
The controlled comparison between dense experts, merged models, random-partition controls, and \mom is therefore meant to isolate the effect of persona-specialized sparse modeling within a shared language-modeling framework, rather than to position \mom as an alternative to state-of-the-art chess engines.

The playing-strength evaluation follows a fixed Stockfish protocol with bounded search, greedy decoding, a no-retry illegality policy, and a finite game horizon.
These choices make the benchmark reproducible and computationally feasible, but they capture only one operational view of chess competence.
Accordingly, the reported ratings and scores should be interpreted comparatively within this tournament setup, rather than as absolute estimates transferable to online platforms, human play, or other evaluation protocols.

The player-specialization analysis is bounded by the selected grandmaster cohort and by the available historical data for each player.
The chosen experts provide a tractable and interpretable testbed, but they cover only a small and highly elite region of the chess population.
It remains open whether the same specialization and routing sharpness would emerge for larger cohorts, lower-rated players, players with fewer games, or experts defined by openings, eras, time controls, or strategic archetypes.
Dataset size, opponent distribution, color balance, and historical period may also influence both expert strength and apparent stylistic separability.

The stylometric evidence should be interpreted as behavioral and statistical rather than causal.
Own-master likelihood advantages, activation displacement, and concentrated routing patterns indicate that player-specific regularities are present in the trained models, but they do not prove that the models recover a complete or uniquely identifiable notion of human style.
Some signal may arise from opening repertoires, repeated structures in historical games, opponent pools, or other contextual regularities correlated with player identity.
Similarly, router activations provide a useful computational trace, but
they should not be read as literal explanations of human decision-making.

Finally, the reinforcement-learning experiments are deliberately narrow.
GRPO is used as a single-move post-training procedure because the experimental setting is next-move prediction from a given board state, rather than long-form trajectory generation.
This is aligned with the game setting considered in the paper, but it does not exploit the regime in which GRPO is typically most effective: conditioning the on-policy behavior of a model over long completions, where rewards can shape extended generation dynamics.
The GRPO results should therefore be understood as targeted move-level refinement rather
than a full exploration of long-horizon reinforcement learning for chess language models.

\clearpage
\newpage

\section{Ethical Concerns}
\label{app:ethical-concerns}

The main ethical concern is related to the potential misuse of stylometric components for player profiling or behavioral attribution.
To mitigate this risk, the weights of the ablative stylometry model are not provided, preventing the most directly reusable profiling component from being applied outside the intended scientific analysis.
Apart from this point, we do not identify additional ethical concerns specific to the proposed chess modeling framework.

\end{appendices}

\clearpage
\newpage



\newpage
\section*{NeurIPS Paper Checklist}

\begin{enumerate}

\item {\bf Claims}
    \item[] Question: Do the main claims made in the abstract and introduction accurately reflect the paper's contributions and scope?
    \item[] Answer: \answerYes{}
    \item[] Justification: All quantitative and qualitative claims mentioned in the abstract and \cref{sec:introduction} are substantiated by the results presented in \cref{subsec:results}. Additional experimental evidence supporting these claims is provided in ~\cref{app:behavioral_stylometry}, ~\cref{app:rl}, and ~\cref{app:merging}.

\item {\bf Limitations}
    \item[] Question: Does the paper discuss the limitations of the work performed by the authors?
    \item[] Answer: \answerYes{}
    \item[] Justification: Limitations and ethical considerations are comprehensively discussed in ~\cref{app:limitations} and ~\cref{app:ethical-concerns}, respectively.

\item {\bf Theory assumptions and proofs}
    \item[] Question: For each theoretical result, does the paper provide the full set of assumptions and a complete (and correct) proof?
    \item[] Answer: \answerNA{}.
    \item[] Justification: Our equations build on well-established theoretical results and are used primarily to support the practical contributions of our work, which lies in chess MoE and expert specialization through behavioral cloning of grandmaster play. The mathematical formulations presented involve verifiable manipulations that illustrate the underlying mechanics of our approach, rather than requiring formal proofs or new assumptions.

\item {\bf Experimental result reproducibility}
    \item[] Question: Does the paper fully disclose all the information needed to reproduce the main experimental results of the paper to the extent that it affects the main claims and/or conclusions of the paper (regardless of whether the code and data are provided or not)?
    \item[] Answer: \answerYes{}
    \item[] Justification: We provide detailed descriptions of the experimental setup in \cref{subsec:experimental_setup} and ~\cref{app:implementation_hw} to ensure that our results can be reliably reproduced.

\item {\bf Open access to data and code}
    \item[] Question: Does the paper provide open access to the data and code, with sufficient instructions to faithfully reproduce the main experimental results, as described in supplemental material?
    \item[] Answer: \answerYes{}
    \item[] Justification: Following open-science principles, all data, code, and model weights are publicly released at \url{https://anonymous.4open.science/r/mixture-of-masters}. The only exception is the set of behavioral style models, which is withheld for ethical reasons outlined in ~\cref{app:ethical-concerns}.

\item {\bf Experimental setting/details}
    \item[] Answer: \answerYes{}
    \item[] Justification: We specify all training and test details necessary to understand and reproduce our results, including the explored hyperparameter space and the underlying rationales. This information is comprehensively documented in ~\cref{app:implementation_hw}.

\item {\bf Experiment statistical significance}
    \item[] Question: Does the paper report error bars suitably and correctly defined or other appropriate information about the statistical significance of the experiments?
    \item[] Answer: \answerYes{}
    \item[] Justification: For all bootstrapped analyses presented in the main paper and Supplementary Material, we report standard errors to convey statistical variability (e.g., Table~\ref{tab:expert_results}).

\item {\bf Experiments compute resources}
    \item[] Question: For each experiment, does the paper provide sufficient information on the computer resources (type of compute workers, memory, time of execution) needed to reproduce the experiments?
    \item[] Answer: \answerYes{} 
    \item[] Justification: We provide sufficient information to reproduce our experiments, including details on VRAM requirements and compute time, as documented in ~\cref{app:implementation_hw}.
    
\item {\bf Code of ethics}
    \item[] Question: Does the research conducted in the paper conform, in every respect, with the NeurIPS Code of Ethics \url{https://neurips.cc/public/EthicsGuidelines}?
    \item[] Answer: \answerYes{}
    \item[] Justification: We affirm that the research presented in this paper fully complies with the NeurIPS Code of Ethics.

\item {\bf Broader impacts}
    \item[] Question: Does the paper discuss both potential positive societal impacts and negative societal impacts of the work performed?
    \item[] Answer: \answerYes{}
    \item[] Justification: We elaborate on the potential societal impacts of this work in ~\cref{app:impact-statement}.
    
\item {\bf Safeguards}
    \item[] Question: Does the paper describe safeguards that have been put in place for responsible release of data or models that have a high risk for misuse (e.g., pre-trained language models, image generators, or scraped datasets)?
    \item[] Answer: \answerYes{}
    \item[] Justification: For ethical reasons, we make available only the code and weights for our chess language models, while excluding the chess behavioral style models to mitigate potential misuse. All data used in this work are sourced from publicly available, open-source resources. A more detailed discussion of these safeguards is provided in ~\cref{app:ethical-concerns}

\item {\bf Licenses for existing assets}
    \item[] Question: Are the creators or original owners of assets (e.g., code, data, models), used in the paper, properly credited and are the license and terms of use explicitly mentioned and properly respected?
    \item[] Answer: \answerYes{}
    \item[] Justification: We properly acknowledge the creators of all external assets used in our work, including code, data, and models. All licenses and terms of use have been carefully reviewed, respected, and are detailed in ~\cref{app:data}.

\item {\bf New assets}
    \item[] Question: Are new assets introduced in the paper well documented and is the documentation provided alongside the assets?
    \item[] Answer: \answerYes{}
    \item[] Justification: The assets introduced in this work include the grandmaster dataset, the MoE and single-expert chess language models, as well as the accompanying implementation code. Dataset construction is documented in ~\cref{app:data}. Model architecture and implementation details are provided in \cref{subsec:experts} and \cref{subsec:experimental_setup}, and further elaborated in ~\cref{app:implementation_hw}.

\item {\bf Crowdsourcing and research with human subjects}
    \item[] Question: For crowdsourcing experiments and research with human subjects, does the paper include the full text of instructions given to participants and screenshots, if applicable, as well as details about compensation (if any)? 
    \item[] Answer: \answerNA{}
    \item[] Justification: The paper does not involve crowdsourcing or annotations from human subjects. All qualitative examples were analyzed directly by the authors.

\item {\bf Institutional review board (IRB) approvals or equivalent for research with human subjects}
    \item[] Question: Does the paper describe potential risks incurred by study participants, whether such risks were disclosed to the subjects, and whether Institutional Review Board (IRB) approvals (or an equivalent approval/review based on the requirements of your country or institution) were obtained?
    \item[] Answer: \answerNA{}
    \item[] Justification: The paper does not involve crowdsourcing nor research with human subjects.

\item {\bf Declaration of LLM usage}
    \item[] Question: Does the paper describe the usage of LLMs if it is an important, original, or non-standard component of the core methods in this research? Note that if the LLM is used only for writing, editing, or formatting purposes and does \emph{not} impact the core methodology, scientific rigor, or originality of the research, declaration is not required.
    \item[] Answer: \answerNA{}
    \item[] Justification: The core method development in this research does not involve LLMs as any important, original, or non-standard components.

\end{enumerate}

\end{document}